\documentclass[11pt]{article}
\usepackage[final]{acl}

\usepackage{times}
\usepackage{latexsym}
\usepackage[T1]{fontenc}
\usepackage[utf8]{inputenc}
\usepackage{microtype}
\IfFileExists{inconsolata.sty}{\usepackage{inconsolata}}{}

\usepackage{graphicx}
\usepackage{amsmath}
\usepackage{amssymb}
\usepackage{tablefootnote}
\usepackage{caption}
\usepackage{amsthm}
\usepackage{booktabs}
\usepackage{multirow}
\usepackage{xcolor}

\usepackage{tikz}
\usepackage{pgfplots}
\pgfplotsset{compat=1.18}
\usetikzlibrary{positioning,arrows.meta,patterns,calc,fit}

\usepackage{enumitem}
\setlist{nosep}

\usepackage[most]{tcolorbox}
\tcbset{
  pseudocodebox/.style={
    breakable,
    colback=gray!4,
    colframe=gray!55!black,
    boxsep=2pt,
    left=3pt, right=3pt, top=3pt, bottom=3pt,
    arc=2pt,
    fonttitle=\bfseries\footnotesize,
    coltitle=black,
    colbacktitle=gray!12,
    title=#1,
    enhanced,
  }
}

\definecolor{tHRL}{HTML}{2563EB}
\definecolor{tLRL}{HTML}{D97706}
\definecolor{tHO}{HTML}{DC2626}
\definecolor{tFr}{HTML}{475569}
\definecolor{tCal}{HTML}{0EA5E9}

\DeclareRobustCommand{\HRL}{\textsc{hrl}}
\DeclareRobustCommand{\LRL}{\textsc{lrl}}

\DeclareRobustCommand{\MCtwo}{MetaCLIP\nobreakdash-2}
\DeclareRobustCommand{\SLtwo}{SigLIP\nobreakdash-2}
\DeclareRobustCommand{\Rat}{\textsc{r}@1}
\newcommand{\cka}{\mathrm{CKA}}
\DeclareMathOperator{\patch}{patch}
\newcommand{\Esem}{E}
\newcommand{\Ecal}{E_{\mathrm{cal}}}

\title{Where Do Multilingual Vision-Language Encoders Fail on Low-Resource Languages?}
\hypersetup{pdftitle={Where Do Multilingual Vision-Language Encoders Fail on Low-Resource Languages?}}

\author{
  Donghoon Han\thanks{These authors contributed equally.} \quad
  SungHyun Moon\footnotemark[2] \quad
  Aidyn Zhakatayev \quad
  Junghun Cha \quad
  SeungJae Lee \\
  Dnotitia Inc.\\
  \texttt{\{dhk1349,\,sunghyun,\,aidyn,\,jhcha,\,seungjae.lee\}@dnotitia.com}
}

\begin{document}
\maketitle

\begin{abstract}
Recent multilingual vision--language encoders cover hundreds of languages in a single model, yet on two state-of-the-art instances retrieval on low-resource languages (\LRL{}; e.g.\ Swahili) trails high-resource ones (\HRL{}; e.g.\ English) by $30^+$\,pp. We ask where in the trained encoder this gap is located. Prior modality-gap and cross-lingual subspace work suggests a linear language direction at the output crowds out alignment-relevant geometry. We falsify this: LEACE drives the linear language classifier from $>99\%$ to near chance and iterated INLP to $37$--$50\%$ while \LRL{} retrieval moves within $\pm 1.5$\,pp and all tier means within $2.2$\,pp, tracking random controls. The linear bias is a \emph{symptom}, not the cause. Instead, the alignment-causal factor lies along the encoder's forward path: the EOS (end-of-sequence) hidden state's per-language trajectory diverges with depth. Substituting the EOS with its parallel English value three blocks before the projector lifts Swahili from $22.1\%$ to $69.1\%$ on one encoder (and reproduces on the other); three controls rule out pooled-position tautology and English specificity. A front-layer trunk that pulls each language's projection toward the parallel-content centroid corroborates the diagnosis at training time, recovering $+9.6$ / $+17.1$\,pp on \LRL{} XM3600 retrieval (1{,}000-image subset), with consistent gains across three further benchmarks while preserving \HRL{} performance.
\end{abstract}

\section{Introduction}
\label{sec:intro}

Multilingual dual-encoder models map images and text into a shared embedding space for retrieval, classification, and downstream multimodal pipelines across dozens to hundreds of languages \citep{conneau2020unsupervised,xue2021mt5}. Recent single-encoder releases---\SLtwo{} \citep{tschannen2025siglip2} with bidirectional attention and \MCtwo{} \citep{xu2025metaclip2} with causal attention---scale this coverage, but standard image--text retrieval still shows a $30^+$\,pp Recall@1(\Rat{}) gap between high-resource languages (\HRL{}; e.g.\ English, French, German) and low-resource languages (\LRL{}; e.g.\ Bengali, Swahili, Telugu) on \MCtwo{} and \SLtwo{}, reproduced on XM3600 \citep{thapliyal2022xm3600}, Flickr30k-200 and XTD-200 \citep{visheratin2024nllbclip}, and Babel-ImageNet \citep{geigle2024babelimagenet}. This suggests that the curse-of-multilinguality issue \citep{conneau2020unsupervised} persists at the multimodal layer and may affect downstream multilingual multimodal tasks; benchmark-level analyses, however, do not explain where or why representations diverge along the encoder forward path. We localise this gap, test the linear-bias account, and evaluate an alternative through controlled intervention.

\paragraph{A natural starting point.} At the output text embedding, language identity is linearly extractable ($>99\%$ classifier accuracy on both encoders). Motivated by modality-gap analyses \citep{liang2022modalitygap} and cross-lingual subspace work \citep{conneau2020unsupervised,xue2021mt5}, we test whether this direction displaces alignment-relevant geometry and whether erasing it improves retrieval. With LEACE \citep{belrose2023leace} and INLP \citep{ravfogel2020inlp} (Section~\ref{sec:orth}), LEACE reduces classifier accuracy to near chance and INLP attenuates it to $37$--$50\%$ at rank 128, but \LRL{} retrieval stays within $\pm 1.5$\,pp and matches same-rank random controls. The signal is easy to erase but appears symptomatic rather than alignment-causal.

\paragraph{Where the causal factor lies.} We probe the forward path with EOS (end-of-sequence) activation patching \citep{vig2020causal,meng2022rome}: at layer $\ell$, we overwrite the target caption's EOS hidden state with its parallel English value, run the frozen back-half, and score against the image gallery (Section~\ref{sec:causal}). Three blocks before the projector, this rescues \LRL{} retrieval to near the single-source English reference on both encoders; mid-token-patch, random-English-EOS, and source-language generalisation controls rule out pooled-position tautology and English specificity. The causal factor is better described as the EOS hidden state's per-language \emph{forward-path trajectory}: \LRL{} values diverge from English at the first measured block and stay further apart with depth, and this divergence---not the output bias direction---controls retrieval.

\paragraph{Trunk calibration as a training-time test of the diagnosis.} Because patching requires oracle parallel-content substitution, we fine-tune the first $M$ text-encoder blocks---a \emph{front-layer trunk}---as a general-input test: each language's projection moves toward the parallel-content anchor while the back-half, projection head, and vision tower remain frozen (Section~\ref{sec:approx}). Consistent with the trajectory account, the trunk reshapes per-layer geometry (Section~\ref{sec:analysis}) and recovers a substantial fraction of \LRL{} retrieval on XM3600 and three further benchmarks while preserving \HRL{} performance (Section~\ref{sec:results}). We treat this as convergent evidence for the trajectory account, not a re-derivation of patching: a same-depth single-row patch recovers far less than the trunk, while the deep-layer patch reaches a single-source English reference the trunk does not, and a multi-\HRL{} centroid exceeds even that (Section~\ref{sec:generalization_english}).

\paragraph{Scope of the causal claim.} What we establish is an \emph{interventional localisation within trained encoders}: in a fixed released encoder the alignment-relevant failure can be located on, and repaired through, the EOS hidden state's forward-path trajectory. We do not claim a complete causal account of why that divergence arises during pretraining; data prevalence, tokenisation, attention regime and optimisation dynamics all remain candidate upstream causes (Section~\ref{sec:discussion}).

\paragraph{Contributions.}
\begin{itemize}
  \item \textbf{Rejecting linear-bias displacement.} LEACE erases the language-identifying direction and INLP strongly attenuates it, yet \LRL{} retrieval remains within $\pm 1.5$\,pp of baseline and no tier mean moves by more than $2.2$\,pp.
  \item \textbf{Mechanistic localisation.} EOS-position activation patching with three controls localises the alignment-causal factor to the EOS hidden state's forward-path trajectory and rescues \LRL{} retrieval to the English-reference level on causal and bidirectional encoders.
  \item \textbf{Training-time corroboration.} A front-layer trunk yields substantial \LRL{} gains across four benchmarks and reshapes per-layer cross-lingual similarity as predicted, targeting the same trajectory-level object as patching without reducing to it.
\end{itemize}

\paragraph{Scope.} We study the \emph{text encoder} in dual-encoder image-text models. The two main-paper encoders cover the family’s main architectural variants (causal \MCtwo{}; bidirectional \SLtwo{}), and three additional encoders (AltCLIP, NLLB-CLIP-L, mSigLIP; Appendix~\ref{sec:app-morevlm}) test cross-family generalisation. Generative VLMs and training-time dynamics are out of scope.
\section{Related Work}
\label{sec:related}

\paragraph{Representation geometry and concept erasure.} The text--image modality gap was characterised by \citet{liang2022modalitygap}. Cross-lingual representation analyses using linear probing \citep{conneau2020emerging,pires2019mbert} and CKA \citep{kornblith2019cka} suggest universal subspace structure \citep{chang2022multigeom,jha2025platonic}. LEACE \citep{belrose2023leace}, together with earlier INLP \citep{ravfogel2020inlp} and all-but-the-top \citep{mu2017abtt}, provides closed-form linear concept erasure; related debiasing results show that a linearly extractable concept need not be causal. Motivated by prior geometry work, Section~\ref{sec:orth} applies such erasure to multilingual vision-language encoder embeddings up to rank 128 and finds no evidence that the language-identifying direction is load-bearing.

\paragraph{Activation patching and partial-encoder distillation.} EOS-position activation substitution is a special case of mechanistic-interpretability patching \citep{vig2020causal,meng2022rome,wang2022ioi}, related to ``language surgery'' at intermediate layers of multilingual LLMs \citep{lopo2025langsurgery}. Section~\ref{sec:patching} adapts patching to dual-encoder retrieval with parallel-content and random-content controls, separating substantive rescue from pooled-position tautology. Section~\ref{sec:approx}'s trunk adapts multilingual sentence-embedding distillation \citep{reimers2020multilingual} to a partial-encoder setting (front-$M$-only, InfoNCE plus centroid-anchored cosine rather than MSE); its novel element is that it targets the trajectory locus identified by patching. The \HRL{}-averaging boost in Section~\ref{sec:hrl-avg} is consistent with meta-embedding variance reduction \citep{coates2018meta}.

\paragraph{Data sources.} \textbf{XM3600} \citep{thapliyal2022xm3600} carries the headline retrieval and all per-layer/back-half probes; \textbf{Flickr30k-200} and \textbf{XTD-200} \citep{visheratin2024nllbclip}, and \textbf{Babel-ImageNet} \citep{geigle2024babelimagenet} provide cross-benchmark replication (Appendix~\ref{sec:app-benchmarks}); \textbf{FLORES-200 devtest} \citep{nllb2022flores200} with \textbf{COMET-22} \citep{rei2022comet22} serve only as the reference parallel corpus and metric for the GLM-5.1 translator-quality check selecting the 11-language trunk training pool (Appendix~\ref{sec:app-flores-correlation}).
\section{Setup and Notation}
\label{sec:setup}

\paragraph{Models and states.} We use two encoders: \MCtwo{} Worldwide-Huge \citep{xu2025metaclip2}, a 24-block causal-attention text tower with dynamic EOS at token-id 2 and a 77-token context, and \SLtwo{} SO400M \citep{tschannen2025siglip2}, a 27-block bidirectional-attention text tower with sticky EOS at position $-1$ and a 64-token context. Each is evaluated as \emph{frozen} (released weights) and \emph{trunk-calibrated} (the first $M$ text-encoder blocks replaced by a translation-aligned trunk; $M{=}4$ for \MCtwo{} and $M{=}3$ for \SLtwo{}; Section~\ref{sec:approx}).

\paragraph{Notation.} Let $\Esem : \mathcal{X} \to \mathbb{R}^D$ be the text encoder, decomposed as $\Esem = \pi \circ \varphi_{N-1} \circ \cdots \circ \varphi_0$, with the $N$ blocks numbered from $0$ and $\pi$ the projection head including the final LayerNorm. For a tokenized caption $x^L$ in language $L$, let $H_k(x^L) \in \mathbb{R}^{n_L \times d}$ be the hidden-state matrix after $k$ blocks, $h_k(x^L) \in \mathbb{R}^d$ its pooled (EOS) row, and $t_L(x) := \Esem(x^L)$ the output text embedding. Block $\ell$ emits $H_{\ell+1}$, so $\ell = N{-}1$ is the projector's input and $\ell = N{-}4$ leaves three blocks. Trunk calibration (Section~\ref{sec:approx}) replaces $\varphi_0, \ldots, \varphi_{M-1}$, so its depth-$M$ state is $H_M$, pooled row $h_M$.

\paragraph{Trajectory and its divergence.} The \emph{forward-path trajectory} of a caption is the sequence of pooled-row hidden states $\bigl(h_1(x^L), \ldots, h_N(x^L)\bigr)$ it produces, and we measure how far language $L$'s trajectory runs from its parallel English counterpart by
\begin{equation}
  d_\ell(L) \;:=\; \mathbb{E}_x\bigl[\,1 - \cos\bigl(h_{\ell+1}(x^L),\, h_{\ell+1}(x^{\mathrm{en}})\bigr)\bigr],
  \label{eq:dL-def}
\end{equation}
with $x$ ranging over parallel caption groups. Equation~\ref{eq:dL-def} is invariant to rotation and to the per-layer scale of $h_{\ell+1}$; the residual $r(L)$ of Section~\ref{sec:lipschitz} is its unnormalised $\ell_2$ counterpart at the single depth $M$, where an absolute distance is needed.

\paragraph{Languages and data.} The language pool contains 11 languages: \HRL{}-6 (English [en], French [fr], German [de], Spanish [es], Chinese [zh], Korean [ko]) and \LRL{}-5 (Bengali [bn], Filipino [fil], Hindi [hi], Swahili [sw], Telugu [te]). The \LRL{}-5 are the trained-pool subset of the seven low-resource languages designated by \citet{wang2025scaling} based on WebLI-100B prevalence ($0.001$--$0.267\%$, retrieval-independent); Hebrew and Maori are excluded because Hebrew lacks coverage in our XM3600/Flickr30k-200/XTD-200 protocol, and Maori fails the COMET-22 $\geq 0.82$ translator-quality bar under both GLM-5.1 and GPT-5.2 (Appendix~\ref{sec:app-flores-correlation}). All other pool languages appear in the standard test splits of XM3600 \citep{thapliyal2022xm3600}, and Flickr30k-200 and XTD-200 \citep{visheratin2024nllbclip} (Appendix~\ref{sec:app-benchmarks}). The mechanistic image--text experiments use the $1{,}000$-image XM3600 retrieval subset, so cross-lingual pairs share content and visual grounding.

\section{The Linear Language Direction}
\label{sec:orth}

We establish two facts about $t_L(x)$: language identity is linearly extractable, and cross-lingual averaging lifts retrieval even on the \HRL{} pool. A natural explanation is that averaging cancels this direction and that the same direction gates \LRL{} retrieval; LEACE tests this link and does not support it, suggesting a symptom rather than an alignment-causal factor.

\subsection{Language Identifiability of the Projected Embedding}
\label{sec:clf}

We train classifiers on COCO-translated captions and test on disjoint XM3600 to avoid corpus-specific surface patterns. GPT-5.2\footnote{GPT-5.2-2025-12-11.} produced translations into the 13 probe languages; these files will be released with the codebase, so the probe is reproducible without translator access (Appendix~\ref{sec:app-clf-layerwise}). A 13-way logistic-regression classifier on $t_L(x) \in \mathbb{R}^D$ detects source language with $99.8\%$ accuracy on \MCtwo{} and $99.7\%$ on \SLtwo{} (chance $1/13 \approx 7.7\%$); linear SVM and single-hidden-layer MLP reach $96$--$98\%$, so the signal is not classifier-specific. A per-layer probe shows the signal from $\ell{=}0$, not only after the projection head: \MCtwo{} stays at $96$--$99\%$, while \SLtwo{} stays at $71$--$83\%$ because bidirectional attention spreads the signal across positions (full curves in Appendix~\ref{sec:app-clf-layerwise}).

\paragraph{\HRL{}-averaging boost.}\label{sec:hrl-avg}
Averaging the six \HRL{} embeddings of a parallel caption lifts \MCtwo{} retrieval by $+17.4$\,pp on XM3600 ($78.1 \to 95.5$ on \HRL{}-6; Appendix~\ref{sec:app-extra-avgresidual}). This generalizes: \HRL{}-pool averaging yields $+5.6$ to $+38.5$\,pp across five multilingual vision-language encoders (\MCtwo{}, \SLtwo{}, AltCLIP, NLLB-CLIP-L, mSigLIP) on XM3600 and Flickr30k-200 \citep{visheratin2024nllbclip}, and \LRL{}-pool averaging is positive in every (encoder, benchmark) cell, with $\geq +10$\,pp on the \LRL{}-only pool (Appendix~\ref{sec:app-extra-avg-broad}); the centroid geometry behind the boost --- pool dispersion, pre-normalisation norms, centroid--image similarity --- is in Appendix~\ref{sec:app-avg-geometry}. If the boost comes from canceling the language-identity direction in Section~\ref{sec:clf}, erasing that direction should produce a similar lift; we test this with LEACE.

\subsection{LEACE: Erasing the Linear Language Direction}
\label{sec:leace}

LEACE \citep{belrose2023leace} returns the minimum-perturbation closed-form projector $P_{\mathrm{lang}}$ under which no linear classifier exceeds chance on the eraser output. It drops classifier accuracy from $\geq 99\%$ to chance, but image-to-text retrieval changes by $\leq 1.5$\,pp and matches a same-rank random orthogonal \emph{ablation} (rand-remove-12, $I - V V^\top$; Table~\ref{tab:leace}). Higher-rank Iterated Null-space Projection (INLP, \citealp{ravfogel2020inlp}) gives the same pattern: at ranks $\{12, 64, 128\}$, rank-128 erasure drops the classifier to $37$--$50\%$, yet \LRL{} retrieval still changes by $\leq 1.5$\,pp and no tier mean by more than $2.2$\,pp (Appendix~\ref{sec:app-inlp}). Thus, the language direction is recoverable, but removing it does not appear retrieval-causal up to rank 128.

\begin{table}[t]
\centering
\small
\setlength{\tabcolsep}{4pt}
\begin{tabular}{l|cc|cc}
\toprule
 & \multicolumn{2}{c|}{\MCtwo{}} & \multicolumn{2}{c}{\SLtwo{}} \\
 & \HRL{} & \LRL{} & \HRL{} & \LRL{} \\
\midrule
base \Rat{}     & 78.1 & 44.1 & 63.3 & 21.7 \\
LEACE \Rat{}    & 77.5 & 43.0 & 64.0 & 21.6 \\
rand-remove-12 \Rat{}  & 78.2 & 43.6 & 63.0 & 21.8 \\
\bottomrule
\end{tabular}
\caption[LEACE ablation on XM3600]{
LEACE ablation on XM3600 ($1{,}000$ images, $11$ trained languages).
Entries are i2t \Rat{} in \% for the original embedding, LEACE-projected embedding, and same-rank random orthogonal ablation ($I - V V^\top$). LEACE drops classifier accuracy to chance ($\sim\!99.8\% \to 3$--$5\%$, chance $1/13 \approx 7.7\%$)\protect\footnotemark{}, while retrieval changes by $\leq 1.5$\,pp and matches rand-remove-12. Reported INLP checkpoints $\{12,64,128\}$ replicate this; the full sweep
additionally includes rank 32 (Appendix~\ref{sec:app-inlp}).
}
\label{tab:leace}
\end{table}
\footnotetext{Below-chance post-erasure accuracy can occur because the eraser is fit on COCO activations and the classifier is refit OOD on XM3600; chance is uniform $1/13$.}

\paragraph{Reconciling the averaging boost and the LEACE null.} The content redundancy exploited by averaging and the language-identity direction erased by LEACE appear to occupy \emph{different subspaces} (Appendix~\ref{sec:app-extra-avgresidual}). Linear bias displacement is not supported up to rank 128, while the non-linear regime remains inconclusive under an adversarial-MLP eraser (Appendix~\ref{sec:app-r1-nonlinear}); the detectable bias is therefore best read as language-identity encoding, not the cause of the retrieval gap.
\section{Pinpointing the Locus on the Forward-Path Trajectory}
\label{sec:causal}

Section~\ref{sec:orth} showed that erasing the linear language direction at the output text embedding leaves \LRL{} retrieval within $\pm 1.5$\,pp, so that direction is unlikely to be the sole alignment-causal factor; we therefore inspect the encoder's forward-path trajectory with single-row activation patching \citep{vig2020causal,meng2022rome}.

\subsection{Protocol}
\label{sec:patching}

For target language $T$ and source language $A$,
\begin{equation*}
  \patch_{\ell,A\to T}(x^T) \;:=\; \pi \circ \varphi_{N-1} \circ \cdots \circ \varphi_{\ell+1}(\widetilde{H}_{\ell+1}),
\end{equation*}
where $\widetilde{H}_{\ell+1}$ is $H_{\ell+1}(x^T)$ with its pooled row replaced by $h_{\ell+1}(x^A)$ and all other rows kept; the later blocks are unchanged, so this is a single-row residual-stream intervention.

For each XM3600 image-caption row ($n{=}1{,}000$), we cache $h_{\ell+1}(x^{\mathrm{en}})$ from the parallel English caption at every block; for each $T \neq \mathrm{en}$ and stride-2 patch layer $\ell$, we substitute only the target EOS row and evaluate against the fixed image embedding gallery. The parallel-English anchor is the EN-swap reference, and norm-matched random Gaussian substitution is the basic control.

\subsection{Rescue Curves}
\label{sec:patch-curves}

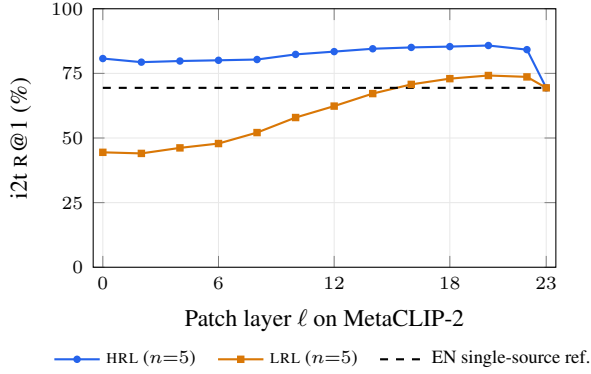
\begin{figure}[t]
\centering
\begin{tikzpicture}
\begin{axis}[
  width=\columnwidth, height=5.0cm,
  xlabel={Patch layer $\ell$ on \MCtwo{}},
  ylabel={i2t \Rat{} (\%)},
  xmin=-0.5, xmax=23.5, ymin=0, ymax=100,
  xtick={0,6,12,18,23}, ytick={0,25,50,75,100},
  axis background/.style={fill=none},
  axis on top=false,
  legend style={font=\scriptsize, at={(0.5,-0.28)}, anchor=north, fill=none, draw=none, row sep=-2pt, inner sep=2pt},
  legend columns=3, /tikz/every even column/.append style={column sep=6pt},
  tick label style={font=\scriptsize}, label style={font=\footnotesize},
  grid=both, grid style={gray!18, very thin},
]
\addplot[color=tHRL, mark=*, mark size=1pt, thick] coordinates {
(0,80.74)(2,79.34)(4,79.78)(6,80.06)(8,80.36)(10,82.34)(12,83.42)(14,84.52)(16,85.04)(18,85.36)(20,85.78)(22,84.18)(23,69.40)
};
\addlegendentry{\HRL{} ($n{=}5$)}
\addplot[color=tLRL, mark=square*, mark size=1pt, thick] coordinates {
(0,44.46)(2,44.04)(4,46.18)(6,47.86)(8,52.10)(10,57.92)(12,62.34)(14,67.18)(16,70.74)(18,72.96)(20,74.18)(22,73.64)(23,69.40)
};
\addlegendentry{\LRL{} ($n{=}5$)}
\addplot[color=black, dashed, thick] coordinates {(0,69.4)(23,69.4)};
\addlegendentry{EN single-source ref.}
\end{axis}
\end{tikzpicture}
\caption{Tier-averaged XM3600 image-to-text \Rat{} ($1{,}000$ images) across EOS-patch layers $\ell$ for frozen \MCtwo{}; dashed line: single-source English reference ($69.4\%$).}
\label{fig:patch-sweep}
\end{figure}

\paragraph{Rescue curves.}
Figure~\ref{fig:patch-sweep} shows the \MCtwo{} patch sweep. We headline $\ell=N{-}4$, which passes through three unmodified back-half blocks: averaged over the \LRL{}-5 pool, retrieval rises from $44.1\%$ to $73.4\%$ ($+29.3$\,pp), Swahili from $22.1\%$ to $69.1\%$ and Filipino from $39.9\%$ to $73.5\%$. The best single layer adds little (Swahili $71.0\%$ at $\ell=22$). Cluster bootstrap ($B{=}1{,}000$ row resamples, pairing preserved) gives $95\%$ CIs of $[+45.5,+52.2]$ for Swahili and $[+30.1,+37.1]$ for Filipino at their best layers ($22$, $20$; Appendix~\ref{sec:app-r1-bootstrap}). The norm-matched random Gaussian control collapses retrieval (\HRL{} $\sim 80 \to 8.5\%$, \LRL{} $\to 1.6\%$), so the rescue is signal-specific rather than norm-driven; \SLtwo{} shows the same pattern with larger deltas because its \LRL{} baselines are lower. Unlike the layer-confined peak typical of prior patching work \citep{wang2022ioi}, the curve improves toward the near-final region and tends to shrink when more target-language back-half blocks follow the patch; because $\ell=N{-}1$ is partly a pooling tautology (see below), the key non-tautological checkpoint is $\ell=N{-}4$. Table~\ref{tab:patch} reports tier-level rescue at $\ell=N{-}4$ and random-control failure for both encoders.

\begin{table}[t]
\centering
\small
\setlength{\tabcolsep}{4pt}
\begin{tabular}{l|c|cc}
\toprule
&  & \multicolumn{2}{c}{i2t \Rat{} (\%): base $\to$ $\ell{=}N{-}4$} \\
\cmidrule(lr){3-4}
Encoder & EN ref. & \HRL{} & \LRL{} \\
\midrule
\MCtwo{} & 69.4 & $79.8 \to \mathbf{85.8}$ & $44.1 \to \mathbf{73.4}$ \\
\addlinespace[2pt]
\hspace{1em}\textit{random control} & --- & rand $\,8.5$ & rand $\,1.6$ \\
\midrule
\SLtwo{} & 66.1 & $62.8 \to \mathbf{81.8}$ & $21.7 \to \mathbf{66.6}$ \\
\addlinespace[2pt]
\hspace{1em}\textit{random control} & --- & rand $16.4$ & rand $\,3.2$ \\
\bottomrule
\end{tabular}
\caption{XM3600 i2t \Rat{} (\%), baseline to the $\ell{=}N{-}4$ patch, averaged within non-English \HRL{} and \LRL{} targets (n=5 each). The best layer $\ell^\star$ adds $\leq{+}4.2$ pp beyond $\ell{=}N{-}4$ (\LRL{} peaks $74.3$/$70.8$). EN ref.: single-source English reference; \textit{random control}: norm-matched Gaussian EOS at $\ell^\star$. CIs: Appendix~\ref{sec:app-r1-bootstrap}.}
\label{tab:patch}
\end{table}

\paragraph{Disentangling pooled-position tautology from substantive rescue.}
\label{sec:tautology}
At $\ell=N{-}1$, the EOS row is the projector's pooled input, making substitution a near-tautological upper reference. Three controls separate this endpoint from substantive rescue: \textbf{(a)}~a non-EOS mid-sentence patch at $\ell=N{-}1$ leaves \LRL{} retrieval essentially unchanged; \textbf{(b)}~the EOS patch at $\ell=N{-}4$ passes through three unmodified back-half blocks and still rescues Swahili $22.1 \to 69.1\%$ and Filipino $39.9 \to 73.5\%$ on \MCtwo{}; and \textbf{(c)}~a \emph{random} English EOS at $\ell=N{-}4$ collapses retrieval below baseline (\MCtwo{} \LRL{} $44.1 \to 7.7$; \SLtwo{} $21.7 \to 5.2$). Thus, the rescue relies on parallel-content English structure, not merely an English-distributed vector. Full per-language and per-layer numbers are in Appendix~\ref{sec:app-r1-controls}.

\paragraph{Generalization beyond English.}
\label{sec:generalization_english}
The rescue is not English-specific: at $\ell=N{-}4$, a parallel French EOS lifts Swahili to $78.7\%$ (vs.\ $69.1\%$ from English at the same depth), and the \emph{mean} of the six \HRL{} EOS states lifts it to $88.0\%$, above the single-source English reference. This is consistent with the trajectory account: if alignment depends on the EOS hidden state's location along a parallel-content cone, a multi-\HRL{} centroid can sit closer to the cone's interior than any single EOS and act as a variance-reduced source, paralleling projector-output averaging (Section~\ref{sec:hrl-avg}). The single-source English level is therefore an EN-EOS-swap reference, not a hard upper bound: at $\ell=N{-}4$, the English-EOS patch brings Swahili essentially to the EN reference ($69.1$ vs.\ $69.4$; $-0.3$\,pp), while $71.0\%$ is the separate best-layer English-EOS peak at $\ell=22$. Full numbers are in Appendix~\ref{sec:app-r1-multi-hrl}.
\section{Trunk Calibration: A Controlled Intervention}
\label{sec:approx}

Sections~\ref{sec:orth}--\ref{sec:causal} rule out an output linear-bias account and instead point to the EOS hidden state's per-language forward-path trajectory. Trunk calibration tests this by re-fitting the first $M$ blocks so their depth-$M$ outputs, after the frozen back-half, move toward a shared parallel-content target. We choose depth $M$ as the earliest tractable point on this trajectory: front blocks are trainable and the back-half is frozen. We ask whether shaping that state yields a population-level retrieval lift; its magnitude need not match single-row patching, which uses oracle parallel content at inference and acts deeper along the trajectory (Appendix~\ref{sec:app-r1-patch-at-M}).

\paragraph{Architecture and objective.} Let $M$ be the front-layer depth ($M{=}4$ for \MCtwo{}, $M{=}3$ for \SLtwo{}; sweep in Appendix~\ref{sec:app-sweep}). The trunk-calibrated text encoder replaces blocks $\{\varphi_0, \ldots, \varphi_{M-1}\}$ with trainable copies $\{\varphi'_0, \ldots, \varphi'_{M-1}\}$ initialised from the frozen weights (\emph{soft fine-tune}; hard-reinit ablation in Appendix~\ref{sec:app-extra-hardinit}), while the back-half $\varphi_M, \ldots, \varphi_{N-1}$, projector $\pi$, and vision tower stay frozen. The loss is at the calibrated projected output, not directly at depth $M$:
\begin{equation}
\begin{aligned}
  \mathcal{L}_{\mathrm{trunk}} \;:=\;& \mathcal{L}_{\mathrm{InfoNCE}}\bigl(\{z_g^L\},\, \{a_g\}\bigr) \\
  &{}+ \lambda \,\mathbb{E}_{g, L}\bigl[\,1 - \langle z_g^L,\, a_g \rangle\,\bigr].
\end{aligned}
\label{eq:trunkloss}
\end{equation}
Here $z_g^L := \Ecal(x_g^L)$ is the calibrated projected embedding of the language-$L$ caption in parallel group $g$, and $a_g$ is the frozen-encoder anchor for that group: the renormalised mean of the normalised frozen embeddings over the anchor set $\mathcal{A}$ (written out in Appendix~\ref{sec:app-pseudo}). Embeddings are $\ell_2$-normalised, so the inner product is the cosine. Canonically $\mathcal{A}$ is all 11 trained languages; we select it on a joint criterion --- strongest training-time paired cosine and centroid alignment at near-best held-out \LRL{} retrieval --- and ablate the \HRL{}-6-only and English-only alternatives in Appendix~\ref{sec:app-sweep}. Because gradients pass through the frozen back-half, the trunk can reshape only the depth-$M$ state; the sensitivity probe below gives a sanity envelope.

\paragraph{Training data and hyperparameters.}
Training uses text-only parallel-translation groups over the 11-language pool: English captions from a 1M-caption subset of CC12M \citep{changpinyo2021cc12m} are translated by GLM-5.1, a point release in the GLM-5 family \citep{glm2026glm5}. We train with $128$ caption-groups for $20{,}000$ AdamW steps, lr $1.2{\times}10^{-4}$ (\MCtwo{}) / $2.4{\times}10^{-4}$ (\SLtwo{}), and $\sim 50$M / $\sim 40$M trainable parameters. Full recipe and pseudocode are in Appendix~\ref{sec:app-pseudo}.\footnote{Code, checkpoints, and diagnostic probes will be released at \url{https://github.com/dnotitia/geometric-bottleneck}.}

\paragraph{Translation-quality sanity check.} On the trained pool, GLM-5.1 reaches per-language COMET-22 quality 0.816--0.905; Hindi is the boundary case under GLM-5.1 (0.816) and clears the \(\geq 0.82\) bar under GPT-5.2, while the other ten trained-pool languages clear it under both (Appendix~\ref{sec:app-flores-correlation}). Downstream evaluation uses natively-multilingual XM3600 and other professionally-translated benchmarks (Appendix~\ref{sec:app-benchmarks}), so the parallel-translation data enters only through the training objective.

\paragraph{Lipschitz bridge from projected-space loss to depth-$M$ intervention.}
\label{sec:lipschitz}
The loss is computed at the projected output, but only the first $M$ blocks are trainable; the back-half $\psi := \pi \circ \varphi_{N-1} \circ \cdots \circ \varphi_M$ is a fixed map from depth-$M$ hidden-state matrices to projected embeddings, probed along the pooled row with the other rows fixed. Define the per-language depth-$M$ residual
\begin{equation}
  r(L) \;:=\; \mathbb{E}_x \bigl\lVert h_M(x^L) - h_M(x^{\mathrm{en}}) \bigr\rVert,
  \label{eq:rL-def}
\end{equation}
taking both terms in the same state, so that $r(\mathrm{en}) = 0$. The trunk objective targets $a_g$, the 11-language projected anchor of Equation~\ref{eq:anchor}. Write $\Ecal^{L\leftarrow\mathrm{en}}(x) := \psi\bigl(\widetilde{H}_M(x^L; \mathrm{en})\bigr)$ for the pooled-row counterfactual, in which the calibrated depth-$M$ matrix of $x^L$ keeps every non-pooled row and takes the calibrated English pooled row. If $\psi$ is $K$-Lipschitz along that coordinate on the trunk-produced states and their pooled-row counterfactuals, then in expectation
\begin{equation}
  \mathbb{E}_x \bigl\lVert \Ecal(x^L) - \Ecal^{L\leftarrow\mathrm{en}}(x) \bigr\rVert \;\leq\; K \cdot r(L).
  \label{eq:cal-bound}
\end{equation}
The two inputs differ only in the pooled row, and by exactly the residual of Equation~\ref{eq:rL-def}. It therefore concerns only the pooled-row component: it does not bound the full language-to-English discrepancy, because the non-pooled rows also differ. We read it as a sanity envelope, not an identifiability proof; the converse does not follow from $K$-Lipschitzness and would require a bi- or inverse-Lipschitz assumption we do not establish. Because only the front blocks are trainable, the loss can only act through the depth-$M$ distribution. Sampled sensitivity ratios average $0.27$--$0.46$ across both encoders and states, and the per-language scatter of $\bar\rho_L \cdot r(L)$ vs.\ trunk-calibration $\Delta$\Rat{} is moderate (Pearson $0.55$--$0.69$; Appendix~\ref{sec:app-lipschitz}). Within this envelope, optimizing at the projector is operationally a depth-$M$ intervention---the earliest trainable point on the trajectory identified by Sections~\ref{sec:orth}--\ref{sec:causal}.
\section{Retrieval as Evidence for the Diagnosis}
\label{sec:results}

The forward-path-trajectory diagnosis (Section~\ref{sec:causal}) predicts that retrieval should recover if trunk calibration pulls each language's depth-$M$ hidden state toward a parallel-content anchor. We test this prediction through XM3600 per-language retrieval, multi-benchmark replication, and additional-encoder replication, using lift size, language coverage, and propagation through the frozen back-half as evidence.

\paragraph{XM3600 per-language retrieval.}
\phantomsection
\label{sec:results-xm3600}

\begin{table}[t]
\centering
\small
\setlength{\tabcolsep}{4pt}
\begin{tabular}{ll|cc|cc}
\toprule
& & \multicolumn{2}{c|}{\MCtwo{}} & \multicolumn{2}{c}{\SLtwo{}} \\
Lang & Tier & frozen & cal. & frozen & cal. \\
\midrule
en   & \HRL{} & 69.4 & 72.0 & 66.1 & 70.5 \\
fr   & \HRL{} & 84.4 & 86.2 & 78.2 & 82.6 \\
de   & \HRL{} & 86.5 & 86.8 & 33.7 & \textbf{43.8} \\
es   & \HRL{} & 78.0 & 77.6 & 64.1 & 67.4 \\
zh   & \HRL{} & 74.6 & 77.5 & 63.8 & \textbf{70.5} \\
ko   & \HRL{} & 75.5 & 78.9 & 74.0 & 77.0 \\
\midrule
bn   & \LRL{} & 64.7 & 69.0 & 30.2 & \textbf{42.7} \\
fil  & \LRL{} & 39.9 & \textbf{52.7} & 26.9 & \textbf{45.0} \\
hi   & \LRL{} & 49.4 & 52.9 & 33.6 & 38.5 \\
sw   & \LRL{} & 22.1 & \textbf{42.1} & 12.4 & \textbf{35.7} \\
te   & \LRL{} & 44.3 & \textbf{52.0} & \phantom{0}5.2 & \textbf{31.9} \\
\midrule
\textit{\HRL{} mean} & & 78.1 & 79.8 & 63.3 & \textbf{68.6} \\
\textit{\LRL{} mean} & & 44.1 & \textbf{53.7} & 21.7 & \textbf{38.8} \\
\bottomrule
\end{tabular}
\caption{Per-language image-to-text \Rat{} (\%) on the $1{,}000$-image XM3600 subset, frozen vs.\ trunk-calibrated. Bold marks calibrated-vs-frozen lift $\geq +5$\,pp.}
\label{tab:perlang-xm3600}
\end{table}

Calibration yields double-digit \LRL{} gains in $6$ of $10$ language--model pairs: two on \MCtwo{} (sw $+20.0$, fil $+12.8$) and four on \SLtwo{} (bn $+12.5$, fil $+18.1$, sw $+23.3$, te $+26.7$), while \HRL{} \Rat{} is preserved or slightly improved (Table~\ref{tab:perlang-xm3600}); de-\SLtwo{} also rises $+10.1$\,pp from a weak \HRL{} baseline ($33.7 \to 43.8$).

\paragraph{Cross-benchmark replication.} Trunk calibration does not appear to overfit XM3600: $\Delta$ is positive on Flickr30k-200, XTD-200, XM3600, and Babel-ImageNet for both encoders (Table~\ref{tab:bench-mini}). Across the four benchmarks, the \LRL{}-5 pool lifts by $+7.8$\,pp on \MCtwo{} and $+14.9$\,pp on \SLtwo{}, while \HRL{}-6 rises slightly ($+1.7$ / $+3.0$\,pp). Per-language breakdowns are in Appendix~\ref{sec:app-perlang-benchmarks}; CVQA (QA-style) is in Appendix~\ref{sec:app-benchmarks}.

\begin{table}[h]
\centering
\small
\setlength{\tabcolsep}{4pt}
\begin{tabular}{l|cc|cc}
\toprule
 & \multicolumn{2}{c|}{\MCtwo{}} & \multicolumn{2}{c}{\SLtwo{}} \\
Benchmark & $\Delta$\HRL{} & $\Delta$\LRL{} & $\Delta$\HRL{} & $\Delta$\LRL{} \\
\midrule
Flickr30k-200    & $+0.7$ & $+8.4$ & $+3.3$ & $+20.4$ \\
XTD-200          & $+2.2$ & $+9.0$ & $+3.2$ & $+19.1$ \\
XM3600 (full)    & $+1.4$ & $+7.5$ & $+4.3$ & $+10.7$ \\
Babel-ImageNet   & $+2.3$ & $+6.3$ & $+1.3$ & $+9.5$ \\
\midrule
\textit{mean (4 benchmarks)} & $+1.7$ & $\mathbf{+7.8}$ & $+3.0$ & $\mathbf{+14.9}$ \\
\bottomrule
\end{tabular}
\caption{Trunk-calibration $\Delta$ in tier-mean performance (pp), by benchmark, for both encoders (canonical 11-language anchor). Metric is i2t \Rat{} for Flickr30k-200, XTD-200, and XM3600, and top-1 accuracy for Babel-ImageNet. The XM3600 row is the full $3{,}600$-image evaluation suite, not the $1{,}000$-image subset of Table~\ref{tab:perlang-xm3600}. Full table including CVQA in Appendix~\ref{sec:app-benchmarks}; per-language detail in Appendix~\ref{sec:app-perlang-benchmarks}.}
\label{tab:bench-mini}
\end{table}

\paragraph{Replication on additional vision-language encoders.} The diagnosis and trunk recipe also transfer to AltCLIP, NLLB-CLIP-L, and mSigLIP (Appendix~\ref{sec:app-morevlm}), which differ from the main pair in pooling, text-tower scale, and pretraining objective. Pooler-row patching at $\ell{=}N{-}2$ substantially rescues \LRL{} retrieval, and the same front-layer trunk recovers $+10$ to $+37$\,pp on \LRL{} XM3600 and $+9$ to $+51$\,pp on \LRL{} Flickr30k-200. These results suggest that the phenomenon and intervention are not specific to the \MCtwo{}/\SLtwo{} pair.

\section{Geometric Analysis Through Trunk Calibration}
\label{sec:analysis}

Patching (Section~\ref{sec:patching}) localised the alignment-causal factor along the forward path, and the trunk (Section~\ref{sec:approx}) tests the same diagnosis at training time. We use the calibrated state as a controlled comparison to characterise how trunk calibration changes per-layer geometry. Specifically, we compute cross-lingual CKA \citep{kornblith2019cka}, a rotation-invariant similarity, at every text-encoder block on $1{,}000$ image-aligned XM3600 captions. We track \HRL{}$\leftrightarrow$\HRL{} and \HRL{}$\leftrightarrow$\LRL{} similarities, averaged over the relevant unordered pairs; their gap serves as the geometric counterpart of the retrieval tier gap. Top-$K$ Jaccard and modality-gap corroborations on the same XM3600 caption set are reported in Appendix~\ref{sec:app-extra}.

\begin{figure*}[t]
\centering
\begin{minipage}{0.49\textwidth}\centering
\begin{tikzpicture}
\begin{axis}[
  width=\textwidth, height=5.4cm,
  xlabel={Text encoder layer $\ell$}, ylabel={Cross-lingual CKA},
  xmin=-0.5, xmax=24.5, ymin=0.05, ymax=0.70,
  xtick={0,4,8,12,16,20,23}, ytick={0.1,0.2,0.3,0.4,0.5,0.6},
  axis background/.style={fill=none},
  axis on top=false,
  legend to name=cka-shared-legend,
  legend style={font=\scriptsize, draw=none, fill=none, inner sep=2pt},
  legend columns=4, /tikz/every even column/.append style={column sep=8pt},
  tick label style={font=\scriptsize}, label style={font=\footnotesize},
  title style={font=\footnotesize}, title={(a) \MCtwo{} (24 text blocks)},
  grid=both, grid style={gray!18, very thin},
]
\fill[gray!22] (axis cs:-0.5,0.05) rectangle (axis cs:3.5,0.70);
\node[anchor=north west, font=\scriptsize, text=gray!90!black] at (axis cs:-0.2,0.69) {trained ($\ell{<}M$)};
\addplot[color=tHRL, mark=*, mark size=1pt, thick] coordinates {(0,0.2426)(1,0.2369)(2,0.2358)(3,0.1766)(4,0.1621)(5,0.1725)(6,0.1851)(7,0.1915)(8,0.2097)(9,0.2538)(10,0.3078)(11,0.3811)(12,0.4467)(13,0.4743)(14,0.5075)(15,0.5250)(16,0.5566)(17,0.5603)(18,0.5616)(19,0.5662)(20,0.5794)(21,0.5510)(22,0.5609)(23,0.5752)};
\addlegendentry{\HRL{}--\HRL{} (frozen)}
\addplot[color=tHRL, mark=*, mark size=1pt, thick, dashed] coordinates {(0,0.1772)(1,0.1408)(2,0.1460)(3,0.0751)(4,0.0767)(5,0.0914)(6,0.1117)(7,0.1273)(8,0.1681)(9,0.2435)(10,0.3205)(11,0.4034)(12,0.4898)(13,0.5333)(14,0.5772)(15,0.6021)(16,0.6438)(17,0.6376)(18,0.6376)(19,0.6495)(20,0.6483)(21,0.6239)(22,0.6201)(23,0.6211)};
\addlegendentry{\HRL{}--\HRL{} (calibrated)}
\addplot[color=tLRL, mark=square*, mark size=1pt, thick] coordinates {(0,0.1837)(1,0.1907)(2,0.1894)(3,0.1661)(4,0.1571)(5,0.1645)(6,0.1650)(7,0.1678)(8,0.1798)(9,0.1982)(10,0.2212)(11,0.2550)(12,0.2993)(13,0.3175)(14,0.3493)(15,0.3573)(16,0.3870)(17,0.3896)(18,0.3942)(19,0.3998)(20,0.4185)(21,0.4055)(22,0.4254)(23,0.4583)};
\addlegendentry{\HRL{}--\LRL{} (frozen)}
\addplot[color=tLRL, mark=square*, mark size=1pt, thick, dashed] coordinates {(0,0.1466)(1,0.1233)(2,0.1266)(3,0.0860)(4,0.0880)(5,0.1010)(6,0.1161)(7,0.1316)(8,0.1668)(9,0.2211)(10,0.2812)(11,0.3544)(12,0.4360)(13,0.4764)(14,0.5148)(15,0.5369)(16,0.5809)(17,0.5745)(18,0.5734)(19,0.5855)(20,0.5840)(21,0.5640)(22,0.5621)(23,0.5560)};
\addlegendentry{\HRL{}--\LRL{} (calibrated)}
\end{axis}\end{tikzpicture}
\end{minipage}\hfill
\begin{minipage}{0.49\textwidth}\centering
\begin{tikzpicture}
\begin{axis}[
  width=\textwidth, height=5.4cm,
  xlabel={Text encoder layer $\ell$}, ylabel={Cross-lingual CKA},
  xmin=-0.5, xmax=27.5, ymin=0.03, ymax=0.60,
  xtick={0,4,8,12,16,20,24,26}, ytick={0.1,0.2,0.3,0.4,0.5},
  axis background/.style={fill=none},
  axis on top=false,
  tick label style={font=\scriptsize}, label style={font=\footnotesize},
  title style={font=\footnotesize}, title={(b) \SLtwo{} SO400M (27 text blocks)},
  grid=both, grid style={gray!18, very thin},
]
\fill[gray!22] (axis cs:-0.5,0.03) rectangle (axis cs:2.5,0.60);
\node[anchor=north west, font=\scriptsize, text=gray!90!black] at (axis cs:-0.2,0.59) {trained ($\ell{<}M$)};
\addplot[color=tHRL, mark=*, mark size=1pt, thick] coordinates {(0,0.1121)(1,0.1094)(2,0.1029)(3,0.1034)(4,0.1014)(5,0.1137)(6,0.1290)(7,0.1468)(8,0.1499)(9,0.1507)(10,0.1790)(11,0.1797)(12,0.1621)(13,0.1901)(14,0.1820)(15,0.1964)(16,0.1844)(17,0.1794)(18,0.1826)(19,0.2061)(20,0.2203)(21,0.2306)(22,0.2920)(23,0.3780)(24,0.4097)(25,0.4156)(26,0.3882)};
\addplot[color=tHRL, mark=*, mark size=1pt, thick, dashed] coordinates {(0,0.1071)(1,0.0807)(2,0.0503)(3,0.0512)(4,0.0513)(5,0.0552)(6,0.0604)(7,0.0715)(8,0.0725)(9,0.0743)(10,0.0864)(11,0.0882)(12,0.0854)(13,0.1206)(14,0.1266)(15,0.1432)(16,0.1502)(17,0.1546)(18,0.1882)(19,0.2366)(20,0.2660)(21,0.3018)(22,0.3665)(23,0.4679)(24,0.5062)(25,0.5433)(26,0.5232)};
\addplot[color=tLRL, mark=square*, mark size=1pt, thick] coordinates {(0,0.0866)(1,0.0878)(2,0.0862)(3,0.0874)(4,0.0865)(5,0.0936)(6,0.0989)(7,0.1089)(8,0.1088)(9,0.1110)(10,0.1220)(11,0.1203)(12,0.1129)(13,0.1319)(14,0.1282)(15,0.1329)(16,0.1294)(17,0.1285)(18,0.1357)(19,0.1552)(20,0.1659)(21,0.1751)(22,0.2169)(23,0.2758)(24,0.3031)(25,0.3158)(26,0.3022)};
\addplot[color=tLRL, mark=square*, mark size=1pt, thick, dashed] coordinates {(0,0.0913)(1,0.0788)(2,0.0514)(3,0.0523)(4,0.0526)(5,0.0561)(6,0.0607)(7,0.0699)(8,0.0710)(9,0.0731)(10,0.0819)(11,0.0834)(12,0.0818)(13,0.1085)(14,0.1136)(15,0.1253)(16,0.1305)(17,0.1323)(18,0.1566)(19,0.1937)(20,0.2162)(21,0.2455)(22,0.2998)(23,0.3930)(24,0.4326)(25,0.4663)(26,0.4512)};
\end{axis}\end{tikzpicture}
\end{minipage}

\vspace{2pt}
\centerline{\ref{cka-shared-legend}}

\caption{Per-layer cross-lingual CKA on $1{,}000$ image-aligned XM3600 captions and $11$ trained languages. Solid: frozen; dashed: trunk-calibrated. The shaded region marks the trunk-trained front blocks ($\ell < M$; $M{=}4$ on \MCtwo{}, $M{=}3$ on \SLtwo{}). Curves show within-pool \HRL{}--\HRL{} and cross-pool \HRL{}--\LRL{} averages. In the frozen state, \HRL{}--\LRL{} lies below \HRL{}--\HRL{}; after calibration, \HRL{}--\LRL{} rises at the projector input on both encoders (numbers in §\ref{sec:analysis} body).}
\label{fig:cka-frozen}
\end{figure*}
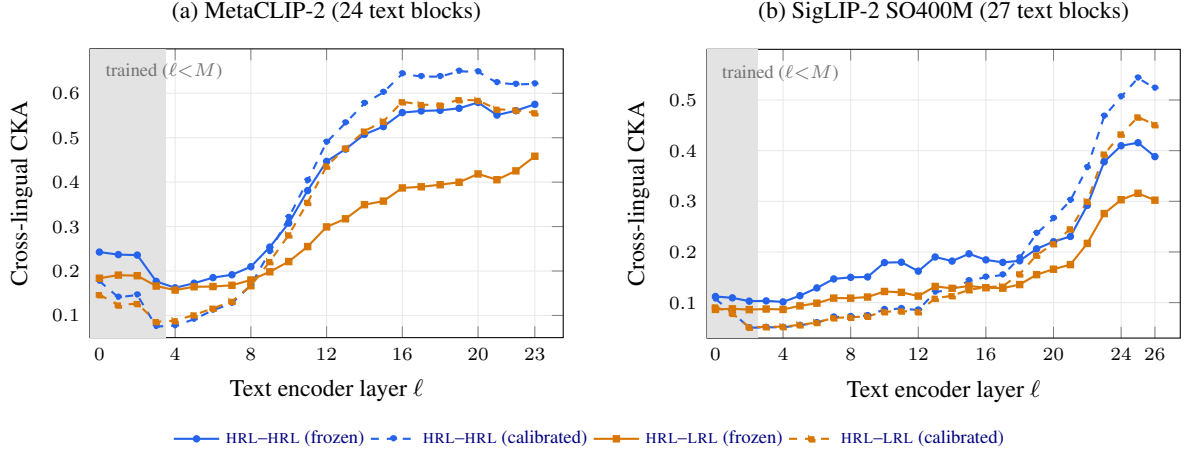

\paragraph{Frozen geometry.} In the frozen state, both within-\HRL{} and \HRL{}--\LRL{} CKA generally rise with depth, but the within-\HRL{} curve remains higher. At the last text block, within-\HRL{}/\HRL{}--\LRL{} CKA is $0.575$/$0.458$ on \MCtwo{} and $0.388$/$0.302$ on \SLtwo{}; near the projector input, the within-\HRL{} curve is consistently $\approx 0.10$--$0.12$ above the \HRL{}--\LRL{} curve. This mirrors the retrieval tier gap: at the last block, an \LRL{} caption remains farther from its \HRL{} parallel than two \HRL{} parallels are from each other.

\paragraph{What trunk calibration changes.} Trunk calibration raises \HRL{}--\LRL{} CKA at the projector input and narrows the gap to the within-\HRL{} curve. On \MCtwo{}, \HRL{}--\LRL{} CKA increases from $0.458$ to $0.556$ at $\ell{=}23$ ($+0.10$); on \SLtwo{}, it increases from $0.302$ to $0.451$ at $\ell{=}26$ ($+0.15$). Although only the first $M$ blocks are trained, the lift appears at the final block, supporting the trunk mechanism: reshaping the front-half can move depth-$M$ states into a region that the frozen back-half maps to higher cross-lingual similarity. This is consistent with the trajectory account (Section~\ref{sec:causal}): shifting the depth-$M$ representation can be sufficient when the frozen back-half remains competent on the shifted region.

\paragraph{How calibration reaches the later blocks.} Only the first $M$ blocks are trained, so the effect on the $N-M$ blocks that follow is what the trunk has to explain. Four measurements converge: cross-lingual CKA rises at the projector input (above), the depth-$M$ residual $r(L)$ and the pooled-row envelope of Equation~\ref{eq:cal-bound} partly account for it (Appendix~\ref{sec:app-lipschitz}), the frozen back-half responds less to pooled-row change when calibrated, and the divergence $d_\ell(L)$ of Equation~\ref{eq:dL-def} falls at every depth with the tier gap shrinking from $0.119$ to $0.022$ on \MCtwo{} and $0.143$ to $0.045$ on \SLtwo{}. Appendix~\ref{sec:app-traj-div} collects these with a frozen-versus-calibrated $d_\ell(L)$ figure, and states the reference points: calibration partially recovers the patched failure mode on general input, while the deep patch and a same-depth multi-\HRL{} centroid are stronger.

\paragraph{Divergence and pretraining prevalence.} The divergence this section tracks is also the quantity that lines up with pretraining data volume: across the ten non-English trained languages, the WebLI-100B prevalence proxy correlates negatively with the frozen depth-$M$ residual (Spearman $-0.88$ on \MCtwo{}, $-0.64$ on \SLtwo{}; Appendix~\ref{sec:app-prevalence}). We read this as corroborating context rather than a second finding: $n{=}10$, and prevalence co-varies with corpus quality and tokeniser coverage, so it cannot separate those accounts (Section~\ref{sec:discussion}).

\paragraph{Scope and caveats.} CKA measures cross-lingual \emph{textual} similarity at each layer, not image--text alignment directly. We therefore use it as the geometric quantity through which the trajectory account predicts alignment recovery, not as a standalone alignment metric. Top-$K$ Jaccard and modality-gap scale on the same image-aligned XM3600 captions are reported in Appendix~\ref{sec:app-extra}.
\section{Discussion}
\label{sec:discussion}

\paragraph{Why two interventions, not one.} Patching (Section~\ref{sec:patching}) provides the cleanest causal test---single-row replacement yields full recovery and noise destroys it---but requires a parallel English caption for each query; calibration (Section~\ref{sec:approx}) needs none, and is not a parameterised form of it: at the calibration depth a single-row patch moves \LRL{} by only $+1.4$ / $-0.1$ pp, against calibration's $+9.6$ / $+17.1$ (Appendix~\ref{sec:app-r1-patch-at-M}). The trained pool already makes the asymmetry concrete: at $\ell{=}N{-}4$ the patch lifts \MCtwo{} \LRL{} to $73.4\%$ where the trunk reaches $53.7\%$, because patching is handed parallel content at inference while the trunk has to generalise from translation supervision, and Equation~\ref{eq:cal-bound} covers only its pooled-row component (Appendix~\ref{sec:app-lipschitz}).


\paragraph{What the linear bias is.} A linear language-identifying direction is learned ($>99\%$ classifier accuracy), yet erasing it leaves \LRL{} retrieval within $\pm 1.5$\,pp up to rank 128, so it is causally dissociable from retrieval-relevant structure. The alignment-causal factor is better described as the non-linear, depth-dependent EOS hidden-state trajectory: bias and alignment coexist linearly, and the failure is structural rather than displacement.

\paragraph{Trajectory divergence as the back-half's distribution boundary.} The trajectory and back-half-distribution framings describe the same observation: per-language paths leave the \HRL{} hidden-state region on which the back-half is competent. This explains both interventions: an \HRL{}-distribution hidden state at a deep block restores back-half competence under patching, while the trunk pulls front-half states back toward that region. We use the trajectory framing because the divergence is per-language and forward-path-localised; the difference is mainly narrative emphasis.

\paragraph{Alternative upstream causes.} Our interventions localise where the failure is expressed in a trained encoder, not what produced it, and three candidates remain live. \textbf{Tokenisation:} the token-count premium over English\footnote{Measured on FLORES-200 devtest ($1{,}012$ parallel sentences per language) as the total token count for language $L$ divided by the total for English, under each encoder's own tokeniser.} is $1.19$--$1.44$ for the trained \LRL{}s on \MCtwo{} but $1.59$--$2.83$ on \SLtwo{}, so \LRL{} captions consume more of a fixed context; yet German is \SLtwo{}'s weakest \HRL{} at a premium of $1.31$, and calibration lifts it $+10.1$\,pp without touching the tokeniser. \textbf{Attention regime:} a re-initialised front block is partly absorbed by causal \MCtwo{} but collapses bidirectional \SLtwo{} at every depth (Appendix~\ref{sec:app-extra-hardinit}), so the regime governs how far a front-half edit propagates---though the tier ordering reproduces across every pooling regime we test (Appendix~\ref{sec:app-morevlm}). \textbf{Optimisation dynamics and data volume:} pretraining prevalence correlates negatively with front-half divergence (Section~\ref{sec:analysis}), consistent with sparse \LRL{} gradient signal leaving the forward path uncorrected, but prevalence co-varies with corpus quality and tokeniser coverage across our pool. Separating these accounts requires controlled pretraining runs, which is out of scope here: the claim of this paper is the interventional localisation, not the explanation.

\paragraph{Diagnostic protocol for new multilingual vision-language encoder releases.} These measurements form a retraining-free checklist for new multilingual vision-language encoders: per-layer cross-lingual CKA (Section~\ref{sec:analysis}) diagnoses the forward-path trajectory; LEACE on the output text embedding (Section~\ref{sec:orth}) tests the linear-bias-displacement hypothesis; EOS-patch sweeps at $\ell \in \{N{-}4, N{-}1\}$ with parallel-EN / random-EN / mid-token controls (Section~\ref{sec:patching}) separate substantive rescue from pooled-position tautology; and per-language back-half sensitivity (Appendix~\ref{sec:app-lipschitz}) characterises its local pooled-row response. We validate the checklist on \MCtwo{} and \SLtwo{} and replicate it on AltCLIP (XLM-R-Large CLS, 24L), NLLB-CLIP-L (M2M100 language-code, 24L), and mSigLIP (SigLIP-base sticky-EOS, 12L), which span different pooling regimes and text-tower scales; Appendix~\ref{sec:app-morevlm}. Extension to generative-VLM text encoders such as BLIP-2 and EVA-CLIP-multilingual remains future work.
\section{Conclusion}
\label{sec:conclusion}

Linearly extractable language identity at the output and the alignment-causal mechanism are different phenomena. Concept erasure removes the identity direction with little change to retrieval; the causal factor appears to be the per-language forward-path trajectory of the pooled hidden state. A single-row substitution from a parallel-content anchor three blocks before the projector recovers \LRL{} retrieval on causal and bidirectional encoders, while a front-layer trunk corroborates the diagnosis and reshapes geometry as predicted.

For multilingual dual-encoder design, this suggests intervening on the trajectory rather than the output embedding: per-language drift is expressed at the depth-$M$ EOS state, making the front blocks a correction site while the back-half remains frozen. The result gives a forward-path-trajectory account of cross-lingual divergence across tested variants---causal, bidirectional, CLS-pooled, and language-code-pooled---and we will release the EOS-swap protocol, per-language sensitivity probe, and trunk pipeline as benchmark-agnostic diagnostics.

Two steps remain: identifying when the trajectory diverges during pretraining and which optimisation signals shape it, and testing whether the structure appears in text-only multilingual encoders. The take-away is that a linearly visible feature is not, by itself, evidence of a causal one.
\section*{Limitations}
\label{sec:limit}

\noindent \textbf{Model coverage:} two pretrained text encoders in the main paper; appendix extensions to AltCLIP, NLLB-CLIP-L, and mSigLIP (Appendix~\ref{sec:app-morevlm}) span four architectures and three text-encoder families, but generalisation to BLIP-2, EVA-CLIP-multilingual, etc.\ is an open empirical question.

\noindent \textbf{Vision tower:} treated as a fixed reference; vision-side bias contribution to the modality-gap denominator is not addressed.

\noindent \textbf{Token-length truncation:} \SLtwo{}'s 64-token cap can truncate long XM3600 captions in non-Latin-script \LRL{}s, but measured per-language truncation rates on the $1{,}000$-image retrieval subset stay $\leq 0.3\%$ in every (model, language) cell, so the probes of Sections~\ref{sec:analysis} and~\ref{sec:lipschitz} are uncontaminated.

\noindent \textbf{Low absolute cosine:} retrieval is driven by relative ordering, not alignment magnitude; trunk-calibration/patching deltas should be read accordingly.

\noindent \textbf{Trunk calibration narrows but does not close the gap:} \HRL{}--\LRL{} CKA rises and the tier gap shrinks (Sections~\ref{sec:analysis},~\ref{sec:results}), but \LRL{} retrieval does not reach \HRL{} parity. Two plausible contributors: (i)~the objective targets centroid-aligned projector-space representations but does not fully equalise depth-$M$ distributions across languages; (ii)~training data is a 1M-caption CC12M subset translated into the 11-language pool, under-representing genuine \LRL{} surface forms relative to natively-multilingual corpora. Richer objectives, longer training, and natively-multilingual data are natural next steps.

\noindent \textbf{Meaning of \emph{low-resource}:} we use the term operationally, for low prevalence in web-scale vision--language pretraining data, rather than linguistic resource scarcity in general.

\noindent \textbf{Locus, not sole cause:} our intervention identifies a correctable locus on the text-encoder path; the frozen projection head and vision tower may also contribute to the observed gap.

\noindent \textbf{Supervision requirement:} trunk calibration needs adequate parallel-text supervision and is demonstrated on five languages passing our MT-quality screen, so languages below that threshold may exhibit additional failure modes, and generalisation beyond the trained \LRL{}-5 remains open.

\noindent \textbf{Scope of the Lipschitz bridge:} Equation~\ref{eq:cal-bound} constrains only the pooled-row component of the calibrated cross-lingual discrepancy, under an assumed conditional Lipschitz constant; trunk calibration also moves the non-pooled rows at depth $M$, and we do not measure that remainder.

\noindent \textbf{Sensitivity estimates are diagnostic:} the ratios of Appendix~\ref{sec:app-lipschitz} are finite-difference quotients at sampled points, so neither their mean nor their maximum certifies a Lipschitz constant --- the sampled maximum is a lower bound on the supremum, and the only upper bound we compute, the block-wise spectral product, is too loose to be informative.

\noindent \textbf{Beyond-linear erasure:} the falsification covers linear erasure up to rank 128; an adversarial-MLP eraser does not achieve comparable non-linear erasure (Appendix~\ref{sec:app-r1-nonlinear}), so whether a non-linear language encoding is alignment-causal remains open.

\noindent \textbf{Run-to-run variance:} we do not report seed variance for the calibration runs; the bootstrap intervals we do report resample evaluation rows rather than training runs, and are taken at each cell's best patch layer rather than at the $\ell{=}N{-}4$ anchor of Table~\ref{tab:patch}. Accordingly, these intervals quantify row-sampling uncertainty conditional on a trained checkpoint and selected layer, rather than variability of the complete training-and-selection pipeline.

\noindent \textbf{Shared evaluation subset:} the patching, CKA, residual and sensitivity measurements use the same $1{,}000$ image--caption rows of XM3600, so they are not statistically independent. Agreement among them is therefore convergent evidence within one sample, not independent replication. The cross-benchmark suite provides independently sampled retrieval evidence, but does not repeat the mechanistic probes or test whether their relationships reproduce across datasets.


\section*{Acknowledgements}

This research was supported by the `Advanced GPU Utilization Support Program'
funded by the Government of the Republic of Korea (Ministry of Science and ICT).

\paragraph{Author Contributions.}
Donghoon Han led methodology, implementation, experiments, analysis, visualization, and the initial draft. SungHyun Moon led conceptualization, methodological direction, research framing, and substantive revision of the manuscript and rebuttals. Together, they handled the rebuttal and author–reviewer discussion, the resubmission, and the camera-ready manuscript, and jointly shaped the study from conception to completion. Aidyn Zhakatayev and Junghun Cha contributed to discussions of the core idea and experimental validation. SeungJae Lee reviewed the final manuscript.

\clearpage
\bibliography{custom}

@inproceedings{jain2021mural,
  title={MURAL: Multimodal, multitask representations across languages},
  author={Jain, Aashi and Guo, Mandy and Srinivasan, Krishna and Chen, Ting and Kudugunta, Sneha and Jia, Chao and Yang, Yinfei and Baldridge, Jason},
  booktitle={Findings of the Association for computational Linguistics: EMNLP 2021},
  pages={3449--3463},
  year={2021}
}

@inproceedings{conneau2020unsupervised,
  title={Unsupervised cross-lingual representation learning at scale},
  author={Conneau, Alexis and Khandelwal, Kartikay and Goyal, Naman and Chaudhary, Vishrav and Wenzek, Guillaume and Guzm{\'a}n, Francisco and Grave, Edouard and Ott, Myle and Zettlemoyer, Luke and Stoyanov, Veselin},
  booktitle={Proceedings of the 58th annual meeting of the association for computational linguistics},
  pages={8440--8451},
  year={2020}
}

@inproceedings{xue2021mt5,
  title={mT5: A massively multilingual pre-trained text-to-text transformer},
  author={Xue, Linting and Constant, Noah and Roberts, Adam and Kale, Mihir and Al-Rfou, Rami and Siddhant, Aditya and Barua, Aditya and Raffel, Colin},
  booktitle={Proceedings of the 2021 conference of the North American chapter of the association for computational linguistics: Human language technologies},
  pages={483--498},
  year={2021}
}

@article{tschannen2025siglip2,
  title={Siglip 2: Multilingual vision-language encoders with improved semantic understanding, localization, and dense features},
  author={Michael Tschannen and Alexey Gritsenko and Xiao Wang and Muhammad Ferjad Naeem and Ibrahim Alabdulmohsin and Nikhil Parthasarathy and Talfan Evans and Lucas Beyer and Ye Xia and Basil Mustafa and Olivier Hénaff and Jeremiah Harmsen and Andreas Steiner and Xiaohua Zhai},
  journal={arXiv preprint arXiv:2502.14786},
  year={2025}
}

@inproceedings{xu2025metaclip2,
 author = {Chuang, Yung-Sung and Li, Yang and Wang, Dong and Yeh, Ching-Feng and Lyu, Kehan and Raghavendra, Ramya and Glass, Jim and Huang, Lifei and Weston, Jason and Zettlemoyer, Luke and Chen, Xinlei and Liu, Zhuang and Xie, Saining and Yih, Scott and Li, Shang-Wen and Xu, Hu},
 booktitle = {Advances in Neural Information Processing Systems},
 doi = {10.52202/085713-1601},
 editor = {D. Belgrave and C. Zhang and H. Lin and R. Pascanu and P. Koniusz and M. Ghassemi and N. Chen},
 pages = {48009--48036},
 publisher = {Curran Associates, Inc.},
 title = {Meta CLIP 2: A Worldwide Scaling Recipe},
 url = {https://proceedings.neurips.cc/paper_files/paper/2025/file/449fb670956a93b3d4d95167f72093e1-Paper-Conference.pdf},
 volume = {38, Main Conference},
 year = {2025}
}

@inproceedings{thapliyal2022xm3600,
  title={Crossmodal-3600: A massively multilingual multimodal evaluation dataset},
  author={Thapliyal, Ashish V and Tuset, Jordi Pont and Chen, Xi and Soricut, Radu},
  booktitle={Proceedings of the 2022 Conference on Empirical Methods in Natural Language Processing},
  pages={715--729},
  year={2022}
}

@article{nllb2022flores200,
  title={No language left behind: Scaling human-centered machine translation},
  author={{NLLB Team} and Marta R. Costa-jussà and James Cross and Onur Çelebi and Maha Elbayad and Kenneth Heafield and Kevin Heffernan and Elahe Kalbassi and Janice Lam and Daniel Licht and Jean Maillard and Anna Sun and Skyler Wang and Guillaume Wenzek and Al Youngblood and Bapi Akula and Loic Barrault and Gabriel Mejia Gonzalez and Prangthip Hansanti and John Hoffman and Semarley Jarrett and Kaushik Ram Sadagopan and Dirk Rowe and Shannon Spruit and Chau Tran and Pierre Andrews and Necip Fazil Ayan and Shruti Bhosale and Sergey Edunov and Angela Fan and Cynthia Gao and Vedanuj Goswami and Francisco Guzmán and Philipp Koehn and Alexandre Mourachko and Christophe Ropers and Safiyyah Saleem and Holger Schwenk and Jeff Wang},
  journal={arXiv preprint arXiv:2207.04672},
  year={2022}
}

@inproceedings{rei2022comet22,
  title={COMET-22: Unbabel-IST 2022 submission for the metrics shared task},
  author={Rei, Ricardo and De Souza, Jos{\'e} GC and Alves, Duarte and Zerva, Chrysoula and Farinha, Ana C and Glushkova, Taisiya and Lavie, Alon and Coheur, Luisa and Martins, Andr{\'e} FT},
  booktitle={Proceedings of the Seventh Conference on Machine Translation (WMT)},
  pages={578--585},
  year={2022}
}

@article{liang2022modalitygap,
  title={Mind the gap: Understanding the modality gap in multi-modal contrastive representation learning},
  author={Liang, Victor Weixin and Zhang, Yuhui and Kwon, Yongchan and Yeung, Serena and Zou, James Y},
  journal={Advances in neural information processing systems},
  volume={35},
  pages={17612--17625},
  year={2022}
}

@inproceedings{kornblith2019cka,
  title={Similarity of neural network representations revisited},
  author={Kornblith, Simon and Norouzi, Mohammad and Lee, Honglak and Hinton, Geoffrey},
  booktitle={International conference on machine learning},
  pages={3519--3529},
  year={2019},
  organization={PMLR}
}

@inproceedings{conneau2020emerging,
  title={Emerging cross-lingual structure in pretrained language models},
  author={Conneau, Alexis and Wu, Shijie and Li, Haoran and Zettlemoyer, Luke and Stoyanov, Veselin},
  booktitle={Proceedings of the 58th annual meeting of the association for computational linguistics},
  pages={6022--6034},
  year={2020}
}

@inproceedings{pires2019mbert,
  title={How multilingual is multilingual BERT?},
  author={Pires, Telmo and Schlinger, Eva and Garrette, Dan},
  booktitle={Proceedings of the 57th annual meeting of the association for computational linguistics},
  pages={4996--5001},
  year={2019}
}

@inproceedings{vig2020causal,
author = {Vig, Jesse and Gehrmann, Sebastian and Belinkov, Yonatan and Qian, Sharon and Nevo, Daniel and Singer, Yaron and Shieber, Stuart},
title = {Investigating gender bias in language models using causal mediation analysis},
year = {2020},
isbn = {9781713829546},
publisher = {Curran Associates Inc.},
address = {Red Hook, NY, USA},
booktitle = {Proceedings of the 34th International Conference on Neural Information Processing Systems},
articleno = {1039},
numpages = {14},
location = {Vancouver, BC, Canada},
series = {NIPS '20}
}

@article{meng2022rome,
  title={Locating and editing factual associations in gpt},
  author={Meng, Kevin and Bau, David and Andonian, Alex and Belinkov, Yonatan},
  journal={Advances in neural information processing systems},
  volume={35},
  pages={17359--17372},
  year={2022}
}

@inproceedings{wang2022ioi,
  title={Interpretability in the Wild: a Circuit for Indirect Object Identification in {GPT}-2 Small},
  author={Kevin Ro Wang and Alexandre Variengien and Arthur Conmy and Buck Shlegeris and Jacob Steinhardt},
  booktitle={The Eleventh International Conference on Learning Representations },
  year={2023},
}

@article{belrose2023leace,
  title={Leace: Perfect linear concept erasure in closed form},
  author={Belrose, Nora and Schneider-Joseph, David and Ravfogel, Shauli and Cotterell, Ryan and Raff, Edward and Biderman, Stella},
  journal={Advances in Neural Information Processing Systems},
  volume={36},
  pages={66044--66063},
  year={2023}
}

@inproceedings{ravfogel2020inlp,
  title={Null it out: Guarding protected attributes by iterative nullspace projection},
  author={Ravfogel, Shauli and Elazar, Yanai and Gonen, Hila and Twiton, Michael and Goldberg, Yoav},
  booktitle={Proceedings of the 58th annual meeting of the association for computational linguistics},
  pages={7237--7256},
  year={2020}
}

@inproceedings{geva2020feedforward,
  title={Transformer feed-forward layers are key-value memories},
  author={Geva, Mor and Schuster, Roei and Berant, Jonathan and Levy, Omer},
  booktitle={Proceedings of the 2021 conference on empirical methods in natural language processing},
  pages={5484--5495},
  year={2021}
}

@article{aggarwal2020xtd,
  title={Towards zero-shot cross-lingual image retrieval},
  author={Aggarwal, Pranav and Kale, Ajinkya},
  journal={arXiv preprint arXiv:2012.05107},
  year={2020}
}

@inproceedings{geigle2024babelimagenet,
  title={Babel-imagenet: Massively multilingual evaluation of vision-and-language representations},
  author={Geigle, Gregor and Timofte, Radu and Glava{\v{s}}, Goran},
  booktitle={Proceedings of the 62nd Annual Meeting of the Association for Computational Linguistics (Volume 1: Long Papers)},
  pages={5064--5084},
  year={2024}
}

@inproceedings{romero2024cvqa,
 author = {Romero, David and Lyu, Chenyang and Wibowo, Haryo Akbarianto and Lynn, Teresa and Hamed, Injy and Kishore, Aditya Nanda and Mandal, Aishik and Dragonetti, Alina and Abzaliev, Artem and Tonja, Atnafu Lambebo and Balcha, Bontu Fufa and Whitehouse, Chenxi and Salamea, Christian and Velasco, Dan John and Adelani, David Ifeoluwa and Le Meur, David and Villa-Cueva, Emilio and Koto, Fajri and Farooqui, Fauzan and Belcavello, Frederico and Batnasan, Ganzorig and Vallejo, Gisela and Caulfield, Grainne and Ivetta, Guido and Song, Haiyue and Ademtew, Henok Biadglign and Maina, Hern\'{a}n and Lovenia, Holy and Azime, Israel Abebe and Cruz, Jan Christian Blaise and Gala, Jay and Geng, Jiahui and Ortiz-Barajas, Jesus-German and Baek, Jinheon and Dunstan, Jocelyn and Alemany, Laura Alonso and Nagasinghe, Kumaranage Ravindu Yasas and Benotti, Luciana and D\textquotesingle Haro, Luis Fernando and Viridiano, Marcelo and Estecha-Garitagoitia, Marcos and Cabrera, Maria Camila Buitrago and Rodr\'{\i}guez-Cantelar, Mario and Jouitteau, M\'{e}lanie and Mihaylov, Mihail and Etori, Naome and Imam, Mohamed Fazli Mohamed and Adilazuarda, Muhammad Farid and Gochoo, Munkhjargal and Otgonbold, Munkh-Erdene and Niyomugisha, Olivier and Silva, Paula M\'{o}nica and Chitale, Pranjal and Dabre, Raj and Chevi, Rendi and Zhang, Ruochen and Diandaru, Ryandito and Cahyawijaya, Samuel and G\'{o}ngora, Santiago and Jeong, Soyeong and Purkayastha, Sukannya and Kuribayashi, Tatsuki and Clifford, Teresa and Jayakumar, Thanmay and Torrent, Tiago Timponi and Ehsan, Toqeer and Araujo, Vladimir and Kementchedjhieva, Yova and Burzo, Zara and Lim, Zheng Wei and Yong, Zheng Xin and Ignat, Oana and Nwatu, Joan and Mihalcea, Rada and Solorio, Thamar and Aji, Alham Fikri},
 booktitle = {Advances in Neural Information Processing Systems},
 doi = {10.52202/079017-0366},
 editor = {A. Globerson and L. Mackey and D. Belgrave and A. Fan and U. Paquet and J. Tomczak and C. Zhang},
 pages = {11479--11505},
 publisher = {Curran Associates, Inc.},
 title = {CVQA: Culturally-diverse Multilingual Visual Question Answering Benchmark},
 url = {https://proceedings.neurips.cc/paper_files/paper/2024/file/1568882ba1a50316e87852542523739c-Paper-Datasets_and_Benchmarks_Track.pdf},
 volume = {37},
 year = {2024}
}

@inproceedings{reimers2020multilingual,
  title={Making monolingual sentence embeddings multilingual using knowledge distillation},
  author={Reimers, Nils and Gurevych, Iryna},
  booktitle={Proceedings of the 2020 conference on empirical methods in natural language processing (EMNLP)},
  pages={4512--4525},
  year={2020}
}

@inproceedings{coates2018meta,
  title={Frustratingly easy meta-embedding--computing meta-embeddings by averaging source word embeddings},
  author={Coates, Joshua and Bollegala, Danushka},
  booktitle={Proceedings of the 2018 Conference of the North American Chapter of the Association for Computational Linguistics: Human Language Technologies, Volume 2 (Short Papers)},
  pages={194--198},
  year={2018}
}

@InProceedings{jha2025platonic,
  title = 	 {Position: The Platonic Representation Hypothesis},
  author =       {Huh, Minyoung and Cheung, Brian and Wang, Tongzhou and Isola, Phillip},
  booktitle = 	 {Proceedings of the 41st International Conference on Machine Learning},
  pages = 	 {20617--20642},
  year = 	 {2024},
  editor = 	 {Salakhutdinov, Ruslan and Kolter, Zico and Heller, Katherine and Weller, Adrian and Oliver, Nuria and Scarlett, Jonathan and Berkenkamp, Felix},
  volume = 	 {235},
  series = 	 {Proceedings of Machine Learning Research},
  month = 	 {21--27 Jul},
  publisher =    {PMLR},
  url = 	 {https://proceedings.mlr.press/v235/huh24a.html}
}

@inproceedings{lopo2025langsurgery,
  title={Language surgery in multilingual large language models},
  author={Lopo, Joanito Agili and Habibi, Muhammad Ravi Shulthan and Wong, Tack Hwa and Ghozali, Muhammad Ilham and Koto, Fajri and Winata, Genta Indra and Limkonchotiwat, Peerat and Aji, Alham Fikri and Cahyawijaya, Samuel},
  booktitle={Proceedings of the 5th Workshop on Multilingual Representation Learning (MRL 2025)},
  pages={438--467},
  year={2025}
}

@article{mu2017abtt,
  title={All-but-the-top: Simple and effective postprocessing for word representations},
  author={Mu, Jiaqi and Bhat, Suma and Viswanath, Pramod},
  journal={arXiv preprint arXiv:1702.01417},
  year={2017}
}

@inproceedings{chang2022multigeom,
  title={The geometry of multilingual language model representations},
  author={Chang, Tyler A and Tu, Zhuowen and Bergen, Benjamin K},
  booktitle={Proceedings of the 2022 Conference on Empirical Methods in Natural Language Processing},
  pages={119--136},
  year={2022}
}

@inproceedings{changpinyo2021cc12m,
  title={Conceptual 12m: Pushing web-scale image-text pre-training to recognize long-tail visual concepts},
  author={Changpinyo, Soravit and Sharma, Piyush and Ding, Nan and Soricut, Radu},
  booktitle={2021 IEEE/CVF Conference on Computer Vision and Pattern Recognition (CVPR)},
  pages={3557--3567},
  year={2021},
  organization={IEEE}
}

@article{glm2026glm5,
  title={Glm-5: from vibe coding to agentic engineering},
  author={{GLM-5 Team} and Aohan Zeng and Xin Lv and Zhenyu Hou and Zhengxiao Du and Qinkai Zheng and Bin Chen and Da Yin and Chendi Ge and Chenghua Huang and Chengxing Xie and Chenzheng Zhu and Congfeng Yin and Cunxiang Wang and Gengzheng Pan and Hao Zeng and Haoke Zhang and Haoran Wang and Huilong Chen and Jiajie Zhang and Jian Jiao and Jiaqi Guo and Jingsen Wang and Jingzhao Du and Jinzhu Wu and Kedong Wang and Lei Li and Lin Fan and Lucen Zhong and Mingdao Liu and Mingming Zhao and Pengfan Du and Qian Dong and Rui Lu and Shuang-Li and Shulin Cao and Song Liu and Ting Jiang and Xiaodong Chen and Xiaohan Zhang and Xuancheng Huang and Xuezhen Dong and Yabo Xu and Yao Wei and Yifan An and Yilin Niu and Yitong Zhu and Yuanhao Wen and Yukuo Cen and Yushi Bai and Zhongpei Qiao and Zihan Wang and Zikang Wang and Zilin Zhu and Ziqiang Liu and Zixuan Li and Bojie Wang and Bosi Wen and Can Huang and Changpeng Cai and Chao Yu and Chen Li and Chengwei Hu and Chenhui Zhang and Dan Zhang and Daoyan Lin and Dayong Yang and Di Wang and Ding Ai and Erle Zhu and Fangzhou Yi and Feiyu Chen and Guohong Wen and Hailong Sun and Haisha Zhao and Haiyi Hu and Hanchen Zhang and Hanrui Liu and Hanyu Zhang and Hao Peng and Hao Tai and Haobo Zhang and He Liu and Hongwei Wang and Hongxi Yan and Hongyu Ge and Huan Liu and Huanpeng Chu and Jia'ni Zhao and Jiachen Wang and Jiajing Zhao and Jiamin Ren and Jiapeng Wang and Jiaxin Zhang and Jiayi Gui and Jiayue Zhao and Jijie Li and Jing An and Jing Li and Jingwei Yuan and Jinhua Du and Jinxin Liu and Junkai Zhi and Junwen Duan and Kaiyue Zhou and Kangjian Wei and Ke Wang and Keyun Luo and Laiqiang Zhang and Leigang Sha and Liang Xu and Lindong Wu and Lintao Ding and Lu Chen and Minghao Li and Nianyi Lin and Pan Ta and Qiang Zou and Rongjun Song and Ruiqi Yang and Shangqing Tu and Shangtong Yang and Shaoxiang Wu and Shengyan Zhang and Shijie Li and Shuang Li and Shuyi Fan and Wei Qin and Wei Tian and Weining Zhang and Wenbo Yu and Wenjie Liang and Xiang Kuang and Xiangmeng Cheng and Xiangyang Li and Xiaoquan Yan and Xiaowei Hu and Xiaoying Ling and Xing Fan and Xingye Xia and Xinyuan Zhang and Xinze Zhang and Xirui Pan and Xu Zou and Xunkai Zhang and Yadi Liu and Yandong Wu and Yanfu Li and Yidong Wang and Yifan Zhu and Yijun Tan and Yilin Zhou and Yiming Pan and Ying Zhang and Yinpei Su and Yipeng Geng and Yong Yan and Yonglin Tan and Yuean Bi and Yuhan Shen and Yuhao Yang and Yujiang Li and Yunan Liu and Yunqing Wang and Yuntao Li and Yurong Wu and Yutao Zhang and Yuxi Duan and Yuxuan Zhang and Zezhen Liu and Zhengtao Jiang and Zhenhe Yan and Zheyu Zhang and Zhixiang Wei and Zhuo Chen and Zhuoer Feng and Zijun Yao and Ziwei Chai and Ziyuan Wang and Zuzhou Zhang and Bin Xu and Minlie Huang and Hongning Wang and Juanzi Li and Yuxiao Dong and Jie Tang},
  journal={arXiv preprint arXiv:2602.15763},
  year={2026}
}

@inproceedings{chen2023altclip,
  title={Altclip: Altering the language encoder in clip for extended language capabilities},
  author={Chen, Zhongzhi and Liu, Guang and Zhang, Bo-Wen and Yang, Qinghong and Wu, Ledell},
  booktitle={Findings of the Association for Computational Linguistics: ACL 2023},
  pages={8666--8682},
  year={2023}
}

@article{visheratin2024nllbclip,
  title={NLLB-CLIP--train performant multilingual image retrieval model on a budget},
  author={Visheratin, Alexander},
  journal={arXiv preprint arXiv:2309.01859},
  year={2023}
}

@InProceedings{wang2025scaling,
    author    = {Wang, Xiao and Alabdulmohsin, Ibrahim and Salz, Daniel and Li, Zhe and Rong, Keran and Zhai, Xiaohua},
    title     = {Scaling Pre-training to One Hundred Billion Data for Vision Language Models},
    booktitle = {Proceedings of the IEEE/CVF Conference on Computer Vision and Pattern Recognition (CVPR) Findings},
    month     = {June},
    year      = {2026},
    pages     = {6185-6196}
}

\clearpage
\appendix
\makeatletter
\let\acl@oldseccntformat\@seccntformat
\renewcommand{\@seccntformat}[1]{%
  \ifnum\pdfstrcmp{#1}{section}=0 Appendix \csname the#1\endcsname:\ %
  \else \acl@oldseccntformat{#1}\fi
}
\makeatother
\section*{Appendix Roadmap}

The appendices are grouped into four parts. (i)~\textbf{Evidence for main-text claims}: per-language patching (Appendix~\ref{sec:app-perlang}), pseudocode (Appendix~\ref{sec:app-pseudo}), and the trunk-training sweep selecting the canonical configuration (Appendix~\ref{sec:app-sweep}). (ii)~\textbf{Broader evaluation}: five-benchmark tier aggregates (Appendix~\ref{sec:app-benchmarks}) and per-language results (Appendix~\ref{sec:app-perlang-benchmarks}). (iii)~\textbf{Robustness probes}: translation quality, intra-language clustering (Appendix~\ref{sec:app-flores-correlation}), and further ablations (Appendix~\ref{sec:app-extra}). (iv)~\textbf{Supporting measurements}: sampled back-half sensitivity for Equation~\ref{eq:cal-bound} (Appendix~\ref{sec:app-lipschitz}) and additional patching controls (Appendix~\ref{sec:app-r1-controls}).

\section{Per-Language Patching Sweep, All Cells}
\label{sec:app-perlang}

Table~\ref{tab:per-language-patch} complements the $\ell=N{-}4$ tier averages of Table~\ref{tab:patch} across all four (model, state) cells on XM3600 (1{,}000 images), reporting the no-patch baseline \Rat{}, the \emph{best-layer} patched peak \Rat{} with its layer index, and delta; peaks at $\ell = N{-}1$ are pooling references rather than non-tautological rescues. The single-source English reference (``EN ref.'') is the direct English-caption \Rat{} on the same image set; rows matching it recover to that reference under EOS swap.

\begin{table*}[h]
\centering
\small
\setlength{\tabcolsep}{4pt}
\begin{tabular}{ll|ccc|ccc|ccc|ccc}
\toprule
& & \multicolumn{3}{c|}{\MCtwo{} frozen} & \multicolumn{3}{c|}{\MCtwo{} cal.} & \multicolumn{3}{c|}{\SLtwo{} frozen} & \multicolumn{3}{c}{\SLtwo{} cal.} \\
& & \multicolumn{3}{c|}{(EN ref. 69.4)} & \multicolumn{3}{c|}{(EN ref. 72.0)} & \multicolumn{3}{c|}{(EN ref. 66.1)} & \multicolumn{3}{c}{(EN ref. 70.5)} \\
Lang & Grp & base & peak\,($\ell$) & $\Delta$ & base & peak\,($\ell$) & $\Delta$ & base & peak\,($\ell$) & $\Delta$ & base & peak\,($\ell$) & $\Delta$ \\
\midrule
fr  & \HRL{} & 84.4 & 88.1\,(20) & $+3.7$  & 86.2 & 88.6\,(14) & $+2.4$  & 78.2 & 85.7\,(22) & $+7.5$  & 82.6 & 87.6\,(22) & $+5.0$ \\
de  & \HRL{} & 86.5 & 88.6\,(20) & $+2.1$  & 86.8 & 89.4\,(20) & $+2.6$  & 33.7 & 76.5\,(24) & $+42.8$ & 43.8 & 80.9\,(24) & $+37.1$ \\
es  & \HRL{} & 78.0 & 83.9\,(20) & $+5.9$  & 77.6 & 83.4\,(20) & $+5.8$  & 64.1 & 81.2\,(24) & $+17.1$ & 67.4 & 82.7\,(24) & $+15.3$ \\
zh  & \HRL{} & 74.6 & 84.6\,(20) & $+10.0$ & 77.5 & 87.3\,(20) & $+9.8$  & 63.8 & 82.3\,(24) & $+18.5$ & 70.5 & 83.7\,(24) & $+13.2$ \\
ko  & \HRL{} & 75.5 & 83.9\,(20) & $+8.4$  & 78.9 & 86.9\,(20) & $+8.0$  & 74.0 & 85.4\,(24) & $+11.4$ & 77.0 & 86.7\,(24) & $+9.7$ \\
\midrule
bn  & \LRL{} & 64.7 & 79.3\,(20) & $+14.6$ & 69.0 & 82.2\,(20) & $+13.2$ & 30.2 & 74.1\,(24) & $+43.9$ & 42.7 & 77.7\,(24) & $+35.0$ \\
fil & \LRL{} & 39.9 & 73.5\,(20) & $+33.6$ & 52.7 & 76.8\,(20) & $+24.1$ & 26.9 & 70.7\,(24) & $+43.8$ & 45.0 & 77.5\,(24) & $+32.5$ \\
hi  & \LRL{} & 49.4 & 73.9\,(22) & $+24.5$ & 52.9 & 76.9\,(20) & $+24.0$ & 33.6 & 73.7\,(24) & $+40.1$ & 38.5 & 74.9\,(24) & $+36.4$ \\
sw  & \LRL{} & 22.1 & 71.0\,(22) & $+48.9$ & 42.1 & 76.2\,(22) & $+34.1$ & 12.4 & 69.3\,(24) & $\mathbf{+56.9}$ & 35.7 & 75.9\,(24) & $+40.2$ \\
te  & \LRL{} & 44.3 & 73.8\,(22) & $+29.5$ & 52.0 & 77.0\,(22) & $+25.0$ &  5.2 & 66.1\,(26) & $\mathbf{+60.9}$ & 31.9 & 73.5\,(24) & $+41.6$ \\
\midrule
\bottomrule
\end{tabular}
\caption{Per-language image-to-text \Rat{} (\%) on XM3600 (1{,}000 images): no-patch baseline, best-layer patched peak with layer index, and delta (pp), for each (model, state) cell. The single-source English reference (EN ref.) appears above each block; rows matching it recover to that reference under EOS swap. Bold $\Delta$ entries are $\geq +50$ pp.}
\label{tab:per-language-patch}
\end{table*}
\section{Calibration: Detailed Recipe and Pseudo-code}
\label{sec:app-pseudo}

This appendix gives the training recipe and pseudocode for trunk calibration (Section~\ref{sec:approx}). The loss is computed at the encoder's \emph{projected output}, after the frozen back-half and projection head, not at the depth-$M$ hidden state; gradients pass through the frozen back-half into the trainable front blocks, shaping the depth-$M$ distribution via the Lipschitz bridge of Section~\ref{sec:lipschitz}.

\paragraph{Training recipe.} Encoder: \MCtwo{} Worldwide-Huge ($N{=}24$, hidden $1024$) or \SLtwo{} SO400M ($N{=}27$, hidden $1152$). Trainable: first $M$ text-encoder blocks ($M{=}4$ / $3$), initialised from frozen weights (\emph{soft fine-tune}); frozen: back-half blocks $\varphi_M, \ldots, \varphi_{N-1}$, projection head $\pi$, vision tower, and all LayerNorms outside the trainable range. Trainable parameters are $\sim 50$M / $\sim 40$M. Optimiser: AdamW ($\beta_1{=}0.9$, $\beta_2{=}0.999$, weight decay $0.01$), gradient clipping at norm $1.0$, no LR schedule. Learning rate is $1.2{\times}10^{-4}$ on \MCtwo{} and $2.4{\times}10^{-4}$ on \SLtwo{}; $\lambda_{\mathrm{align}}{=}2$ / $1$ are the canonical values from Appendix~\ref{sec:app-sweep}. Training uses $128$ caption groups for $20{,}000$ steps ($\sim 5$ GPU-hours per encoder on one H100), drawn from a 1M-caption CC12M subset whose English captions are translated into the other 10 trained-pool languages by GLM-5.1; images are never used during trunk training. The canonical anchor is the per-row centroid of frozen-encoder projected outputs over the full 11-language trained pool, matching Section~\ref{sec:approx} and Appendix~\ref{sec:app-benchmarks}; \HRL{}-6-only and English-only anchors are ablated in Appendix~\ref{sec:app-sweep}. Captions use the encoder's native tokeniser and sequence cap ($77$ / $64$). Block numbering follows the implementation, with $H_{k+1} = \varphi_k(H_k)$ and $\Esem = \pi(H_N)$ for the hidden-state matrices of Section~\ref{sec:setup}: the trainable blocks are \texttt{text\_model.encoder.layers[0:M]}, matching $\varphi_0, \ldots, \varphi_{M-1}$ of Section~\ref{sec:setup}, and a patch or trajectory read at block $\ell$ hooks \texttt{layers[$\ell$]}'s output. Code and checkpoints will be released at \url{https://github.com/dnotitia/geometric-bottleneck}.

\paragraph{The anchor and the alignment term.} For a parallel group $g$ with captions $\{x_g^L\}_{L\in\mathcal{L}}$, the anchor of Equation~\ref{eq:trunkloss} is
\begin{equation}
  a_g \;=\; \mathrm{norm}\Bigl(\tfrac{1}{|\mathcal{A}|}\textstyle\sum_{L' \in \mathcal{A}} \mathrm{norm}\bigl(\Esem(x_g^{L'})\bigr)\Bigr),
  \label{eq:anchor}
\end{equation}
i.e.\ normalise each frozen projected embedding, average over the anchor set $\mathcal{A}$, then renormalise; $\mathcal{A} = \mathcal{L}$ for the canonical variant. Writing the alignment term out over the batch's groups $\mathcal{G}$,
\begin{equation}
  \mathcal{L}_{\mathrm{align}} \;=\; \frac{1}{|\mathcal{G}||\mathcal{L}|} \sum_{g \in \mathcal{G}} \sum_{L \in \mathcal{L}} \bigl[\,1 - \langle z_g^L,\, a_g \rangle\,\bigr],
  \label{eq:align-full}
\end{equation}
which is the \texttt{L\_align} line of the pseudocode below: a sample-wise cosine loss between each language's calibrated embedding and its own group's anchor, not a distance between per-language centroids. The InfoNCE term uses the same $a_g$ as the positive for all $|\mathcal{L}|$ languages of group $g$, with the other groups' anchors in the batch as negatives.

\paragraph{Pseudo-code.} The text-only forward returns the projected text embedding (\texttt{text\_proj}); for both \MCtwo{} and \SLtwo{}, this is the L2-normalised \texttt{get\_text\_features} output in the shared image--text space.

\begin{tcolorbox}[pseudocodebox={Training: trunk calibration}]
\begin{footnotesize}
\begin{verbatim}
# E_frozen = pi o phi_{N-1} o ... o phi_0
# E_calib  = pi o phi_{N-1} o ... o phi_M
#            o phi'_{M-1} o ... o phi'_0
# ANCHOR_LANGS: all 11 trained languages
#   (ablations: HRL-6 or EN-only)
# TRAIN_LANGS: 11

phi_prime = [copy(phi[l]) for l in range(M)]
freeze(back-half, pi, vision)

for step in 1..T:
  batch = sample_parallel(
    B groups; each group = 1 EN source
    + 10 GLM-5.1 translations)

  # Frozen anchor: centroid of projections
  with no_grad():
    anchors = []
    for L in ANCHOR_LANGS:
      f = E_frozen.get_text_features(
            **tokenize(batch[L]))
      anchors.append(F.normalize(f, dim=-1))
    z_anchor = F.normalize(
      stack(anchors).mean(0), dim=-1)

  # Trainable projected outputs
  caps, gids = [], []
  for gi, group in enumerate(batch):
    for L in TRAIN_LANGS:
      caps.append(group[L]); gids.append(gi)

  z_query = E_calib.get_text_features(
              **tokenize(caps))
  z_query = F.normalize(z_query, dim=-1)

  # Losses at projected output
  logits  = (z_query @ z_anchor.T) / tau
  L_nce   = cross_entropy(logits, gids)
  L_align = (1 - (z_query
              * z_anchor[gids]).sum(-1)).mean()
  loss    = L_nce + lambda_align * L_align

  loss.backward()
  clip_grad(1.0)
  optimizer.step()

save({"trainable_layers_state_dict":
        [phi_prime[l].state_dict()
         for l in range(M)],
      "config": {"M": M,
        "lambda_align": lambda_align,
        "trained_langs": TRAIN_LANGS}})
\end{verbatim}
\end{footnotesize}
\end{tcolorbox}

\begin{tcolorbox}[pseudocodebox={Inference: load the trunk}]
\begin{footnotesize}
\begin{verbatim}
for l in range(M):
  E.text.encoder.layers[l] \
    .load_state_dict(state_dicts[l])
\end{verbatim}
\end{footnotesize}
\end{tcolorbox}

\noindent The two architectures share the training loop and differ only in encoder loading and tokeniser. The temperature $\tau$ is fixed at $0.02$ throughout.
\section{Translation-Quality Sanity Check and Intra-Language Clustering}
\label{sec:app-flores-correlation}

This appendix reports two probes: a COMET-22 sanity check for the translation system used in trunk training (Appendix~\ref{sec:app-comet-intra}), and per-language intra-language cosine structure (Appendix~\ref{sec:app-intra-cos}), which Section~\ref{sec:leace} cites as distinct from the LEACE direction.

\subsection{Translation-Quality Sanity Check}
\label{sec:app-comet-intra}

The trunk is trained on GLM-5.1-translated parallel captions (Section~\ref{sec:approx}). Reference-based COMET-22 on FLORES-200 devtest \citep{rei2022comet22} places GLM-5.1 translations in a narrow adequate-quality range over the 11-language trained pool (\HRL{}-6 + \LRL{}-5): $0.816$ (Hindi) to $0.905$ (Korean), with Hindi only slightly below the nominal $0.82$ cutoff at three-decimal precision. A second reference system, GPT-5.2\footnote{GPT-5.2-2025-12-11.}, gives nearly identical scores ($\pm 0.01$) and puts all trained-pool languages at or above $0.825$. The pool was selected by this LLM-translator quality screen; candidate languages below the bar were excluded from trunk training. We do not claim a correlation between translation quality and downstream metrics, since the trained-pool COMET range is narrow ($\leq 0.09$).

\begin{table}[h]
\centering
\small
\setlength{\tabcolsep}{4pt}
\begin{tabular}{lc|cc}
\toprule
Lang & Tier & glm-5.1 & gpt-5.2 \\
\midrule
fr   & \HRL{} & 0.888 & 0.892 \\
de   & \HRL{} & 0.889 & 0.893 \\
es   & \HRL{} & 0.870 & 0.875 \\
zh   & \HRL{} & 0.895 & 0.896 \\
ko   & \HRL{} & 0.905 & 0.909 \\
\midrule
bn   & \LRL{} & 0.876 & 0.881 \\
fil  & \LRL{} & 0.853 & 0.865 \\
hi   & \LRL{} & 0.816 & 0.825 \\
sw   & \LRL{} & 0.845 & 0.863 \\
te   & \LRL{} & 0.871 & 0.876 \\
\midrule
\bottomrule
\end{tabular}
\caption{COMET-22 reference-based translation quality on FLORES-200 devtest, using GLM-5.1 and GPT-5.2 as translation systems. Scores are near or above the $0.82$ selection threshold for all trained-pool languages; Hindi is the boundary case under GLM-5.1 and clears the threshold under GPT-5.2.}
\label{tab:exp2-perlang}
\end{table}

\subsection{Intra-Language Clustering on \MCtwo{}}
\label{sec:app-intra-cos}

The LEACE result of Section~\ref{sec:leace} rules out the language-identity direction as the alignment-causal factor, but the text embedding still has per-language geometric structure. Define
\begin{equation}
  \alpha(L) \;:=\; \mathbb{E}_{i \neq j}\,\bigl[1 - \cos(t_L(x_i), t_L(x_j))\bigr],
  \label{eq:intracos}
\end{equation}
the mean pairwise cosine distance between distinct same-language captions; smaller $\alpha(L)$ means a narrower same-language cone in $\mathbb{R}^D$. On frozen \MCtwo{} at the final projection layer ($3{,}600$ XM3600 captions per language; $10{,}000$ within-language pairs), $\alpha(L)$ shows a tier pattern: \HRL{} ranges $0.56$--$0.71$ (en $0.58$, fr $0.68$, de $0.71$, es $0.66$, zh $0.56$, ko $0.61$), while trained \LRL{} is often tighter ($0.38$--$0.65$: sw $0.38$, fil $0.43$, te $0.54$, hi $0.62$, bn $0.65$). Weaker retrieval is associated with tighter within-language clustering.

This intra-language collapse is distinct from the linear language-identity direction tested in Section~\ref{sec:leace}: LEACE removes a low-rank between-language separation direction, whereas collapse is a per-language radial structure that compresses content into a narrow cone. Removing the identity direction therefore does not undo the collapse, explaining why LEACE leaves retrieval unchanged (Section~\ref{sec:leace}) while \HRL{}-averaging (Section~\ref{sec:hrl-avg}) lifts it by mixing cross-lingual content. The collapse is a separate symptom of low-resource-language representation, and calibration partially mitigates it by reshaping the front-of-encoder distribution.
\section{Multi-Benchmark Calibration Gains}
\label{sec:app-benchmarks}

Table~\ref{tab:bench} reports calibrated-vs-frozen gains for the canonical trunk (11-language centroid anchor; $M{=}4$ for \MCtwo{} and $M{=}3$ for \SLtwo{}; Section~\ref{sec:approx}). \HRL{}-6 and EN-only anchors are ablated in Table~\ref{tab:anchor-ablation} (Appendix~\ref{sec:app-sweep}). Each cell is the change in mean performance on the benchmark's language pool. Throughout this appendix and Table~\ref{tab:bench-mini}, \HRL{} is the \HRL{}-6 average \emph{including} English and \LRL{} the \LRL{}-5 average, both following Section~\ref{sec:setup}; the patching tables (Table~\ref{tab:patch}, Appendix~\ref{sec:app-r1-controls}) instead average the five non-English \HRL{} languages, because English is the patch source there. We evaluate five multilingual suites: Flickr30k-200, XTD-200 \citep{visheratin2024nllbclip}, XM3600 \citep{thapliyal2022xm3600}, Babel-ImageNet \citep{geigle2024babelimagenet}, and CVQA native-text split \citep{romero2024cvqa}. CVQA's translation-evaluation split has smaller and noisier deltas.

\begin{table}[h]
\centering
\small
\setlength{\tabcolsep}{4pt}
\begin{tabular}{l|cc|cc}
\toprule
& \multicolumn{2}{c|}{\MCtwo{}} & \multicolumn{2}{c}{\SLtwo{}} \\
benchmark & $\Delta$\HRL{} & $\Delta$\LRL{} & $\Delta$\HRL{} & $\Delta$\LRL{} \\
\midrule
Flickr30k-200    & $+0.7$ & $\mathbf{+8.4}$ & $+3.3$ & $\mathbf{+20.4}$ \\
XTD-200          & $+2.2$ & $\mathbf{+9.0}$ & $+3.2$ & $\mathbf{+19.1}$ \\
XM3600 (full)    & $+1.4$ & $\mathbf{+7.5}$ & $+4.3$ & $\mathbf{+10.7}$ \\
Babel-ImageNet   & $+2.3$ & $\mathbf{+6.3}$ & $+1.3$ & $\mathbf{+9.5}$ \\
CVQA (native)    & $+1.1$ & $+1.8$          & $-3.4$ & $\sim 0$        \\
\midrule
mean (4 benchmarks) & $+1.7$ & $\mathbf{+7.8}$ & $+3.0$ & $\mathbf{+14.9}$ \\
\bottomrule
\end{tabular}
\caption{Trunk-calibration gains in tier-mean performance (pp) over the frozen baseline. Metrics are i2t \Rat{} for Flickr30k-200, XTD-200, and XM3600; top-1 accuracy for Babel-ImageNet; and multiple-choice accuracy for CVQA. Tiers follow Section~\ref{sec:setup}. The bottom row averages the four non-CVQA benchmarks; CVQA is QA-style and excluded.}
\label{tab:bench}
\end{table}

Overall, the trunk intervention of Section~\ref{sec:approx} yields measurable \LRL{} gains on all four non-CVQA benchmarks for both encoders, averaging $+7.8$ pp on \MCtwo{} and $+14.9$ pp on \SLtwo{} across the five-language \LRL{} pool. The larger \SLtwo{} gains reflect its lower frozen \LRL{} baseline and larger room to recover toward the single-source English reference. \HRL{} \Rat{} is preserved or slightly improved on retrieval benchmarks; the exception is CVQA, where \SLtwo{} regresses on \HRL{} ($-3.4$ pp) and is flat on \LRL{}, consistent with that benchmark's higher noise floor on SO400M. The \HRL{}-6-only and EN-only anchor ablations (Table~\ref{tab:anchor-ablation}) are within $\pm 1$--$2$ pp of the canonical 11-language anchor on trained-\LRL{} mean, so the replication conclusions are not anchor-specific.
\section{Per-Language Performance on All Benchmarks}
\label{sec:app-perlang-benchmarks}

\paragraph{Benchmark coverage.} This appendix reports per-language image-to-text \Rat{} (or accuracy for classification-style benchmarks) for frozen and trunk-calibrated states across six multilingual suites. \textbf{Flickr30k-200} extends Flickr30k captions to 200 languages by professional translation, with a $1{,}000$-image split and i2t \Rat{}. \textbf{XTD-200} \citep{visheratin2024nllbclip} is a 200-language COCO-based retrieval suite ($1{,}000$ images, i2t \Rat{}). \textbf{XM3600} \citep{thapliyal2022xm3600} is a natively multilingual 36-language retrieval suite; we report the full evaluation suite, distinct from the $1{,}000$-image main-paper subset (Table~\ref{tab:perlang-xm3600}). \textbf{Babel-ImageNet} \citep{geigle2024babelimagenet} is multilingual zero-shot ImageNet classification over $1{,}000$ classes. \textbf{CVQA} \citep{romero2024cvqa} is a culturally grounded 4-way multiple-choice VQA benchmark scored by image--choice-text cosine similarity. \textbf{XTD-10} \citep{aggarwal2020xtd} is the original 10-language XTD retrieval suite; four languages (en, es, zh, ko) overlap our trained pool. All evaluations are zero-shot: the trunk is trained once per encoder on the parallel-translation corpus in Section~\ref{sec:approx} and is not re-tuned per benchmark or language. All retrieval numbers in the per-benchmark subsections below are i2t; Appendix~\ref{sec:app-perdirection} additionally reports i2t and t2i separately, for both encoders and both states, on the $1{,}000$-image XM3600 subset and Flickr30k-200.

\paragraph{Calibrated state.} The \emph{cal} column uses the canonical 11-language centroid anchor (Section~\ref{sec:approx}) for all (encoder, benchmark, language) cells. The anchor was selected on a held-out 200-image XM3600 dev split (the anchor-variant panel of Appendix~\ref{sec:app-sweep-sl2}, Table~\ref{tab:sweep-sl2-anchor}), not on the test benchmarks below. \HRL{}-6-only and EN-only anchors are ablated in Table~\ref{tab:anchor-ablation} (Appendix~\ref{sec:app-sweep}); both land within $\pm 1$--$2$ pp of the canonical anchor on \LRL{} retrieval for both encoders.

\subsection{Per-direction results (i2t and t2i)}
\label{sec:app-perdirection}

Retrieval elsewhere in this paper is reported in the image-to-text (i2t) direction. This subsection reports i2t and text-to-image (t2i) \Rat{} separately, for both encoders and both states, on the $1{,}000$-image XM3600 subset of Table~\ref{tab:perlang-xm3600} and on the Flickr30k-200 $1{,}000$-image split. Table~\ref{tab:perdir-tier} gives tier-level means; Tables~\ref{tab:perdir-xm3600} and~\ref{tab:perdir-flickr} give per-language detail. The i2t columns coincide with Table~\ref{tab:perlang-xm3600} (XM3600) and Table~\ref{tab:bench-flickr30k} (Flickr30k-200). The calibration gains are not an artefact of the reported direction: the \LRL{}-5 tier mean also rises in t2i in all four (encoder, benchmark) cells --- $+9.7$ / $+9.0$\,pp on \MCtwo{} and $+21.2$ / $+25.9$\,pp on \SLtwo{} for XM3600 / Flickr30k-200, comparable to the i2t lifts --- while \HRL{}-6 t2i means stay within about a point of frozen; the single negative tier-mean delta in either direction is Flickr30k-200 \MCtwo{} \HRL{}-6 t2i ($-0.7$\,pp).

\begin{table}[h]
\centering
\small
\setlength{\tabcolsep}{4pt}
\begin{tabular}{ll|cc|cc}
\toprule
& & \multicolumn{2}{c|}{frozen} & \multicolumn{2}{c}{calibrated} \\
Encoder & Tier & i2t & t2i & i2t & t2i \\
\midrule
\multicolumn{6}{l}{\emph{XM3600 ($1{,}000$-image subset)}} \\
\MCtwo{} & \HRL{}-6 & 78.1 & 75.3 & 79.8 & 76.5 \\
\MCtwo{} & \LRL{}-5 & 44.1 & 37.5 & 53.7 & 47.2 \\
\SLtwo{} & \HRL{}-6 & 63.3 & 62.4 & 68.6 & 67.0 \\
\SLtwo{} & \LRL{}-5 & 21.7 & 16.7 & 38.8 & 37.9 \\
\midrule
\multicolumn{6}{l}{\emph{Flickr30k-200}} \\
\MCtwo{} & \HRL{}-6 & 78.3 & 77.5 & 78.9 & 76.8 \\
\MCtwo{} & \LRL{}-5 & 60.4 & 56.4 & 68.7 & 65.4 \\
\SLtwo{} & \HRL{}-6 & 63.7 & 63.4 & 67.0 & 66.9 \\
\SLtwo{} & \LRL{}-5 & 25.3 & 20.9 & 45.7 & 46.8 \\
\bottomrule
\end{tabular}
\caption{Tier-mean \Rat{} (\%) by retrieval direction, frozen vs.\ trunk-calibrated (canonical 11-language anchor), on the $1{,}000$-image XM3600 subset and Flickr30k-200. Tier means average the per-language rows of Tables~\ref{tab:perdir-xm3600} and~\ref{tab:perdir-flickr} (\HRL{}-6 includes English).}
\label{tab:perdir-tier}
\end{table}

\begin{table*}[t]
\centering
\small
\setlength{\tabcolsep}{5pt}
\begin{tabular}{ll|cccc|cccc}
\toprule
& & \multicolumn{4}{c|}{\MCtwo{}} & \multicolumn{4}{c}{\SLtwo{}} \\
& & \multicolumn{2}{c}{frozen} & \multicolumn{2}{c|}{calibrated} & \multicolumn{2}{c}{frozen} & \multicolumn{2}{c}{calibrated} \\
Lang & Grp & i2t & t2i & i2t & t2i & i2t & t2i & i2t & t2i \\
\midrule
en  & \HRL{} & 69.4 & 68.6 & 72.0 & 68.1 & 66.1 & 67.8 & 70.5 & 66.6 \\
fr  & \HRL{} & 84.4 & 80.7 & 86.2 & 81.7 & 78.2 & 80.5 & 82.6 & 79.4 \\
de  & \HRL{} & 86.5 & 83.0 & 86.8 & 82.8 & 33.7 & 26.2 & 43.8 & 43.7 \\
es  & \HRL{} & 78.0 & 74.6 & 77.6 & 75.4 & 64.1 & 65.6 & 67.4 & 66.8 \\
zh  & \HRL{} & 74.6 & 73.0 & 77.5 & 76.3 & 63.8 & 62.5 & 70.5 & 71.4 \\
ko  & \HRL{} & 75.5 & 71.8 & 78.9 & 74.5 & 74.0 & 72.0 & 77.0 & 74.1 \\
\midrule
bn  & \LRL{} & 64.7 & 53.2 & 69.0 & 60.5 & 30.2 & 21.1 & 42.7 & 42.8 \\
fil & \LRL{} & 39.9 & 37.0 & 52.7 & 46.6 & 26.9 & 23.6 & 45.0 & 42.6 \\
hi  & \LRL{} & 49.4 & 40.0 & 52.9 & 45.2 & 33.6 & 27.0 & 38.5 & 36.4 \\
sw  & \LRL{} & 22.1 & 18.2 & 42.1 & 38.1 & 12.4 & \phantom{0}9.4 & 35.7 & 38.7 \\
te  & \LRL{} & 44.3 & 38.9 & 52.0 & 45.6 & \phantom{0}5.2 & \phantom{0}2.2 & 31.9 & 28.9 \\
\midrule
\textit{\HRL{}-6 mean} & & 78.1 & 75.3 & 79.8 & 76.5 & 63.3 & 62.4 & 68.6 & 67.0 \\
\textit{\LRL{}-5 mean} & & 44.1 & 37.5 & 53.7 & 47.2 & 21.7 & 16.7 & 38.8 & 37.9 \\
\bottomrule
\end{tabular}
\caption{Per-language i2t and t2i \Rat{} (\%) on the $1{,}000$-image XM3600 subset, frozen vs.\ trunk-calibrated. The i2t columns equal Table~\ref{tab:perlang-xm3600}.}
\label{tab:perdir-xm3600}
\end{table*}

\begin{table*}[t]
\centering
\small
\setlength{\tabcolsep}{5pt}
\begin{tabular}{ll|cccc|cccc}
\toprule
& & \multicolumn{4}{c|}{\MCtwo{}} & \multicolumn{4}{c}{\SLtwo{}} \\
& & \multicolumn{2}{c}{frozen} & \multicolumn{2}{c|}{calibrated} & \multicolumn{2}{c}{frozen} & \multicolumn{2}{c}{calibrated} \\
Lang & Grp & i2t & t2i & i2t & t2i & i2t & t2i & i2t & t2i \\
\midrule
en  & \HRL{} & 86.3 & 84.3 & 85.4 & 82.2 & 86.4 & 86.5 & 85.3 & 83.1 \\
fr  & \HRL{} & 82.2 & 79.6 & 82.4 & 78.8 & 77.7 & 79.1 & 79.1 & 77.1 \\
de  & \HRL{} & 80.9 & 79.2 & 80.2 & 78.6 & 22.5 & 16.9 & 34.3 & 36.1 \\
es  & \HRL{} & 82.1 & 79.9 & 80.5 & 79.1 & 80.9 & 83.0 & 80.1 & 79.5 \\
zh  & \HRL{} & 70.7 & 69.1 & 71.8 & 70.1 & 50.4 & 50.4 & 57.6 & 59.4 \\
ko  & \HRL{} & 67.5 & 72.9 & 73.4 & 72.1 & 64.5 & 64.5 & 65.6 & 66.0 \\
\midrule
bn  & \LRL{} & 69.2 & 65.0 & 72.3 & 69.1 & 20.3 & 16.2 & 40.1 & 42.6 \\
fil & \LRL{} & 61.2 & 54.6 & 70.3 & 64.8 & 34.1 & 29.4 & 53.6 & 52.0 \\
hi  & \LRL{} & 75.7 & 71.9 & 75.2 & 73.1 & 58.1 & 49.9 & 61.6 & 62.0 \\
sw  & \LRL{} & 35.3 & 31.2 & 56.4 & 53.2 & 11.7 & \phantom{0}8.2 & 42.3 & 43.5 \\
te  & \LRL{} & 60.4 & 59.3 & 69.4 & 66.7 & \phantom{0}2.2 & \phantom{0}0.9 & 31.0 & 34.1 \\
\midrule
\textit{\HRL{}-6 mean} & & 78.3 & 77.5 & 78.9 & 76.8 & 63.7 & 63.4 & 67.0 & 66.9 \\
\textit{\LRL{}-5 mean} & & 60.4 & 56.4 & 68.7 & 65.4 & 25.3 & 20.9 & 45.7 & 46.8 \\
\bottomrule
\end{tabular}
\caption{Per-language i2t and t2i \Rat{} (\%) on the Flickr30k-200 $1{,}000$-image split, frozen vs.\ trunk-calibrated. The i2t columns equal Table~\ref{tab:bench-flickr30k}.}
\label{tab:perdir-flickr}
\end{table*}

\subsection{Flickr30k-200 (i2t \Rat{})}
\label{sec:app-perlang-flickr30k200}

We report i2t \Rat{} on the standard $1{,}000$-image split for both encoder states across the 11 trained languages; per-direction (i2t and t2i) results are in Appendix~\ref{sec:app-perdirection}.

\begin{table}[h]
\centering
\small
\setlength{\tabcolsep}{4pt}
\begin{tabular}{ll|cc|cc}
\toprule
& & \multicolumn{2}{c|}{\MCtwo{}} & \multicolumn{2}{c}{\SLtwo{}} \\
Lang & Grp & frozen & cal & frozen & cal \\
\midrule
en  & \HRL{} & 86.3 & 85.4 & 86.4 & 85.3 \\
fr  & \HRL{} & 82.2 & 82.4 & 77.7 & 79.1 \\
de  & \HRL{} & 80.9 & 80.2 & 22.5 & 34.3 \\
es  & \HRL{} & 82.1 & 80.5 & 80.9 & 80.1 \\
zh  & \HRL{} & 70.7 & 71.8 & 50.4 & 57.6 \\
ko  & \HRL{} & 67.5 & 73.4 & 64.5 & 65.6 \\
\midrule
bn  & \LRL{} & 69.2 & 72.3 & 20.3 & 40.1 \\
fil & \LRL{} & 61.2 & 70.3 & 34.1 & 53.6 \\
hi  & \LRL{} & 75.7 & 75.2 & 58.1 & 61.6 \\
sw  & \LRL{} & 35.3 & 56.4 & 11.7 & 42.3 \\
te  & \LRL{} & 60.4 & 69.4 &  2.2 & 31.0 \\
\bottomrule
\end{tabular}
\caption{Flickr30k-200 i2t \Rat{} per language (t2i in Table~\ref{tab:perdir-flickr}).}
\label{tab:bench-flickr30k}
\end{table}

\subsection{XTD-200 (\Rat{})}
\label{sec:app-perlang-xtd200}

We report image-to-text \Rat{} on the standard $1{,}000$-image split for the 11 trained languages.

\begin{table}[h]
\centering
\small
\setlength{\tabcolsep}{4pt}
\begin{tabular}{ll|cc|cc}
\toprule
& & \multicolumn{2}{c|}{\MCtwo{}} & \multicolumn{2}{c}{\SLtwo{}} \\
Lang & Grp & frozen & cal & frozen & cal \\
\midrule
en  & \HRL{} & 70.1 & 72.6 & 76.8 & 76.6 \\
fr  & \HRL{} & 69.4 & 68.6 & 69.7 & 71.8 \\
de  & \HRL{} & 68.5 & 68.3 & 19.5 & 27.1 \\
es  & \HRL{} & 67.5 & 69.6 & 69.9 & 73.3 \\
zh  & \HRL{} & 60.6 & 62.7 & 49.2 & 52.5 \\
ko  & \HRL{} & 54.2 & 62.0 & 57.7 & 60.9 \\
\midrule
bn  & \LRL{} & 57.5 & 60.7 & 34.4 & 49.7 \\
fil & \LRL{} & 51.3 & 61.0 & 45.4 & 60.2 \\
hi  & \LRL{} & 61.8 & 65.5 & 55.4 & 60.7 \\
sw  & \LRL{} & 27.8 & 47.2 & 21.1 & 44.4 \\
te  & \LRL{} & 50.8 & 59.7 &  4.7 & 41.5 \\
\bottomrule
\end{tabular}
\caption{XTD-200 \Rat{} per language.}
\label{tab:bench-xtd200}
\end{table}

\subsection{XM3600 (\Rat{}, retrieval-style)}
\label{sec:app-perlang-xm3600}

Table~\ref{tab:perlang-xm3600} reports the $1{,}000$-image main-paper subset; below we use the full XM3600 evaluation suite.
\begin{table}[h]
\centering
\small
\setlength{\tabcolsep}{4pt}
\begin{tabular}{ll|cc|cc}
\toprule
& & \multicolumn{2}{c|}{\MCtwo{}} & \multicolumn{2}{c}{\SLtwo{}} \\
Lang & Grp & frozen & cal & frozen & cal \\
\midrule
en  & \HRL{} & 50.9 & 53.1 & 47.5 & 52.9 \\
fr  & \HRL{} & 71.5 & 72.2 & 64.9 & 67.6 \\
de  & \HRL{} & 76.2 & 75.9 & 21.8 & 28.6 \\
es  & \HRL{} & 62.5 & 63.4 & 48.4 & 50.7 \\
zh  & \HRL{} & 61.7 & 63.4 & 50.4 & 56.4 \\
ko  & \HRL{} & 61.0 & 64.2 & 58.1 & 60.7 \\
\midrule
bn  & \LRL{} & 47.2 & 51.9 & 17.5 & 26.3 \\
fil & \LRL{} & 25.7 & 35.2 & 16.9 & 27.7 \\
hi  & \LRL{} & 34.1 & 35.6 & 20.8 & 23.0 \\
sw  & \LRL{} & 12.8 & 27.9 &  6.5 & 22.9 \\
te  & \LRL{} & 28.4 & 35.1 &  2.5 & 17.6 \\
\bottomrule
\end{tabular}
\caption{XM3600 \Rat{} per language on the full evaluation suite.}
\label{tab:bench-xm3600-full}
\end{table}

\subsection{Babel-ImageNet (top-1 classification accuracy)}
\label{sec:app-perlang-babel}

We report top-1 accuracy over the $1{,}000$ ImageNet classes per trained language; the classification metric is not directly comparable to retrieval \Rat{}.

\begin{table}[h]
\centering
\small
\setlength{\tabcolsep}{4pt}
\begin{tabular}{ll|cc|cc}
\toprule
& & \multicolumn{2}{c|}{\MCtwo{}} & \multicolumn{2}{c}{\SLtwo{}} \\
Lang & Grp & frozen & cal & frozen & cal \\
\midrule
en  & \HRL{} & 74.0 & 73.5 & 66.8 & 68.2 \\
fr  & \HRL{} & 66.9 & 69.4 & 59.4 & 60.3 \\
de  & \HRL{} & 65.6 & 69.5 & 23.1 & 29.3 \\
es  & \HRL{} & 62.6 & 65.0 & 57.8 & 57.6 \\
zh  & \HRL{} & 59.0 & 61.4 & 58.2 & 59.7 \\
ko  & \HRL{} & 59.6 & 62.6 & 59.5 & 57.8 \\
\midrule
bn  & \LRL{} & 59.9 & 64.8 & 40.0 & 44.4 \\
fil & \LRL{} & 26.3 & 32.1 & 20.0 & 23.1 \\
hi  & \LRL{} & 53.8 & 58.6 & 43.9 & 46.4 \\
sw  & \LRL{} & 15.1 & 24.3 & 10.9 & 21.1 \\
te  & \LRL{} & 45.4 & 51.9 &  9.8 & 37.4 \\
\bottomrule
\end{tabular}
\caption{Babel-ImageNet top-1 classification accuracy per language.}
\label{tab:bench-babel}
\end{table}

\subsection{CVQA (native split, accuracy)}
\label{sec:app-perlang-cvqa}

CVQA covers es, zh, ko, bn, fil, hi, sw, and te from this paper's pool; en, fr, and de are outside the native split.
\begin{table}[h]
\centering
\small
\setlength{\tabcolsep}{4pt}
\begin{tabular}{ll|cc|cc}
\toprule
& & \multicolumn{2}{c|}{\MCtwo{}} & \multicolumn{2}{c}{\SLtwo{}} \\
Lang & Grp & frozen & cal & frozen & cal \\
\midrule
es  & \HRL{} & 62.5 & 63.7 & 33.3 & 33.9 \\
zh  & \HRL{} & 71.3 & 74.2 & 64.6 & 57.4 \\
ko  & \HRL{} & 70.7 & 70.0 & 55.5 & 52.1 \\
\midrule
bn  & \LRL{} & 62.6 & 59.1 & 44.4 & 39.2 \\
fil & \LRL{} & 49.8 & 48.3 & 28.6 & 31.5 \\
hi  & \LRL{} & 66.7 & 68.2 & 57.2 & 52.7 \\
sw  & \LRL{} & 44.0 & 50.9 & 31.5 & 34.1 \\
te  & \LRL{} & 53.5 & 59.0 & 34.0 & 35.0 \\
\bottomrule
\end{tabular}
\caption{CVQA native-split multiple-choice accuracy per language. CVQA is scored by cosine similarity between image and choice-text embeddings; it is \emph{not} generative VQA. Languages outside the native set (en, fr, de) are excluded.}
\label{tab:bench-cvqa}
\end{table}

\subsection{XTD-10 (\Rat{})}
\label{sec:app-perlang-xtd10}

XTD-10 covers only en, es, zh, and ko from our pool, all \HRL{}; no \LRL{} mean is reported. The \emph{cal} column uses the canonical 11-language anchor (Section~\ref{sec:approx}).
\begin{table}[h]
\centering
\small
\setlength{\tabcolsep}{4pt}
\begin{tabular}{ll|cc|cc}
\toprule
& & \multicolumn{2}{c|}{\MCtwo{}} & \multicolumn{2}{c}{\SLtwo{}} \\
Lang & Grp & frozen & cal & frozen & cal \\
\midrule
en  & \HRL{} & 70.1 & 72.6 & 76.8 & 76.6 \\
es  & \HRL{} & 68.4 & 70.8 & 71.1 & 72.2 \\
zh  & \HRL{} & 65.5 & 66.9 & 56.7 & 59.6 \\
ko  & \HRL{} & 58.2 & 63.0 & 56.8 & 60.7 \\
\bottomrule
\end{tabular}
\caption{XTD-10 \Rat{} per language. Covers only four \HRL{} languages in this paper's pool.}
\label{tab:bench-xtd10}
\end{table}

\paragraph{Reading these tables.} \HRL{} blocks show modest calibration changes (typically $\pm 5$ pp, mostly positive, with occasional small regressions), consistent with non-targeted movement toward the 11-language centroid. \LRL{} blocks show consistent double-digit gains on retrieval and substantial gains on Babel-ImageNet. CVQA is noisier because each item is scored by relative similarity among four candidate texts rather than absolute image--text alignment. Single-digit frozen \SLtwo{} \LRL{} cells (te on Flickr30k-200, XTD-200, and Babel-ImageNet) may partly reflect tokenisation and sequence-length constraints, which we do not isolate on these benchmarks; calibration substantially recovers them despite operating at a fixed mid-layer.
\section{Trunk-Training Hyperparameter and Design Ablation}
\label{sec:app-sweep}

This appendix reports the sweep used to select the canonical calibrated configuration. Candidates were ranked by training-time paired cosine (en$\leftrightarrow$others on the 11-language pool, \HRL{}-6 + \LRL{}-5) and held-out retrieval on a 200-image XM3600 subset. The absolute \Rat{} numbers therefore differ from full-evaluation results (Table~\ref{tab:perlang-xm3600}, Appendix~\ref{sec:app-benchmarks}), since this smaller validation pool is used only for model selection. We sampled 23 \MCtwo{} and 30 \SLtwo{} runs across the axes below; ``\LRL{} pc'' denotes training-time paired cosine averaged over the five \LRL{} languages.

\subsection{MetaCLIP-2 Worldwide-Huge sweep}
\label{sec:app-sweep-mc2}

We vary four axes on \MCtwo{}---trainable depth $M$, alignment-loss weight $\lambda_{\mathrm{align}}$, learning rate, and hard re-init vs.\ soft fine-tune---fixing the other axes in each panel.

\paragraph{Trainable depth $M$.} Front transformer blocks made trainable; remainder frozen. $\lambda_{\mathrm{align}}{=}1$, lr $= 6{\times}10^{-5}$, soft.

\begin{table}[h]
\centering
\small
\setlength{\tabcolsep}{4pt}
\begin{tabular}{cccc}
\toprule
$M$ & \HRL{} \Rat{} & \LRL{} \Rat{} & \LRL{} pc \\
\midrule
2 & 0.907 & 0.669 & 0.218 \\
3 & 0.911 & 0.681 & 0.218 \\
\textbf{4} & \textbf{0.911} & \textbf{0.674} & \textbf{0.218} \\
5 & 0.911 & 0.678 & 0.219 \\
7 & 0.909 & 0.678 & 0.219 \\
8 & 0.910 & 0.687 & 0.219 \\
\bottomrule
\end{tabular}
\caption{\MCtwo{} sweep: trainable depth $M$.}
\label{tab:sweep-mc2-M}
\end{table}

\paragraph{Alignment-loss weight $\lambda_{\mathrm{align}}$.} Fixed $M{=}4$, soft. Two learning rates expose the joint surface.

\begin{table}[h]
\centering
\small
\setlength{\tabcolsep}{4pt}
\begin{tabular}{cccc}
\toprule
$\lambda_{\mathrm{align}}$ (lr) & \HRL{} \Rat{} & \LRL{} \Rat{} & \LRL{} pc \\
\midrule
0.5\ \ (6e-5)   & 0.910 & 0.673 & 0.216 \\
1.0\ \ (6e-5)   & 0.911 & 0.674 & 0.218 \\
2.0\ \ (6e-5)   & 0.911 & 0.690 & 0.221 \\
1.0\ \ (1.2e-4) & 0.908 & 0.689 & 0.221 \\
\textbf{2.0\ \ (1.2e-4)} & \textbf{0.908} & \textbf{0.706} & \textbf{0.224} \\
\bottomrule
\end{tabular}
\caption{\MCtwo{} sweep: alignment-loss weight $\lambda_{\mathrm{align}}$ at two learning rates.}
\label{tab:sweep-mc2-lambda}
\end{table}

\paragraph{Learning rate.} Fixed $M{=}4$, $\lambda_{\mathrm{align}}{=}1$, soft.

\begin{table}[h]
\centering
\small
\setlength{\tabcolsep}{4pt}
\begin{tabular}{cccc}
\toprule
lr & \HRL{} \Rat{} & \LRL{} \Rat{} & \LRL{} pc \\
\midrule
1e-5   & 0.907 & 0.645 & 0.210 \\
3e-5   & 0.906 & 0.658 & 0.215 \\
6e-5   & 0.911 & 0.674 & 0.218 \\
\textbf{1.2e-4} & \textbf{0.908} & \textbf{0.689} & \textbf{0.221} \\
\bottomrule
\end{tabular}
\caption{\MCtwo{} sweep: learning rate at $M{=}4$, $\lambda_{\mathrm{align}}{=}1$.}
\label{tab:sweep-mc2-lr}
\end{table}

\paragraph{Hard re-init vs.\ soft fine-tune.}

\begin{table}[h]
\centering
\small
\setlength{\tabcolsep}{3pt}
\resizebox{\columnwidth}{!}{%
\begin{tabular}{lccc}
\toprule
variant ($M{=}4$) & \HRL{} \Rat{} & \LRL{} \Rat{} & \LRL{} pc \\
\midrule
hard re-init ($\lambda{=}3$, lr 1e-4)    & 0.778 & 0.501 & 0.189 \\
soft ($\lambda{=}1$, lr 1.2e-4)          & 0.908 & 0.689 & 0.221 \\
\textbf{soft ($\lambda{=}2$, lr 1.2e-4)} & \textbf{0.908} & \textbf{0.706} & \textbf{0.224} \\
\bottomrule
\end{tabular}%
}
\caption{\MCtwo{} sweep: hard re-init vs.\ soft fine-tune of the front $M$ blocks.}
\label{tab:sweep-mc2-reinit}
\end{table}

Hard re-init drops \HRL{} by $\approx 13$\,pp and \LRL{} by $\approx 20$\,pp at comparable learning rate; soft fine-tune, initialised from frozen weights, is needed for the trunk objective to converge.

\paragraph{Selected configuration.} The canonical \MCtwo{} calibrated state uses $M{=}4$, $\lambda_{\mathrm{align}}{=}2$, lr $= 1.2{\times}10^{-4}$, soft fine-tune, and the 11-language centroid anchor---the \LRL{} \Rat{} maximizer at \HRL{} parity in the sweep.

\subsection{SigLIP-2 SO400M sweep}
\label{sec:app-sweep-sl2}

We sweep the same axes on \SLtwo{} and add an anchor-variant panel, reflecting the dominant interaction observed on \MCtwo{}. The \SLtwo{} optimum uses a higher learning rate, consistent with different effective gradient magnitudes under bidirectional vs.\ causal attention.

\paragraph{Trainable depth $M$.} $\lambda_{\mathrm{align}}{=}1$, lr $= 1{\times}10^{-5}$, soft.

\begin{table}[h]
\centering
\small
\setlength{\tabcolsep}{4pt}
\begin{tabular}{cccc}
\toprule
$M$ & \HRL{} \Rat{} & \LRL{} \Rat{} & \LRL{} pc \\
\midrule
1 & 0.817 & 0.391 & 0.066 \\
2 & 0.820 & 0.392 & 0.065 \\
\textbf{3} & \textbf{0.818} & \textbf{0.392} & \textbf{0.064} \\
4 & 0.820 & 0.394 & 0.064 \\
6 & 0.813 & 0.398 & 0.063 \\
9 & 0.815 & 0.398 & 0.063 \\
\bottomrule
\end{tabular}
\caption{\SLtwo{} sweep: trainable depth $M$.}
\label{tab:sweep-sl2-M}
\end{table}

\paragraph{Alignment-loss weight $\lambda_{\mathrm{align}}$.} Fixed $M{=}3$, lr $= 6{\times}10^{-5}$, soft.

\begin{table}[h]
\centering
\small
\setlength{\tabcolsep}{4pt}
\begin{tabular}{cccc}
\toprule
$\lambda_{\mathrm{align}}$ & \HRL{} \Rat{} & \LRL{} \Rat{} & \LRL{} pc \\
\midrule
0.5 & 0.829 & 0.475 & 0.075 \\
\textbf{1.0} & \textbf{0.836} & \textbf{0.494} & \textbf{0.075} \\
2.0 & 0.844 & 0.504 & 0.076 \\
\bottomrule
\end{tabular}
\caption{\SLtwo{} sweep: alignment-loss weight $\lambda_{\mathrm{align}}$.}
\label{tab:sweep-sl2-lambda}
\end{table}

\paragraph{Learning rate.} Fixed $M{=}3$, $\lambda_{\mathrm{align}}{=}1$, soft.

\begin{table}[h]
\centering
\small
\setlength{\tabcolsep}{4pt}
\begin{tabular}{cccc}
\toprule
lr & \HRL{} \Rat{} & \LRL{} \Rat{} & \LRL{} pc \\
\midrule
1.5e-5 & 0.818 & 0.395 & 0.063 \\
3e-5   & 0.826 & 0.408 & 0.064 \\
6e-5   & 0.836 & 0.494 & 0.075 \\
1.2e-4 & 0.840 & 0.576 & 0.089 \\
\textbf{2.4e-4} & \textbf{0.833} & \textbf{0.608} & \textbf{0.095} \\
\bottomrule
\end{tabular}
\caption{\SLtwo{} sweep: learning rate at $M{=}3$, $\lambda_{\mathrm{align}}{=}1$.}
\label{tab:sweep-sl2-lr}
\end{table}

\SLtwo{} \LRL{} \Rat{} climbs up to $2.4{\times}10^{-4}$; higher rates were unstable in pilot runs, collapsing around lr $\geq 5{\times}10^{-4}$ under bidirectional attention.

\paragraph{Hard re-init.} No \SLtwo{} hard-reinit runs survive in the sweep file. Pilot experiments in the predecessor codebase collapsed retrieval to single digits for every tested $M$, suggesting that reinitialised front blocks emit out-of-distribution states the fixed bidirectional back-half cannot recover, unlike causal \MCtwo{} where late tokens can partly attend around the perturbation. We therefore use soft fine-tune for \SLtwo{} calibration.

\paragraph{Anchor variant.} Fixed $M{=}3$, $\lambda_{\mathrm{align}}{=}1$, lr $= 2.4{\times}10^{-4}$, soft.

\begin{table}[h]
\centering
\small
\setlength{\tabcolsep}{4pt}
\begin{tabular}{lcccc}
\toprule
anchor & \HRL{} \Rat{} & \LRL{} \Rat{} & \LRL{} pc & centroid-cos \\
\midrule
en-only       & 0.831 & 0.576 & 0.078 & 0.604 \\
\HRL{}-only   & 0.824 & 0.624 & 0.089 & 0.772 \\
\textbf{11-lang} & \textbf{0.833} & \textbf{0.608} & \textbf{0.095} & \textbf{0.823} \\
\bottomrule
\end{tabular}
\caption{\SLtwo{} sweep: anchor variant at $M{=}3$, $\lambda_{\mathrm{align}}{=}1$, lr $2.4{\times}10^{-4}$.}
\label{tab:sweep-sl2-anchor}
\end{table}

The 11-language centroid anchor attains the highest training-time paired cosine and centroid alignment with near-best \LRL{} \Rat{}. Although \HRL{}-only slightly leads on \LRL{} \Rat{}, its lower \LRL{} paired cosine supports the 11-language centroid as the canonical target.

\paragraph{Selected configuration.} The canonical \SLtwo{} calibrated state uses $M{=}3$, $\lambda_{\mathrm{align}}{=}1$, lr $= 2.4{\times}10^{-4}$, soft fine-tune, and the 11-language centroid anchor.

\paragraph{Summary.} Across both encoders, $M$ plateaus between 2 and 8, $\lambda_{\mathrm{align}}{=}2$ helps modestly at higher learning rates, and the main interaction is learning rate with anchor choice. \MCtwo{} peaks near lr $1.2{\times}10^{-4}$ with $\lambda_{\mathrm{align}}{=}2$ and the 11-language anchor, whereas \SLtwo{} needs about $2\times$ that rate. Hard re-init is unviable bidirectionally and dominated by soft fine-tune causally, making soft fine-tune canonical for Equation~\ref{eq:cal-bound}.

\subsection{Anchor-Variant Ablation on All Benchmarks}
\label{sec:app-anchor-ablation}

To check that the sweep-selected 11-language centroid is not a subset artifact, we evaluate three anchor variants (\textbf{11-lang} canonical, \textbf{\HRL{}-6} centroid, \textbf{EN-only}) on the five trained-language benchmarks of Appendix~\ref{sec:app-benchmarks}. Each variant fixes $(M, \lambda_{\mathrm{align}}, \mathrm{lr})$ at the canonical setting and changes only the anchor; \HRL{}-6 and EN-only trunks are retrained from scratch. Table~\ref{tab:anchor-ablation} reports the \LRL{}-5 pool mean i2t \Rat{} for each cell.

\begin{table}[h]
\centering
\small
\setlength{\tabcolsep}{3pt}
\resizebox{\columnwidth}{!}{%
\begin{tabular}{l|ccc|ccc}
\toprule
 & \multicolumn{3}{c|}{\MCtwo{} \LRL{}-5} & \multicolumn{3}{c}{\SLtwo{} \LRL{}-5} \\
Benchmark & 11-lang & \HRL{}-6 & EN & 11-lang & \HRL{}-6 & EN \\
\midrule
XM3600 (full)   & \textbf{37.1} & 35.7 & 34.6 & 23.5 & \textbf{24.1} & 23.4 \\
Flickr30k-200   & \textbf{68.7} & 68.1 & 67.4 & 45.7 & \textbf{47.3} & 47.0 \\
XTD-200         & \textbf{58.8} & 57.4 & 56.8 & \textbf{51.3} & 50.6 & 49.5 \\
Babel-ImageNet  & \textbf{46.4} & 45.2 & 43.9 & \textbf{34.5} & 33.2 & 28.5 \\
CVQA (native)   & \textbf{57.1} & 55.7 & 55.0 & \textbf{38.5} & 36.9 & 35.7 \\
\midrule
\textit{x-bench mean} & \textbf{53.6} & 52.4 & 51.5 & \textbf{38.7} & 38.4 & 36.8 \\
\bottomrule
\end{tabular}%
}
\caption{Anchor-variant ablation: trained-\LRL{}-5 pool mean i2t \Rat{} (\%) for each anchor variant on each benchmark, both encoders (EN $=$ EN-only). The 11-lang canonical variant wins on cross-benchmark mean for both encoders by $1$--$2$\,pp; \HRL{}-6 wins on individual cells (Flickr30k-200 \SLtwo{}, XM3600 \SLtwo{}) but loses on average; EN-only is dominated everywhere. Same evaluation pipeline as Appendix~\ref{sec:app-benchmarks}.}
\label{tab:anchor-ablation}
\end{table}

The 11-language anchor wins on cross-benchmark mean for both encoders, matching the training-time sweep. Per-cell differences are small ($\leq 2$\,pp on \MCtwo{}, $\leq 6$\,pp on \SLtwo{} Babel-ImageNet) relative to the main patching/calibration deltas, so the conclusions of Sections~\ref{sec:results}--\ref{sec:analysis} are not anchor-specific.
\section{Additional Ablations and Detailed Geometric Probes}
\label{sec:app-extra}

This appendix collects ablations and detailed geometric probes that sit outside the main sweep of Appendix~\ref{sec:app-sweep}.

\subsection{Detailed CKA: Definition}
\label{sec:app-cka}

Figure~\ref{fig:cka-frozen} in Section~\ref{sec:analysis} shows the main curves. The CKA definition \citep{kornblith2019cka}: for two layer-$\ell$ hidden-state matrices $H_\ell^A, H_\ell^B \in \mathbb{R}^{N \times d}$ of $N$ parallel inputs in languages $A,B$ (mean-centered rows),
\begin{equation}
\resizebox{0.95\columnwidth}{!}{$\displaystyle
  \cka(H_\ell^A, H_\ell^B) = \frac{\bigl\lVert (H_\ell^B)^\top H_\ell^A \bigr\rVert_F^{\,2}}{\bigl\lVert (H_\ell^A)^\top H_\ell^A \bigr\rVert_F \cdot \bigl\lVert (H_\ell^B)^\top H_\ell^B \bigr\rVert_F}
$}.
  \label{eq:cka}
\end{equation}
\paragraph{Cross-metric corroboration.} CKA is rotation-invariant; the Top-$K{=}64$ positive-activation Jaccard \citep{geva2020feedforward} adds the ``different neurons fire'' interpretation, with the same tier hierarchy at the last block (\MCtwo{} frozen $0.229/0.160$ for \HRL{}--\HRL{}/\LRL{}; \SLtwo{} $0.212/0.158$). The text--image modality gap of \citet{liang2022modalitygap} provides a natural scale: tier-level language gaps grow monotonically (\MCtwo{} $30.4/\mathbf{36.7}\%$ for \HRL{}--\HRL{}/\LRL{}, \SLtwo{} $18.1/\mathbf{25.0}\%$, non-overlapping bootstrap CIs).

\subsection{Layer-wise Language-Identification Classifier}
\label{sec:app-clf-layerwise}

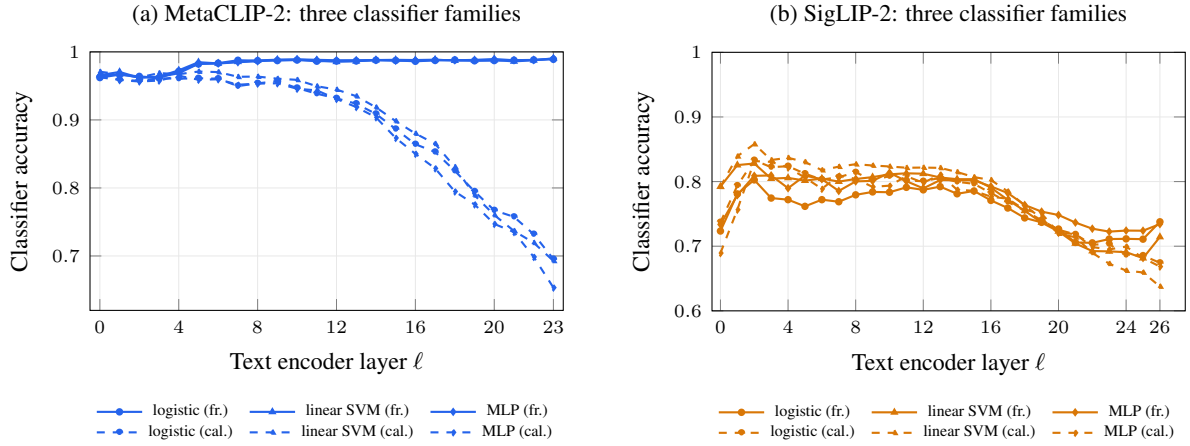
\begin{figure*}[t]
\centering
\begin{minipage}{0.49\textwidth}\centering
\begin{tikzpicture}
\begin{axis}[
  width=\textwidth, height=5.0cm,
  xlabel={Text encoder layer $\ell$}, ylabel={Classifier accuracy},
  xmin=-0.5, xmax=23.5, ymin=0.62, ymax=1.0,
  xtick={0,4,8,12,16,20,23}, ytick={0.7,0.8,0.9,1.0},
  legend style={font=\tiny, at={(0.5,-0.32)}, anchor=north, fill=white, draw=none, row sep=-2pt, inner sep=2pt},
  legend columns=3, /tikz/every even column/.append style={column sep=4pt},
  tick label style={font=\scriptsize}, label style={font=\footnotesize},
  title style={font=\footnotesize}, title={(a) \MCtwo{}: three classifier families},
  grid=both, grid style={gray!18, very thin},
]
\addplot[color=tHRL, mark=*, mark size=1pt, thick] coordinates {(0,0.9622)(1,0.9665)(2,0.9640)(3,0.9617)(4,0.9714)(5,0.9822)(6,0.9834)(7,0.9877)(8,0.9871)(9,0.9872)(10,0.9880)(11,0.9863)(12,0.9865)(13,0.9866)(14,0.9878)(15,0.9874)(16,0.9862)(17,0.9874)(18,0.9882)(19,0.9869)(20,0.9872)(21,0.9878)(22,0.9882)(23,0.9891)};
\addlegendentry{logistic (fr.)}
\addplot[color=tHRL, mark=triangle*, mark size=1pt, thick] coordinates {(0,0.9658)(1,0.9705)(2,0.9620)(3,0.9637)(4,0.9729)(5,0.9848)(6,0.9831)(7,0.9855)(8,0.9874)(9,0.9885)(10,0.9885)(11,0.9874)(12,0.9855)(13,0.9862)(14,0.9880)(15,0.9883)(16,0.9880)(17,0.9885)(18,0.9872)(19,0.9875)(20,0.9880)(21,0.9860)(22,0.9877)(23,0.9905)};
\addlegendentry{linear SVM (fr.)}
\addplot[color=tHRL, mark=diamond*, mark size=1pt, thick] coordinates {(0,0.9652)(1,0.9669)(2,0.9625)(3,0.9631)(4,0.9686)(5,0.9825)(6,0.9829)(7,0.9848)(8,0.9862)(9,0.9882)(10,0.9891)(11,0.9883)(12,0.9877)(13,0.9878)(14,0.9880)(15,0.9871)(16,0.9872)(17,0.9882)(18,0.9877)(19,0.9882)(20,0.9895)(21,0.9877)(22,0.9880)(23,0.9900)};
\addlegendentry{MLP (fr.)}
\addplot[color=tHRL, mark=*, mark size=1pt, thick, dashed] coordinates {(0,0.9626)(1,0.9589)(2,0.9575)(3,0.9612)(4,0.9606)(5,0.9612)(6,0.9606)(7,0.9497)(8,0.9538)(9,0.9568)(10,0.9469)(11,0.9385)(12,0.9317)(13,0.9238)(14,0.9088)(15,0.8866)(16,0.8642)(17,0.8529)(18,0.8252)(19,0.7948)(20,0.7671)(21,0.7575)(22,0.7320)(23,0.6954)};
\addlegendentry{logistic (cal.)}
\addplot[color=tHRL, mark=triangle*, mark size=1pt, thick, dashed] coordinates {(0,0.9708)(1,0.9671)(2,0.9625)(3,0.9688)(4,0.9672)(5,0.9711)(6,0.9703)(7,0.9634)(8,0.9637)(9,0.9606)(10,0.9589)(11,0.9495)(12,0.9446)(13,0.9349)(14,0.9185)(15,0.8983)(16,0.8800)(17,0.8649)(18,0.8308)(19,0.7892)(20,0.7602)(21,0.7368)(22,0.7194)(23,0.6928)};
\addlegendentry{linear SVM (cal.)}
\addplot[color=tHRL, mark=diamond*, mark size=1pt, thick, dashed] coordinates {(0,0.9657)(1,0.9585)(2,0.9565)(3,0.9578)(4,0.9649)(5,0.9591)(6,0.9597)(7,0.9517)(8,0.9534)(9,0.9538)(10,0.9458)(11,0.9434)(12,0.9308)(13,0.9189)(14,0.9038)(15,0.8735)(16,0.8502)(17,0.8288)(18,0.7948)(19,0.7758)(20,0.7471)(21,0.7351)(22,0.6986)(23,0.6535)};
\addlegendentry{MLP (cal.)}
\addplot[color=black, dotted, thin] coordinates {(0,0.0769)(23,0.0769)};
\end{axis}
\end{tikzpicture}
\end{minipage}\hfill
\begin{minipage}{0.49\textwidth}\centering
\begin{tikzpicture}
\begin{axis}[
  width=\textwidth, height=5.0cm,
  xlabel={Text encoder layer $\ell$}, ylabel={Classifier accuracy},
  xmin=-0.5, xmax=27.5, ymin=0.60, ymax=1.0,
  xtick={0,4,8,12,16,20,24,26}, ytick={0.6,0.7,0.8,0.9,1.0},
  legend style={font=\tiny, at={(0.5,-0.32)}, anchor=north, fill=white, draw=none, row sep=-2pt, inner sep=2pt},
  legend columns=3, /tikz/every even column/.append style={column sep=4pt},
  tick label style={font=\scriptsize}, label style={font=\footnotesize},
  title style={font=\footnotesize}, title={(b) \SLtwo{}: three classifier families},
  grid=both, grid style={gray!18, very thin},
]
\addplot[color=tLRL, mark=*, mark size=1pt, thick] coordinates {(0,0.7235)(1,0.7809)(2,0.8026)(3,0.7745)(4,0.7720)(5,0.7617)(6,0.7720)(7,0.7688)(8,0.7795)(9,0.7840)(10,0.7835)(11,0.7911)(12,0.7872)(13,0.7922)(14,0.7811)(15,0.7855)(16,0.7709)(17,0.7591)(18,0.7438)(19,0.7369)(20,0.7266)(21,0.7058)(22,0.7052)(23,0.7111)(24,0.7114)(25,0.7109)(26,0.7378)};
\addlegendentry{logistic (fr.)}
\addplot[color=tLRL, mark=triangle*, mark size=1pt, thick] coordinates {(0,0.7923)(1,0.8255)(2,0.8280)(3,0.8046)(4,0.8057)(5,0.8015)(6,0.8046)(7,0.8000)(8,0.8040)(9,0.8069)(10,0.8112)(11,0.8128)(12,0.8122)(13,0.8063)(14,0.8037)(15,0.8037)(16,0.7906)(17,0.7717)(18,0.7548)(19,0.7366)(20,0.7229)(21,0.7043)(22,0.6925)(23,0.6920)(24,0.6908)(25,0.6815)(26,0.7143)};
\addlegendentry{linear SVM (fr.)}
\addplot[color=tLRL, mark=diamond*, mark size=1pt, thick] coordinates {(0,0.7337)(1,0.7791)(2,0.8089)(3,0.8095)(4,0.7898)(5,0.8092)(6,0.8038)(7,0.7855)(8,0.8003)(9,0.8011)(10,0.8137)(11,0.8006)(12,0.7895)(13,0.8023)(14,0.8020)(15,0.8020)(16,0.7925)(17,0.7823)(18,0.7638)(19,0.7535)(20,0.7483)(21,0.7366)(22,0.7271)(23,0.7226)(24,0.7242)(25,0.7240)(26,0.7342)};
\addlegendentry{MLP (fr.)}
\addplot[color=tLRL, mark=*, mark size=1pt, thick, dashed] coordinates {(0,0.7372)(1,0.7938)(2,0.8331)(3,0.8223)(4,0.8238)(5,0.8122)(6,0.8031)(7,0.8068)(8,0.8143)(9,0.8023)(10,0.8083)(11,0.8017)(12,0.8005)(13,0.8057)(14,0.7997)(15,0.7965)(16,0.7822)(17,0.7686)(18,0.7549)(19,0.7402)(20,0.7255)(21,0.7182)(22,0.7025)(23,0.7031)(24,0.6865)(25,0.6855)(26,0.6743)};
\addlegendentry{logistic (cal.)}
\addplot[color=tLRL, mark=triangle*, mark size=1pt, thick, dashed] coordinates {(0,0.7920)(1,0.8392)(2,0.8580)(3,0.8334)(4,0.8368)(5,0.8303)(6,0.8177)(7,0.8231)(8,0.8269)(9,0.8249)(10,0.8234)(11,0.8214)(12,0.8217)(13,0.8212)(14,0.8148)(15,0.8069)(16,0.8025)(17,0.7848)(18,0.7632)(19,0.7428)(20,0.7195)(21,0.7138)(22,0.6898)(23,0.6729)(24,0.6623)(25,0.6594)(26,0.6377)};
\addlegendentry{linear SVM (cal.)}
\addplot[color=tLRL, mark=diamond*, mark size=1pt, thick, dashed] coordinates {(0,0.6894)(1,0.7565)(2,0.8317)(3,0.8305)(4,0.8217)(5,0.8062)(6,0.7888)(7,0.8091)(8,0.8043)(9,0.7918)(10,0.7937)(11,0.8083)(12,0.7982)(13,0.8109)(14,0.7874)(15,0.7857)(16,0.7766)(17,0.7735)(18,0.7563)(19,0.7505)(20,0.7208)(21,0.7188)(22,0.6986)(23,0.6951)(24,0.6994)(25,0.6818)(26,0.6685)};
\addlegendentry{MLP (cal.)}
\addplot[color=black, dotted, thin] coordinates {(0,0.0769)(27,0.0769)};
\end{axis}
\end{tikzpicture}
\end{minipage}
\caption{Layer-wise language-classifier accuracy on COCO-translated train / XM3600 test (OOD, 13 languages, chance $7.7\%$). Logistic, linear SVM, single-hidden-layer MLP curves track within $\leq 3$\,pp on both encoders in both states---the language signal is a clear linear direction. \MCtwo{}: frozen $96$--$99\%$, calibrated decays to $\sim 65$--$69\%$ at $\ell{=}23$. \SLtwo{}: frozen $71$--$83\%$, calibrated $64$--$86\%$ ($\sim 63$--$70\%$ at $\ell{=}26$).}
\label{fig:clf-vs-layer}
\end{figure*}

Figure~\ref{fig:clf-vs-layer} extends the headline language-classifier probe of Section~\ref{sec:clf} from the final projected embedding to every intermediate transformer block, showing how the linear language signal evolves along the forward path on both encoders and both states. Three classifier families (logistic regression, linear SVM, single-hidden-layer MLP) are reported to verify the signal is linear rather than a non-linear artefact of any one probe.

\subsection{Higher-Rank Linear Erasure: INLP-Iterated Sweep}
\label{sec:app-inlp}

INLP \citep{ravfogel2020inlp} iterated to ranks $\{12, 32, 64, 128\}$ on the projected text embedding, with same-rank random orthogonal-subspace controls. Across both encoders and both states, classifier accuracy decays from $\geq 90\%$ to $37$--$50\%$ at rank 128, while retrieval stays generally close to the same-rank random control, with deviations of up to $2.9$\,pp in the reported cells (\MCtwo{} frozen \LRL{} at rank 128: $43.8$ vs.\ $40.9$). The bias-displacement falsification of Section~\ref{sec:orth} is therefore stable from rank 12 to rank 128 on both architectures.

\begin{table}[h]
\centering
\scriptsize
\setlength{\tabcolsep}{3pt}
\begin{tabular}{l|c|cc|cc}
\toprule
 & clf acc & \multicolumn{2}{c|}{\HRL{} \Rat{} (\%)} & \multicolumn{2}{c}{\LRL{} \Rat{} (\%)} \\
$k$ & $\to$ after & INLP & rand-$k$ & INLP & rand-$k$ \\
\midrule
\multicolumn{6}{l}{\MCtwo{} frozen \ (base: \HRL{} $78.1$, \LRL{} $44.1$)} \\
12  & 0.96 & 78.0 & 78.1 & 44.2 & 43.8 \\
64  & 0.59 & 77.3 & 77.8 & 42.9 & 43.0 \\
128 & 0.47 & 77.3 & 76.5 & 43.8 & 40.9 \\
\midrule
\multicolumn{6}{l}{\MCtwo{} calibrated \ (base: \HRL{} $79.8$, \LRL{} $53.7$)} \\
12  & 0.57 & 79.9 & 79.7 & 53.5 & 53.2 \\
64  & 0.43 & 79.2 & 79.1 & 52.3 & 52.3 \\
128 & 0.37 & 79.2 & 78.1 & 52.3 & 50.6 \\
\midrule
\multicolumn{6}{l}{\SLtwo{} frozen \ (base: \HRL{} $63.3$, \LRL{} $21.7$)} \\
12  & 0.98 & 64.8 & 63.0 & 22.5 & 21.6 \\
64  & 0.66 & 63.6 & 63.7 & 21.8 & 21.3 \\
128 & 0.50 & 62.0 & 61.2 & 21.2 & 20.4 \\
\midrule
\multicolumn{6}{l}{\SLtwo{} calibrated \ (base: \HRL{} $68.6$, \LRL{} $38.8$)} \\
12  & 0.70 & 68.2 & 68.5 & 38.8 & 38.4 \\
64  & 0.48 & 67.2 & 68.9 & 38.0 & 38.6 \\
128 & 0.42 & 66.4 & 67.5 & 37.6 & 37.4 \\
\bottomrule
\end{tabular}
\caption{INLP rank sweep on the projected text embedding.}
\label{tab:inlp-rank}
\end{table}

\subsection{Modality vs.\ Language Gap, Full Table}
\label{sec:app-extra-modality}

The text--image modality gap of \citet{liang2022modalitygap} provides a natural scale against which to read the cross-lingual gap of Section~\ref{sec:analysis}. Table~\ref{tab:gaps} reports the modality gap $\Delta_{\mathrm{mod}}$ and the within-\HRL{} and cross-tier (\HRL{}--\LRL{}) averaged language gaps as fractions of $\Delta_{\mathrm{mod}}$ for both encoders.

\begin{table}[h]
\centering
\small
\setlength{\tabcolsep}{4pt}
\begin{tabular}{lccc}
\toprule
& $\Delta_{\mathrm{mod}}$ & $\frac{\Delta_{\HRL\textrm{-}\HRL}}{\Delta_{\mathrm{mod}}}$ & $\frac{\Delta_{\HRL\textrm{-}\LRL}}{\Delta_{\mathrm{mod}}}$ \\
\midrule
\MCtwo{} & 0.855 & 30.4\% & \textbf{36.7\%} \\
& & {\scriptsize[29.9, 31.1]} & {\scriptsize[36.1, 37.3]} \\
\SLtwo{} & 1.121 & 18.1\% & \textbf{25.0\%} \\
& & {\scriptsize[17.5, 18.7]} & {\scriptsize[24.1, 25.9]} \\
\bottomrule
\end{tabular}
\caption{Modality gap (cosine distance) and within-\HRL{} / cross-tier \HRL{}--\LRL{} averaged language gaps as fractions of the modality gap on 500 XM3600 images. Brackets give $95\%$ bootstrap CIs from $1{,}000$ resamples of the image pool. The tier ordering \HRL{}--\HRL{} $<$ \HRL{}--\LRL{} is preserved with non-overlapping CIs in both architectures. Summarised in Appendix~\ref{sec:app-cka}.}
\label{tab:gaps}
\end{table}

\subsection{Trunk training-pool composition: \HRL{}-only trunk damages \LRL{}}
\label{sec:app-extra-hrlpool}

The canonical calibrated state trains the front-layer trunk on the 11-language pool (\HRL{}-6 + \LRL{}-5) with the 11-language centroid as anchor. A natural alternative is to use only the 6 \HRL{} languages for both supervision and anchor---arguably cheaper to obtain, and an interesting probe of whether \LRL{} retrieval gains come from \LRL{} exposure or merely from \HRL{}-centroid pull.

\begin{table}[h]
\centering
\small
\setlength{\tabcolsep}{3pt}
\resizebox{\columnwidth}{!}{%
\begin{tabular}{lccc}
\toprule
Training pool & \HRL{} \Rat{} & \LRL{} \Rat{} & \LRL{} pc \\
\midrule
Frozen (no trunk)                & 0.797 & 0.381 & --- \\
\HRL{}-only trunk (6 langs)      & 0.799 & \textbf{0.16} & 0.045 \\
\textbf{11-lang trunk (canon.)}  & 0.836 & \textbf{0.494} & 0.075 \\
\bottomrule
\end{tabular}%
}
\caption{Trunk training-pool ablation on \SLtwo{} SO400M. Training the trunk on only the six \HRL{} languages preserves \HRL{} retrieval but \emph{damages} \LRL{} retrieval below the frozen baseline ($0.381 \to 0.16$). The 11-language pool---which exposes the trunk to \LRL{} supervision---is essential for \LRL{} gains. \LRL{} retrieval is therefore not a side-effect of \HRL{}-centroid alignment alone; it requires the front-layer trunk to encode parallel-translation structure for all trained languages.}
\label{tab:pool-ablation}
\end{table}

The \HRL{}-only trunk's \LRL{} collapse rules out the ``just pull everything toward English'' reading: reducing \LRL{} residuals and improving \LRL{} retrieval requires \LRL{} parallel data during training. Equation~\ref{eq:cal-bound} characterises the pooled-row component once a state is produced; the ablation shows that producing a corrected \LRL{} state requires \LRL{} supervision, since the learned correction does not automatically transfer to distributions the trunk was not trained to shape.

\subsection{Cross-architecture hard re-init catastrophe}
\label{sec:app-extra-hardinit}

Hard re-init---drawing the front $M$ blocks from a fresh random init rather than the pretrained weights---underperforms soft fine-tune by 13--20\,pp on \MCtwo{} (Section~\ref{sec:approx}, Appendix~\ref{sec:app-sweep}). On bidirectional-attention encoders the gap is far worse.

\begin{table}[h]
\centering
\small
\setlength{\tabcolsep}{3pt}
\begin{tabular}{llcc}
\toprule
encoder (attn) & regime & \HRL{} & \LRL{} \\
\midrule
\MCtwo{} (causal)         & hard, $\lambda{=}3$       & 0.778 & 0.501 \\
\MCtwo{} (causal)         & \textbf{soft}             & \textbf{0.908} & \textbf{0.706} \\
\midrule
\SLtwo{} base (bidir.)    & hard                      & 0.018 & 0.008 \\
\SLtwo{} base (bidir.)    & \textbf{soft}             & \textbf{0.687} & \textbf{0.275} \\
\midrule
\SLtwo{} SO400M (bidir.)  & hard, M/N$=.33$           & 0.018 & 0.008 \\
\SLtwo{} SO400M (bidir.)  & hard, M/N$=.11$           & 0.009 & 0.004 \\
\SLtwo{} SO400M (bidir.)  & \textbf{soft}             & \textbf{0.833} & \textbf{0.608} \\
\bottomrule
\end{tabular}
\caption{Hard vs.\ soft re-init outcome across attention regimes (\Rat{}). Causal-attention \MCtwo{} tolerates hard re-init with partial degradation; bidirectional-attention \SLtwo{} (both base and SO400M) collapses to near-zero retrieval regardless of $M$. Reading: under bidirectional attention a re-initialised front emits out-of-distribution hidden states the fixed back-half cannot recover; under causal attention late tokens (including pooled EOS) partially route around the perturbation. Soft fine-tune is therefore the only viable regime on \SLtwo{}.}
\label{tab:hardinit}
\end{table}

\subsection{Residual cross-lingual redundancy survives trunk calibration}
\label{sec:app-extra-avgresidual}

Section~\ref{sec:hrl-avg} reports that averaging the six \HRL{} embeddings of a parallel sentence lifts \MCtwo{} \HRL{} \Rat{} from $78.1\%$ to $95.5\%$ on XM3600. A natural question is whether this averaging benefit \emph{disappears} after calibration---i.e., whether the trunk's job is to remove the cross-lingual content redundancy that averaging exploits.

\begin{table}[h]
\centering
\small
\setlength{\tabcolsep}{6pt}
\begin{tabular}{lcc}
\toprule
& Mean per-language & \HRL{}-averaged \\
state & \HRL{}-6 \Rat{} & \Rat{} \\
\midrule
Frozen      & 78.1 & 95.5 ($+17.4$) \\
Calibrated  & 77.1 & 94.9 ($+17.8$) \\
\bottomrule
\end{tabular}
\caption{\MCtwo{} on XM3600 captions: \HRL{}-6 per-language mean \Rat{} and the corresponding \HRL{}-averaged \Rat{}, before and after calibration. Both rows come from the same averaging run, whose calibrated state is an $M{=}4$ sibling of the canonical trunk, so the two are directly comparable. The +17 pp averaging benefit is essentially unchanged by calibration. Calibration leaves the gap between the two bars essentially unchanged rather than collapsing them onto each other, despite modestly lower absolute values in this sibling run: the cross-lingual content redundancy that averaging exploits is not removed by calibration but is preserved alongside the alignment-relevant signal, consistent with the erasure--retrieval dissociation of Section~\ref{sec:orth}.}
\label{tab:avg-residual}
\end{table}

The averaging benefit's near-invariance under calibration is the same observation viewed from a different angle as the LEACE result: a linear cross-lingual structure exists in the projected embedding, it is not what carries alignment, and the trunk is not optimising it away. The trunk intervention of Section~\ref{sec:approx} acts on alignment-relevant structure, leaving the averaging-relevant content redundancy approximately intact.

\subsection{Cross-lingual averaging across architectures and benchmarks}
\label{sec:app-extra-avg-broad}

The \HRL{}-averaging boost is not specific to \MCtwo{} or XM3600. Table~\ref{tab:avg-broad} extends the probe to five frozen multilingual vision-language encoders---\MCtwo{}, \SLtwo{}, AltCLIP, NLLB-CLIP-L, mSigLIP (Appendix~\ref{sec:app-morevlm})---on XM3600 and Flickr30k-200, comparing the per-language mean against a pool centroid (\HRL{}-6 or \LRL{}-5). Embeddings are L2-normalised before averaging and the centroid is renormalised before retrieval; the gallery is the standard $1{,}000$-image subset.

\begin{table*}[t]
\centering
\small
\setlength{\tabcolsep}{4pt}
\begin{tabular}{l|ccc|ccc|ccc|ccc}
\toprule
& \multicolumn{6}{c|}{XM3600 i2t \Rat{} (\%)} & \multicolumn{6}{c}{Flickr30k-200 i2t \Rat{} (\%)} \\
& \multicolumn{3}{c|}{\HRL{}-6 pool} & \multicolumn{3}{c|}{\LRL{}-5 pool} & \multicolumn{3}{c|}{\HRL{}-6 pool} & \multicolumn{3}{c}{\LRL{}-5 pool} \\
Encoder & mean & centroid & $\Delta$ & mean & centroid & $\Delta$ & mean & centroid & $\Delta$ & mean & centroid & $\Delta$ \\
\midrule
\MCtwo{}        & 78.1 & 95.3 & $+17.2$ & 44.1 & 77.8 & $\mathbf{+33.7}$ & 78.3 & 83.9 & $+5.6$  & 60.4 & 76.9 & $\mathbf{+16.5}$ \\
\SLtwo{}        & 63.3 & 93.3 & $+30.0$ & 21.7 & 55.0 & $\mathbf{+33.3}$ & 63.7 & 86.2 & $+22.5$ & 25.3 & 48.3 & $\mathbf{+23.0}$ \\
\midrule
AltCLIP         & 49.2 & 87.7 & $+38.5$ &  6.6 & 18.5 & $+11.9$         & 48.4 & 73.2 & $+24.8$ &  9.1 & 19.5 & $+10.4$ \\
NLLB-CLIP-L     & 55.4 & 87.8 & $+32.4$ & 38.5 & 73.1 & $\mathbf{+34.6}$ & 57.3 & 63.0 & $+5.7$  & 48.7 & 58.9 & $\mathbf{+10.2}$ \\
mSigLIP         & 73.5 & 95.5 & $+22.0$ & 32.4 & 62.0 & $\mathbf{+29.6}$ & 68.1 & 79.4 & $+11.3$ & 29.6 & 46.6 & $\mathbf{+17.0}$ \\
\bottomrule
\end{tabular}
\caption{Per-tier mean i2t \Rat{} on XM3600 / Flickr30k-200 for the per-language baseline (``mean'') vs.\ the pool-centroid embedding (``centroid''), at both pool choices (\HRL{}-6 and \LRL{}-5). All five encoders are frozen. \HRL{}-averaging lifts retrieval $+5.7$ to $+38.5$\,pp in every cell; \LRL{}-averaging is also positive in every cell (two cells clear $+33$\,pp), so cross-lingual content redundancy is not exclusive to the \HRL{} pool. AltCLIP's smaller \LRL{}-centroid lift reflects its weak frozen \LRL{} baseline ($6.6$ / $9.1$). The \MCtwo{} row comes from the independent cross-encoder replication run used for this table; its $0.2$ pp difference from Table~\ref{tab:avg-residual} reflects that separate evaluation artifact.}
\label{tab:avg-broad}
\end{table*}

The \LRL{}-pool centroid is the more striking result: averaging five individually weak \LRL{} embeddings lifts \LRL{} retrieval by $\geq +10$\,pp in every (encoder, benchmark) cell, and above 55\% on XM3600 \LRL{} for four of five encoders (\MCtwo{}~77.8, \SLtwo{}~55.0, NLLB-CLIP-L~73.1, mSigLIP~62.0; AltCLIP lags at $18.5$, matching its weak frozen baseline). This rules out reading \HRL{}-averaging as ``mix in the strong languages until retrieval recovers'': averaging \emph{within} a weak tier also extracts substantial alignment-relevant content, so cross-lingual redundancy is broader than \HRL{}-vs-\LRL{}.

\subsection{Data prevalence vs.\ front-half divergence}
\label{sec:app-prevalence}

Table~\ref{tab:prevalence} pairs each trained language's pretraining prevalence, as reported in Table~12 of \citet{wang2025scaling}, with its frozen depth-$M$ residual $r(L)$ (Equation~\ref{eq:rL-def}, per-language values in Table~\ref{tab:rL-residual}). Rank correlation is negative on both encoders: Spearman $-0.88$ ($p{=}0.002$) on \MCtwo{} and $-0.64$ ($p{=}0.054$) on \SLtwo{} over the ten non-English languages, using two-sided exact permutation tests over all $10!$ label permutations. The last-block trajectory divergence $d_{N-1}(L)$ gives the same sign ($-0.88$ / $-0.79$).

This is corroborating context, not a second finding. With $n{=}10$ a rank correlation carries wide uncertainty; prevalence co-varies with corpus quality, tokeniser coverage and script across this pool; and unlike the interventions of Sections~\ref{sec:orth}--\ref{sec:causal} nothing here is manipulated, so it cannot license a causal reading of pretraining data on divergence.

\begin{table}[h]
\centering
\small
\setlength{\tabcolsep}{4pt}
\begin{tabular}{lc|cc}
\toprule
& & \multicolumn{2}{c}{$r(L)$, frozen} \\
Lang & prevalence (\%) & \MCtwo{} & \SLtwo{} \\
\midrule
en  & $35.353$ & --- & --- \\
es  & \phantom{0}$8.214$ & 7.08 & \phantom{0}6.54 \\
de  & \phantom{0}$3.869$ & 7.50 & \phantom{0}6.63 \\
zh  & \phantom{0}$3.544$ & 8.67 & \phantom{0}8.10 \\
fr  & \phantom{0}$3.354$ & 7.02 & \phantom{0}6.70 \\
ko  & \phantom{0}$2.519$ & 8.76 & \phantom{0}8.46 \\
\midrule
hi  & \phantom{0}$0.267$ & 9.22 & \phantom{0}8.82 \\
bn  & \phantom{0}$0.113$ & 9.18 & 10.49 \\
fil & \phantom{0}$0.111$ & 9.02 & \phantom{0}6.89 \\
sw  & \phantom{0}$0.046$ & 9.58 & \phantom{0}7.21 \\
te  & \phantom{0}$0.036$ & 9.70 & \phantom{0}9.55 \\
\midrule
\multicolumn{2}{l|}{\textit{Spearman} $\rho$} & $-0.88$ & $-0.64$ \\
\bottomrule
\end{tabular}
\caption{Pretraining prevalence (percentage of pages in WebLI-100B, Table~12 of \citealp{wang2025scaling}) against the frozen depth-$M$ residual $r(L)$; rows ordered by prevalence. That table designates seven low-resource languages spanning $0.001$--$0.267\%$ (Section~\ref{sec:setup}), five of which are in our trained pool; it classes Korean as high-resource. English is the reference for $r(L)$ and is excluded from the correlation ($n{=}10$).}
\label{tab:prevalence}
\end{table}

\subsection{Geometry of the cross-lingual averaging boost}
\label{sec:app-avg-geometry}

Appendices~\ref{sec:app-extra-avgresidual} and~\ref{sec:app-extra-avg-broad} establish that pool averaging lifts retrieval and that the lift survives calibration. Table~\ref{tab:avg-geometry} reports the geometry behind it: the pre-normalisation norm of individual projected embeddings, the norm of the pool centroid of L2-normalised embeddings (a dispersion measure --- $1$ would mean the pool is collinear), and the shift in text--image cosine from per-language embeddings to the renormalised centroid.

Two quantities explain the boost. In the frozen state $\lVert c\rVert$ is $0.77$--$0.88$: the pooled languages disagree enough that averaging cancels language-specific components while retaining shared content. And the centroid's cosine to the image exceeds the mean per-language cosine in every cell --- $+0.063$ (\HRL{}) and $+0.059$ (\LRL{}) on frozen \MCtwo{}, $+0.027$ and $+0.010$ on frozen \SLtwo{} --- which is the geometric form of the retrieval lift. After calibration $\lVert c\rVert$ rises ($0.796 \to 0.877$ on \MCtwo{} \HRL{}, $0.812 \to 0.916$ on \SLtwo{} \HRL{}), i.e.\ the pool is more concentrated, and the centroid's advantage shrinks correspondingly ($+0.063 \to +0.036$; $+0.027 \to +0.011$) while absolute per-language similarity rises. Calibration moves the languages toward each other rather than removing the redundancy averaging exploits.

\begin{table}[h]
\centering
\small
\setlength{\tabcolsep}{3pt}
\resizebox{\columnwidth}{!}{%
\begin{tabular}{llcccc}
\toprule
cell & pool & $\lVert t_L\rVert$ raw & $\lVert c\rVert$ & $\cos(\hat c, v)$ & $\Delta$ vs.\ per-lang \\
\midrule
\MCtwo{} fr  & \HRL{}-6 & $25.1 \pm 0.9$ & $0.796$ & $0.309$ & $+0.063$ \\
             & \LRL{}-5 & $26.1 \pm 2.4$ & $0.773$ & $0.259$ & $+0.059$ \\
\MCtwo{} cal & \HRL{}-6 & $24.2 \pm 0.5$ & $0.877$ & $0.290$ & $+0.036$ \\
             & \LRL{}-5 & $24.9 \pm 1.3$ & $0.865$ & $0.258$ & $+0.035$ \\
\midrule
\SLtwo{} fr  & \HRL{}-6 & $33.0 \pm 3.1$ & $0.812$ & $0.144$ & $+0.027$ \\
             & \LRL{}-5 & $35.6 \pm 2.0$ & $0.881$ & $0.079$ & $+0.010$ \\
\SLtwo{} cal & \HRL{}-6 & $34.1 \pm 2.3$ & $0.916$ & $0.136$ & $+0.011$ \\
             & \LRL{}-5 & $32.2 \pm 1.0$ & $0.896$ & $0.105$ & $+0.011$ \\
\bottomrule
\end{tabular}%
}
\caption{Geometry of pool averaging on the $1{,}000$-image XM3600 subset. $\lVert t_L\rVert$ raw: mean $\pm$ sample s.d.\ across the pool's languages of the per-language mean pre-normalisation norm. $\lVert c\rVert$: mean norm of the per-row pool centroid of L2-normalised embeddings. $\cos(\hat c, v)$: cosine between the renormalised centroid and the paired image embedding. $\Delta$: that cosine minus the mean per-language text--image cosine.}
\label{tab:avg-geometry}
\end{table}

\subsection{How front-layer calibration reshapes the later trajectory}
\label{sec:app-traj-div}

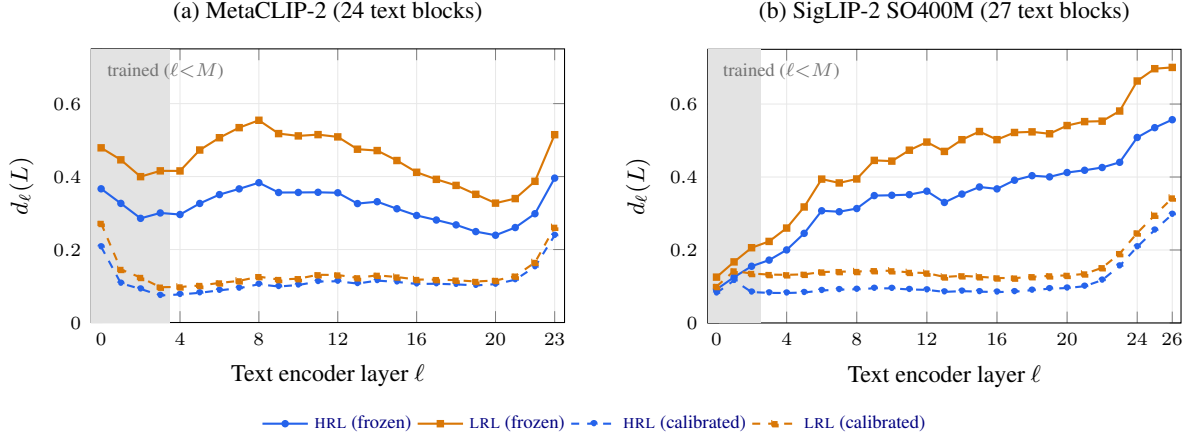
\begin{figure*}[t]
\centering
\begin{minipage}{0.49\textwidth}\centering
\begin{tikzpicture}
\begin{axis}[
  width=\textwidth, height=5.2cm,
  xlabel={Text encoder layer $\ell$}, ylabel={$d_\ell(L)$},
  xmin=-0.5, xmax=23.5, ymin=0.0, ymax=0.75,
  xtick={0,4,8,12,16,20,23}, ytick={0,0.2,0.4,0.6},
  legend to name=trajdiv-legend,
  legend style={font=\scriptsize, draw=none, fill=none, inner sep=2pt},
  legend columns=4, /tikz/every even column/.append style={column sep=8pt},
  tick label style={font=\scriptsize}, label style={font=\footnotesize},
  title style={font=\footnotesize}, title={(a) \MCtwo{} (24 text blocks)},
  grid=both, grid style={gray!18, very thin}, axis on top=false,
]
\fill[gray!22] (axis cs:-0.5,0.0) rectangle (axis cs:3.5,0.75);
\node[anchor=north west, font=\scriptsize, text=gray!90!black] at (axis cs:-0.2,0.74) {trained ($\ell{<}M$)};
\addplot[color=tHRL, mark=*, mark size=1pt, thick] coordinates {(0,0.3665)(1,0.3265)(2,0.2855)(3,0.3003)(4,0.2960)(5,0.3263)(6,0.3506)(7,0.3663)(8,0.3832)(9,0.3563)(10,0.3565)(11,0.3569)(12,0.3556)(13,0.3260)(14,0.3311)(15,0.3118)(16,0.2934)(17,0.2809)(18,0.2676)(19,0.2492)(20,0.2392)(21,0.2603)(22,0.2982)(23,0.3958)};
\addlegendentry{\HRL{} (frozen)}
\addplot[color=tLRL, mark=square*, mark size=1pt, thick] coordinates {(0,0.4788)(1,0.4459)(2,0.3998)(3,0.4157)(4,0.4155)(5,0.4728)(6,0.5065)(7,0.5341)(8,0.5543)(9,0.5178)(10,0.5115)(11,0.5150)(12,0.5088)(13,0.4750)(14,0.4713)(15,0.4440)(16,0.4118)(17,0.3924)(18,0.3758)(19,0.3514)(20,0.3272)(21,0.3397)(22,0.3870)(23,0.5148)};
\addlegendentry{\LRL{} (frozen)}
\addplot[color=tHRL, mark=*, mark size=1pt, thick, dashed] coordinates {(0,0.2075)(1,0.1077)(2,0.0920)(3,0.0742)(4,0.0769)(5,0.0814)(6,0.0882)(7,0.0941)(8,0.1048)(9,0.0986)(10,0.1028)(11,0.1123)(12,0.1131)(13,0.1067)(14,0.1149)(15,0.1112)(16,0.1057)(17,0.1062)(18,0.1045)(19,0.1018)(20,0.1050)(21,0.1173)(22,0.1528)(23,0.2388)};
\addlegendentry{\HRL{} (calibrated)}
\addplot[color=tLRL, mark=square*, mark size=1pt, thick, dashed] coordinates {(0,0.2715)(1,0.1454)(2,0.1234)(3,0.0968)(4,0.0975)(5,0.1014)(6,0.1084)(7,0.1141)(8,0.1257)(9,0.1177)(10,0.1201)(11,0.1308)(12,0.1302)(13,0.1224)(14,0.1293)(15,0.1246)(16,0.1178)(17,0.1176)(18,0.1159)(19,0.1124)(20,0.1152)(21,0.1270)(22,0.1646)(23,0.2609)};
\addlegendentry{\LRL{} (calibrated)}
\end{axis}\end{tikzpicture}
\end{minipage}\hfill
\begin{minipage}{0.49\textwidth}\centering
\begin{tikzpicture}
\begin{axis}[
  width=\textwidth, height=5.2cm,
  xlabel={Text encoder layer $\ell$}, ylabel={$d_\ell(L)$},
  xmin=-0.5, xmax=26.5, ymin=0.0, ymax=0.75,
  xtick={0,4,8,12,16,20,24,26}, ytick={0,0.2,0.4,0.6},
  tick label style={font=\scriptsize}, label style={font=\footnotesize},
  title style={font=\footnotesize}, title={(b) \SLtwo{} SO400M (27 text blocks)},
  grid=both, grid style={gray!18, very thin}, axis on top=false,
]
\fill[gray!22] (axis cs:-0.5,0.0) rectangle (axis cs:2.5,0.75);
\node[anchor=north west, font=\scriptsize, text=gray!90!black] at (axis cs:-0.2,0.74) {trained ($\ell{<}M$)};
\addplot[color=tHRL, mark=*, mark size=1pt, thick] coordinates {(0,0.0920)(1,0.1255)(2,0.1554)(3,0.1723)(4,0.2002)(5,0.2456)(6,0.3075)(7,0.3049)(8,0.3136)(9,0.3491)(10,0.3500)(11,0.3517)(12,0.3612)(13,0.3303)(14,0.3532)(15,0.3727)(16,0.3675)(17,0.3912)(18,0.4037)(19,0.4004)(20,0.4123)(21,0.4185)(22,0.4261)(23,0.4403)(24,0.5084)(25,0.5353)(26,0.5571)};
\addplot[color=tLRL, mark=square*, mark size=1pt, thick] coordinates {(0,0.1256)(1,0.1674)(2,0.2062)(3,0.2235)(4,0.2601)(5,0.3179)(6,0.3943)(7,0.3838)(8,0.3947)(9,0.4456)(10,0.4436)(11,0.4735)(12,0.4958)(13,0.4702)(14,0.5023)(15,0.5247)(16,0.5024)(17,0.5221)(18,0.5238)(19,0.5186)(20,0.5411)(21,0.5519)(22,0.5531)(23,0.5809)(24,0.6632)(25,0.6969)(26,0.7004)};
\addplot[color=tHRL, mark=*, mark size=1pt, thick, dashed] coordinates {(0,0.0822)(1,0.1166)(2,0.0847)(3,0.0824)(4,0.0821)(5,0.0835)(6,0.0890)(7,0.0914)(8,0.0923)(9,0.0950)(10,0.0948)(11,0.0918)(12,0.0907)(13,0.0852)(14,0.0872)(15,0.0858)(16,0.0847)(17,0.0858)(18,0.0898)(19,0.0940)(20,0.0957)(21,0.1007)(22,0.1174)(23,0.1564)(24,0.2088)(25,0.2546)(26,0.2978)};
\addplot[color=tLRL, mark=square*, mark size=1pt, thick, dashed] coordinates {(0,0.0999)(1,0.1425)(2,0.1355)(3,0.1326)(4,0.1319)(5,0.1332)(6,0.1396)(7,0.1407)(8,0.1409)(9,0.1432)(10,0.1429)(11,0.1385)(12,0.1367)(13,0.1261)(14,0.1291)(15,0.1265)(16,0.1233)(17,0.1229)(18,0.1267)(19,0.1290)(20,0.1303)(21,0.1343)(22,0.1513)(23,0.1900)(24,0.2466)(25,0.2952)(26,0.3425)};
\end{axis}\end{tikzpicture}
\end{minipage}

\vspace{2pt}
\centerline{\ref{trajdiv-legend}}

\caption{Per-layer forward-path divergence $d_\ell(L)$ (Equation~\ref{eq:dL-def}) at the EOS position, on $1{,}000$ image-aligned XM3600 captions and the 11 trained languages. Solid: frozen; dashed: trunk-calibrated. Curves are tier means over non-English \HRL{} ($n{=}5$) and \LRL{}-5. Shaded: the trunk-trained front blocks ($M{=}4$ on \MCtwo{}, $M{=}3$ on \SLtwo{}). Axes and shading match Figure~\ref{fig:cka-frozen} for side-by-side reading; no retraining is involved, only a forward pass of the frozen and calibrated encoders.}
\label{fig:traj-div}
\end{figure*}

Section~\ref{sec:analysis} points here for the question the trunk is meant to answer: only the first $M$ blocks are trained, so what happens to the $N-M$ blocks that follow? Four measurements already in this paper answer it from different directions, and Figure~\ref{fig:traj-div} adds the divergence measure itself.

\textbf{(i)~Per-layer similarity.} Cross-lingual CKA rises at the projector input, twenty blocks past the trained region (Figure~\ref{fig:cka-frozen}): $0.458 \to 0.556$ on \MCtwo{}, $0.302 \to 0.451$ on \SLtwo{}. \textbf{(ii)~Front-half residual.} $r(L)$ shows what the trunk does at its own depth (Table~\ref{tab:rL-residual}); the hidden-state norm roughly doubles under calibration, so the informative quantity is $\bar\rho_L \cdot r(L)$ rather than $r(L)$ alone. \textbf{(iii)~Back-half sensitivity.} The frozen back-half responds unevenly across the frozen manifold (sampled ratios up to $0.81$ for \MCtwo{} sw) but evenly on the calibrated one ($[0.27, 0.33]$ for all eleven languages, Table~\ref{tab:lipschitz-perlang}) --- the trunk moves states into a region the frozen back-half handles more stably. \textbf{(iv)~Divergence.} Figure~\ref{fig:traj-div} plots $d_\ell(L)$ (Equation~\ref{eq:dL-def}) at every block. In the frozen state the \LRL{} curve lies above the \HRL{} curve at every depth on both encoders, with a last-block tier gap of $0.119$ (\MCtwo{}) and $0.143$ (\SLtwo{}); after calibration the \LRL{} divergence falls from $0.515$ to $0.261$ and from $0.700$ to $0.343$, and the tier gap to $0.022$ and $0.045$. Per language, every trained language ends inside $[0.22, 0.29]$ on \MCtwo{} and $[0.25, 0.37]$ on \SLtwo{}, against frozen spreads of $[0.36, 0.62]$ and $[0.48, 0.77]$.

Together these say the trunk does not merely shift its own output: it moves depth-$M$ states into a region whose image under the frozen back-half is more cross-lingually aligned. The recovery is partial, and the paper is explicit about the reference points. Calibration recovers $+9.6$ / $+17.1$\,pp on \LRL{} (Table~\ref{tab:perlang-xm3600}) where the deep single-row patch reaches $73.4\%$ from a $44.1\%$ baseline on \MCtwo{} \LRL{}, and a multi-\HRL{} centroid at the same depth exceeds even the single-source English reference (Appendix~\ref{sec:app-r1-multi-hrl}). A same-depth single-row patch at $\ell{=}M$, conversely, recovers far less than the trunk (Appendix~\ref{sec:app-r1-patch-at-M}). Calibration and EOS-swap patching therefore act on the same trajectory-level object and produce convergent geometric changes, calibration on general input rather than an oracle parallel caption. The two remain distinct interventions---the same-depth patch is much weaker, the deep patch stronger---rather than one being a parameterised form of, or a proven bound on, the other.

\section{Sampled Sensitivity of the Frozen Back-Half}
\label{sec:app-lipschitz}

The envelope of Equation~\ref{eq:cal-bound} is useful only if the back-half's response to a change at depth $M$ is numerically small. This appendix reports two measurements per (model, state) cell on XM3600 captions, using the same image-aligned $1{,}000$-caption pool as the headline retrieval experiments: an architecture-only spectral upper bound, and a sampled sensitivity probe along the pooled coordinate. Only the former is a bound; the latter is a diagnostic, for the reasons given at the end of this appendix.

\paragraph{Block-wise spectral-norm upper bound.} Each transformer block in $\psi$ combines $\{q, k, v, \mathrm{out}, \mathrm{fc1}, \mathrm{fc2}\}$ projections with LayerNorms and residual connections. We bound each block by products of spectral norms, then multiply across the back-half. We report a naive ``all-six'' product and a residual-bound variant replacing each block product with $1 + \mathrm{prod}_{\mathrm{attn}} + \mathrm{prod}_{\mathrm{mlp}}$; Table~\ref{tab:lipschitz} gives the resulting numbers.

\paragraph{Sampled pooled-row sensitivity.} For each model state, $1{,}000$ English XM3600 captions yield $h_M(x^{\mathrm{en}}_i) \in \mathbb{R}^d$ for $i = 1{..}1000$. Each $h_M$ is perturbed by norm-controlled Gaussian noise $\delta_{i,k} \sim \mathcal{N}(0, \sigma_k^2 I)$ at six scales $\sigma_k$ from $0.001$ to $0.5$ times $\lVert h_M(x^{\mathrm{en}}_i) \rVert$, with three draws per scale. Because the perturbation moves only the pooled row, we take the ratio through the conditional map
\begin{equation*}
  \psi_{\neg p}(h) \;:=\; \psi\bigl(H_{M,\neg p},\, h\bigr),
\end{equation*}
\noindent which holds the non-pooled rows $H_{M,\neg p}$ fixed, and compute
\begin{equation}
  \rho_{i,k} \;:=\; \frac{\lVert \psi_{\neg p}(h_M + \delta_{i,k}) - \psi_{\neg p}(h_M) \rVert}{\lVert \delta_{i,k} \rVert}
  \label{eq:rho-ratio}
\end{equation}
\noindent and aggregate over $1{,}000 \cdot 18 = 18{,}000$ probe pairs as mean, $\mathrm{p95}$, and maximum. As a perturbation of the full matrix this is supported on the pooled row alone---$\Delta_{i,k}[j,:] = \delta_{i,k}$ for $j = p$ and $0$ otherwise, so $\lVert \Delta_{i,k} \rVert_F = \lVert \delta_{i,k} \rVert$. A pooled-row-only change is exactly the intervention Section~\ref{sec:patching} performs, so this is the matched probe for the patching result; trunk calibration additionally moves the non-pooled rows, which is why Equation~\ref{eq:cal-bound} is an envelope on the pooled coordinate rather than on the full depth-$M$ change.

\begin{table}[h]
\centering
\small
\setlength{\tabcolsep}{6pt}
\begin{tabular}{lcc}
\toprule
cell & mean ratio & max ratio \\
\midrule
\MCtwo{} fr  & $\mathbf{0.46}$ & $4.56$ \\
\MCtwo{} cal & $\mathbf{0.28}$ & $0.76$ \\
\SLtwo{} fr  & $\mathbf{0.40}$ & $0.96$ \\
\SLtwo{} cal & $\mathbf{0.27}$ & $0.37$ \\
\bottomrule
\end{tabular}
\caption{Sampled pooled-row sensitivity ratios $\rho$ (Equation~\ref{eq:rho-ratio}) of the back-half for each (model, state): mean and maximum over $18{,}000$ Gaussian probe pairs at norm-controlled scales on English XM3600 captions. Spectral-norm upper bound (architecture-only): $3.6 \times 10^{50}$ (\MCtwo{}), $1.6 \times 10^{65}$ (\SLtwo{})---valid but uninformative. Observed means stay in $[0.27, 0.46]$ across all cells; under calibration the maximum observed ratio drops by $6.0{\times}$ on \MCtwo{} and $2.6{\times}$ on \SLtwo{}. These are sampled ratios, not bounds (see below).}
\label{tab:lipschitz}
\end{table}

\paragraph{What the sampled ratios do and do not establish.} Equation~\ref{eq:rho-ratio} is a finite-perturbation difference quotient at the sampled points, so neither its mean nor its maximum upper-bounds
\begin{equation}
  \mathrm{Lip}(\psi_{\neg p}; \mathcal{S}) \;=\; \sup_{h \neq h' \in \mathcal{S}} \frac{\lVert \psi_{\neg p}(h) - \psi_{\neg p}(h') \rVert}{\lVert h - h' \rVert} ,
  \label{eq:lip-sup}
\end{equation}
the constant Equation~\ref{eq:cal-bound} assumes, with $\mathcal{S}$ the trunk-produced states and their pooled-row counterfactuals: the sampled maximum is a lower bound on that supremum, and Gaussian probes at six norm scales cover only a thin slice of $\mathcal{S}$. The spectral product below is the only quantity here that is a genuine upper bound on Equation~\ref{eq:lip-sup}, and it is too loose to use. We therefore read Table~\ref{tab:lipschitz} and Table~\ref{tab:lipschitz-perlang} as evidence about how the back-half behaves in the region the trunk actually produces---and as a relative comparison between frozen and calibrated states---not as a certificate that Equation~\ref{eq:cal-bound} holds with $K$ at the tabulated value.

\paragraph{Why the spectral product is uninformative.} The naive bound treats every projection as independently maximal, ignoring cancellations across residual branches, the low-dimensional support of $h_M$, and inter-block LayerNorm renormalisation. The sampled probe instead measures the response of $\psi$ on in-distribution $h_M$ samples, which is the regime Equation~\ref{eq:cal-bound} is applied in.

\paragraph{Why the sampled ratios decrease under calibration.} The $h_M$ inputs to $\psi$ differ by state: frozen cells use pretrained mid-layer activations, whereas calibrated cells use trunk outputs shaped by the projected-centroid objective. Post-trunk states therefore occupy a tighter sub-region in which $\psi$ responds less to a pooled-row change. The mean observed ratio drops by $39\%$ on \MCtwo{} and $32\%$ on \SLtwo{}, and the maximum observed ratio drops by $6.0{\times}$ and $2.6{\times}$ respectively; both directions indicate this tightening and help explain why calibration yields stable retrieval gains rather than amplifying front-half perturbations.

\paragraph{Language-specific sensitivity.} The English-only probe of Table~\ref{tab:lipschitz} hides language variation in back-half contractivity. Repeating the sample-based probe with $h_M$ drawn from each of the 11 trained languages ($1{,}000$ XM3600 captions/lang $\times 18$ perturbation pairs) gives Table~\ref{tab:lipschitz-perlang}.

\begin{table}[h]
\centering
\small
\setlength{\tabcolsep}{3pt}
\begin{tabular}{lc|cc|cc}
\toprule
& & \multicolumn{2}{c|}{\MCtwo{}} & \multicolumn{2}{c}{\SLtwo{}} \\
lang & tier & frozen & cal. & frozen & cal. \\
\midrule
en  & \HRL{} & 0.46 & 0.28 & 0.40 & 0.27 \\
fr  & \HRL{} & 0.48 & 0.29 & 0.38 & 0.27 \\
de  & \HRL{} & 0.45 & 0.29 & 0.39 & 0.29 \\
es  & \HRL{} & 0.48 & 0.29 & 0.39 & 0.28 \\
zh  & \HRL{} & 0.60 & 0.30 & 0.48 & 0.29 \\
ko  & \HRL{} & 0.54 & 0.30 & 0.37 & 0.28 \\
\midrule
bn  & \LRL{} & 0.60 & 0.31 & 0.44 & 0.30 \\
fil & \LRL{} & \textbf{0.77} & 0.31 & 0.41 & 0.30 \\
hi  & \LRL{} & 0.63 & 0.32 & 0.44 & 0.29 \\
sw  & \LRL{} & \textbf{0.81} & 0.33 & 0.47 & 0.30 \\
te  & \LRL{} & 0.66 & 0.31 & 0.40 & 0.31 \\
\bottomrule
\end{tabular}
\caption{Mean sampled sensitivity ratio of the back-half, probed at language-specific $h_M$ regions ($1{,}000$ XM3600 captions/lang, $18{,}000$ Gaussian perturbation pairs/cell). Bold $> 0.75$. Frozen \MCtwo{} is tier-conditioned (\HRL{} $0.45$--$0.60$; trained \LRL{} reaches $0.77$ fil, $0.81$ sw); frozen \SLtwo{} is flatter ($0.37$--$0.48$, bidirectional attention smooths the response). Calibration pulls every probed language into $[0.27, 0.33]$, a narrower and less responsive operating point.}
\label{tab:lipschitz-perlang}
\end{table}

The frozen back-half does not respond uniformly across the $h_M$ manifold: trained \LRL{}s on \MCtwo{}, especially fil and sw, reach ratios $\approx 0.8$, so taking the EN-only value $\approx 0.5$ as $K$ across languages would understate the Equation~\ref{eq:cal-bound} envelope. Calibration pulls all 11 trained languages into a single contractive region: every calibrated cell lies in $[0.27, 0.33]$, roughly a $2\times$ tightening on the worst frozen-\LRL{} cells.

\paragraph{Front-half residual $r(L)$ per language.} The complementary quantity in the Equation~\ref{eq:cal-bound} bound is $r(L) := \mathbb{E}_x \lVert h_M(x^L) - h_M(x^{\mathrm{en}}) \rVert$, the distance from each language's mid-layer pooled state to the parallel English pooled state of the same encoder state. Table~\ref{tab:rL-residual} reports $r(L)$ on both encoders and states, computed on XM3600 captions aligned by image key.

\begin{table}[h]
\centering
\small
\setlength{\tabcolsep}{3pt}
\begin{tabular}{lc|cc|cc}
\toprule
& & \multicolumn{2}{c|}{\MCtwo{}} & \multicolumn{2}{c}{\SLtwo{}} \\
lang & tier & frozen & cal. & frozen & cal. \\
\midrule
en  & \HRL{} & 0.00 & 0.00 & 0.00 & 0.00 \\
fr  & \HRL{} & 7.02 & 7.88 & 6.70 & 10.44 \\
de  & \HRL{} & 7.50 & 7.47 & 6.63 & 15.46 \\
es  & \HRL{} & 7.09 & 7.57 & 6.54 & 10.82 \\
zh  & \HRL{} & 8.67 & 9.80 & 8.10 & 15.03 \\
ko  & \HRL{} & 8.76 & 9.21 & 8.46 & 13.57 \\
\midrule
bn  & \LRL{} & 9.18 & 9.43 & 10.49 & 15.85 \\
fil & \LRL{} & 9.02 & 10.66 & 6.89 & 15.74 \\
hi  & \LRL{} & 9.22 & 8.84 & 8.82 & 14.45 \\
sw  & \LRL{} & 9.58 & 11.19 & 7.21 & 17.09 \\
te  & \LRL{} & 9.70 & 8.04 & 9.55 & 19.17 \\
\midrule
\multicolumn{2}{l|}{\small mean $\lVert h_M\rVert$} & $\sim 10.2$ & $\sim 21.1$ & $\sim 13.6$ & $\sim 32.1$ \\
\bottomrule
\end{tabular}
\caption{Front-half residual $r(L) = \mathbb{E}_x \lVert h_M(x^L) - h_M(x^{\mathrm{en}}) \rVert$, each state measured against its own English reference so that $r(\mathrm{en}) = 0$, on both encoders ($M{=}4$ for \MCtwo{}, $M{=}3$ for \SLtwo{}; $1{,}000$ image-aligned XM3600 captions/lang). Frozen $r(L)$ sits in $\sim 7$--$10$ for \MCtwo{}, $\sim 7$--$10.5$ for \SLtwo{}; under calibration the hidden-state norm $\lVert h_M\rVert$ scales up ($\sim 10\to21$ on \MCtwo{}, $\sim 14\to32$ on \SLtwo{}), so the relevant scale is the product $\bar\rho_L \cdot r(L)$ rather than $r(L)$ alone (Table~\ref{tab:lipschitz-perlang} for $\bar\rho_L$).}
\label{tab:rL-residual}
\end{table}

After convergence, both the observed back-half response $\bar\rho_L$ on the trunk-output support and the relative residual $r(L)/\lVert h_M\rVert$ stay within a narrow range across the 11-language pool, so the calibrated $\bar\rho_L \cdot r(L)$ lies in $[2.2, 3.7]$ on \MCtwo{} and $[2.8, 6.0]$ on \SLtwo{}. This is where the Equation~\ref{eq:cal-bound} envelope is informative as an observed conditional-sensitivity scale rather than a certified bound.

\begin{figure*}[t]
\centering
\begin{tikzpicture}
\begin{axis}[
  width=0.95\textwidth, height=5cm,
  xlabel={$\bar\rho_L \cdot r(L)$},
  ylabel={Trunk $\Delta$ \Rat{} (pp)},
  ymin=-3, ymax=30,
  tick label style={font=\scriptsize}, label style={font=\footnotesize},
  title style={font=\footnotesize}, title={\MCtwo{} (Pearson $0.55$, Spearman $0.79$, $n{=}11$)},
  grid=both, grid style={gray!18, very thin}, axis on top,
]
\addplot[only marks, mark=*, color=tHRL, mark size=2pt] coordinates {
(0.00,2.6) (2.31,1.8) (2.17,0.3) (2.20,-0.4) (2.95,2.9) (2.77,3.4)
};
\addplot[only marks, mark=square*, color=tLRL, mark size=2pt] coordinates {
(2.90,4.3) (3.31,12.8) (2.79,3.5) (3.67,20.0) (2.46,7.7)
};
\node[anchor=west, font=\tiny] at (axis cs:0.00,2.6)  {\,en};
\node[anchor=west, font=\tiny] at (axis cs:2.31,1.8)  {\,fr};
\node[anchor=east, font=\tiny] at (axis cs:2.17,0.3)  {de\,};
\node[anchor=west, font=\tiny] at (axis cs:2.20,-0.4) {\,es};
\node[anchor=west, font=\tiny] at (axis cs:2.95,2.9)  {\,zh};
\node[anchor=east, font=\tiny] at (axis cs:2.77,3.4)  {ko\,};
\node[anchor=west, font=\tiny] at (axis cs:2.79,3.5)  {\,hi};
\node[anchor=west, font=\tiny] at (axis cs:2.90,4.3)  {\,bn};
\node[anchor=west, font=\tiny] at (axis cs:3.31,12.8) {\,fil};
\node[anchor=west, font=\tiny] at (axis cs:3.67,20.0) {\,sw};
\node[anchor=west, font=\tiny] at (axis cs:2.46,7.7)  {\,te};
\end{axis}
\end{tikzpicture}

\vspace{0.4em}
\begin{tikzpicture}
\begin{axis}[
  width=0.95\textwidth, height=5cm,
  xlabel={$\bar\rho_L \cdot r(L)$},
  ylabel={Trunk $\Delta$ \Rat{} (pp)},
  ymin=-3, ymax=30,
  tick label style={font=\scriptsize}, label style={font=\footnotesize},
  title style={font=\footnotesize}, title={\SLtwo{} (Pearson $0.69$, Spearman $0.91$, $n{=}11$)},
  grid=both, grid style={gray!18, very thin}, axis on top,
]
\addplot[only marks, mark=*, color=tHRL, mark size=2pt] coordinates {
(0.00,4.4) (2.83,4.4) (4.47,10.1) (2.97,3.3) (4.38,6.7) (3.84,3.0)
};
\addplot[only marks, mark=square*, color=tLRL, mark size=2pt] coordinates {
(4.67,12.5) (4.75,18.1) (4.17,4.9) (5.20,23.3) (5.95,26.7)
};
\node[anchor=west, font=\tiny] at (axis cs:0.00,4.4)  {\,en};
\node[anchor=west, font=\tiny] at (axis cs:2.83,4.4)  {\,fr};
\node[anchor=west, font=\tiny] at (axis cs:4.47,10.1) {\,de};
\node[anchor=west, font=\tiny] at (axis cs:2.97,3.3)  {\,es};
\node[anchor=south, font=\tiny] at (axis cs:4.38,6.7) {zh};
\node[anchor=west, font=\tiny] at (axis cs:3.84,3.0)  {\,ko};
\node[anchor=west, font=\tiny] at (axis cs:4.67,12.5) {\,bn};
\node[anchor=west, font=\tiny] at (axis cs:4.75,18.1) {\,fil};
\node[anchor=north, font=\tiny] at (axis cs:4.17,4.9) {hi};
\node[anchor=west, font=\tiny] at (axis cs:5.20,23.3) {\,sw};
\node[anchor=west, font=\tiny] at (axis cs:5.95,26.7) {\,te};
\end{axis}
\end{tikzpicture}
\caption{Per-language observed conditional-sensitivity scale $\bar\rho_L \cdot r(L)$ (calibrated state, XM3600) vs.\ trunk lift $\Delta := \mathrm{R@1}_{\mathrm{cal}} - \mathrm{R@1}_{\mathrm{frozen}}$ on XM3600, 11 trained languages per encoder. Blue circles: \HRL{}-6; orange squares: \LRL{}-5. The English point sits at $\bar\rho_{\mathrm{en}} \cdot r(\mathrm{en}) = 0$ by construction. The Spearman ranking is strong on both encoders, dominated by co-variation with baseline difficulty rather than independent predictive power.}
\label{fig:kr-delta-scatter}
\end{figure*}
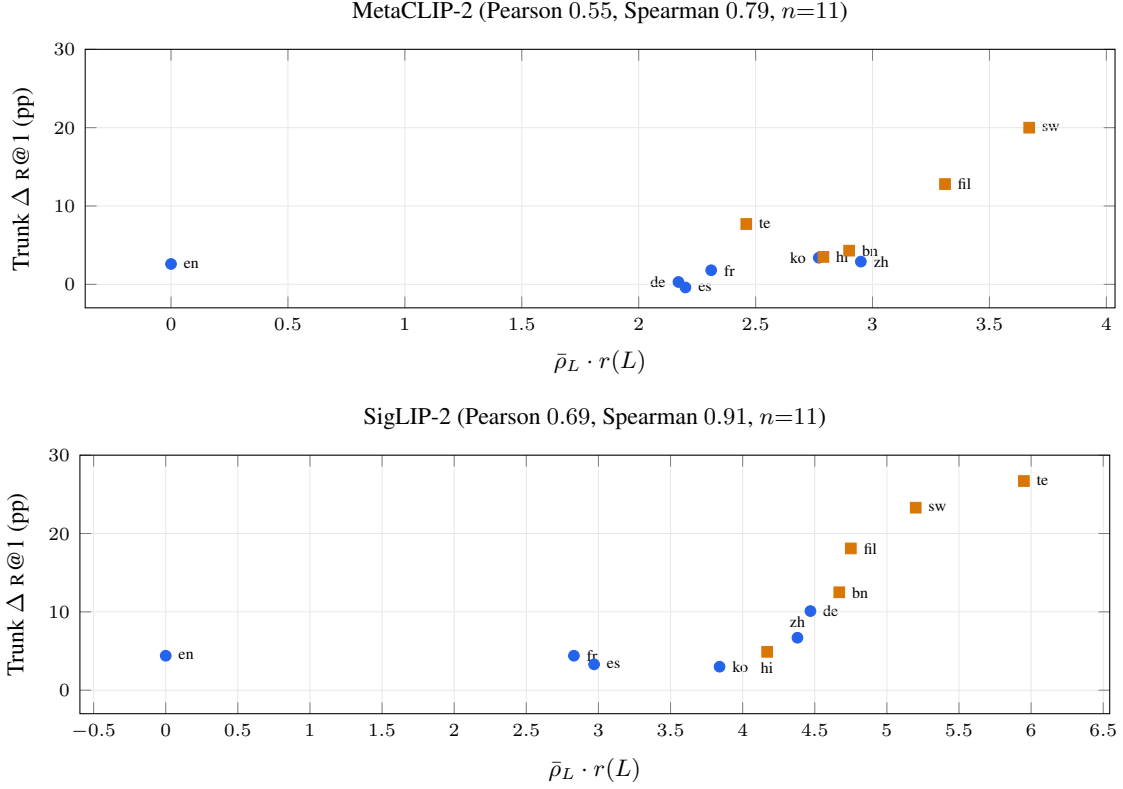

\paragraph{Predictiveness of $\bar\rho_L \cdot r(L)$ for trunk amenability.} Equation~\ref{eq:cal-bound} relates the projected-embedding residual to the pooled-row scale on the trunk's calibrated $h_M$ output; below we substitute the \emph{observed} mean ratio $\bar\rho_L$ (Table~\ref{tab:lipschitz-perlang}) for the assumed constant $K$. Figure~\ref{fig:kr-delta-scatter} compares this quantity with per-language calibration lift $\Delta := \mathrm{R@1}_{\mathrm{cal}} - \mathrm{R@1}_{\mathrm{frozen}}$ across the 11 trained languages on both encoders ($n{=}22$ pooled). Pearson is moderate (\MCtwo{} $\rho{=}0.55$, \SLtwo{} $\rho{=}0.69$, pooled $\rho{=}0.70$), while Spearman is strong (\MCtwo{} $0.79$, \SLtwo{} $0.91$, pooled $0.75$). The positive sign mainly reflects baseline difficulty: weak-baseline languages have larger frozen front-half divergence and more headroom for calibration. Thus, the observed conditional-sensitivity scale ranks trained \LRL{}s above \HRL{}s in amenability, but it does not sharply predict per-language lift.

\section{Additional Patching Controls and Bootstrap Confidence}
\label{sec:app-r1-controls}

This appendix reports six controls for the causal interpretation of Section~\ref{sec:patching}, Section~\ref{sec:approx}, and Section~\ref{sec:orth}: (a)~single-row patching at $\ell = M$ to distinguish calibration from a trunk-depth patch; (b)~non-EOS patching to separate pooling tautology from substantive rescue; (c)~random-English-caption controls, including the non-tautological $\ell = N{-}4$ variant; (d)~bootstrap confidence intervals; (e)~French-source and multi-\HRL{} patching to test English specificity; and (f)~a non-linear adversarial eraser probing whether LEACE's linear-orthogonality result extends beyond linear erasure.

\paragraph{What the patch replaces.} Writing $p_L$ for the pooled (EOS) position of a language-$L$ caption, $p_T$ for the target and $p_A$ for the source, the patched matrix of Section~\ref{sec:patching} is, row by row,
\begin{equation}
  \widetilde{H}_{\ell+1}(x^T; A)_{j,:} =
  \begin{cases}
    h_{\ell+1}(x^A), & j = p_T, \\[2pt]
    H_{\ell+1}(x^T)_{j,:}, & j \neq p_T,
  \end{cases}
  \label{eq:patch-rows}
\end{equation}
so exactly one row changes and every other token state stays at its target-language value. Because the two encoders pool differently (\MCtwo{} at a dynamic EOS token, \SLtwo{} at the sticky last position; Section~\ref{sec:setup}), $p_A$ and $p_T$ generally differ. Control~(b) below patches a $j \neq p_T$ row instead, isolating the pooled row's contribution.

\subsection{Patch at $\ell = M$: trunk calibration $\neq$ single-row patch at trunk depth}
\label{sec:app-r1-patch-at-M}

If calibration were simply a parametric form of $\patch_{M,\,\mathrm{en}\to L}$, patching the frozen encoder at depth $M$ should reproduce its gains. It does not (Table~\ref{tab:patch-at-M}): $\ell=M$ moves \LRL{} by only $-0.1$ to $+2.5$ pp across the four cells, and the best layer in $\{M{-}1,\ldots,M{+}4\}$ tops out at $+3.5$ pp, far below calibration's $+9.6$--$17.1$ pp. Calibration therefore appears to restructure the front-half across $M$ trained layers rather than mimic a single-row same-depth patch.

\begin{table}[h]
\centering
\small
\setlength{\tabcolsep}{3pt}
\begin{tabular}{lc|ccc}
\toprule
& & \HRL{} & \LRL{} & EN ref. \\
cell & $\ell$ & \Rat{} & \Rat{} & \\
\midrule
\MCtwo{} fr & baseline & 79.8 & 44.1 & 69.4 \\
& patch@$M{=}4$ & 81.0 & 45.5 & \\
& best in $\{3..8\}$ & 82.1 & 47.6 & \\
\midrule
\MCtwo{} cal & baseline & 81.4 & 53.7 & 72.0 \\
& patch@$M{=}4$ & 82.6 & 56.1 & \\
& best in $\{3..8\}$ & 83.0 & 56.9 & \\
\midrule
\SLtwo{} fr & baseline & 62.8 & 21.7 & 66.1 \\
& patch@$M{=}3$ & 63.2 & 21.6 & \\
& best in $\{2..7\}$ & 63.9 & 22.4 & \\
\midrule
\SLtwo{} cal & baseline & 68.3 & 38.8 & 70.5 \\
& patch@$M{=}3$ & 69.6 & 41.3 & \\
& best in $\{2..7\}$ & 70.6 & 42.3 & \\
\bottomrule
\end{tabular}
\caption{Patching the EOS row at calibration depth $M$ gives at most $+2.5$ pp on \LRL{}, far below deep-layer patching's $+29.3$ pp at $\ell = N{-}4$ (Table~\ref{tab:patch}); the best layer reaches $+30.2$ pp and the near-tautological $\ell = N{-}1$ endpoint $+25.3$ pp. The best layer in $\{M{-}1, \ldots, M{+}4\}$ reaches only $+3.5$ pp, so calibration is not a single-row inference-time patch at the same depth.}
\label{tab:patch-at-M}
\end{table}

\subsection{Non-EOS position patching: pooling tautology vs.\ deep-layer rescue}
\label{sec:app-r1-nonEOS}

Because the last-block EOS state is the projector input, an EOS-row patch at $\ell=N{-}1$ directly overwrites the pooled representation. Three \MCtwo{} frozen diagnostics separate this tautology from substantive rescue (Table~\ref{tab:nonEOS-controls}).

\begin{table}[h]
\centering
\small
\setlength{\tabcolsep}{3pt}
\begin{tabular}{llcc}
\toprule
& & \HRL{} & \LRL{} \\
$\ell$ & control & \Rat{} & \Rat{} \\
\midrule
& baseline & 79.8 & 44.1 \\
\midrule
$N{-}1=23$ & EOS-patch & 69.4 & 69.4 \\
$N{-}1=23$ & mid-token & 79.8 & 44.1 \\
\midrule
$N{-}4=20$ & EOS-patch & 85.8 & \textbf{73.4} \\
$N{-}4=20$ & mid-token & 81.0 & \textbf{49.6} \\
\bottomrule
\end{tabular}
\caption{Pooling-tautology control on \MCtwo{} frozen. At $\ell{=}23$, EOS patching substitutes the projector input and mid-token patching has no effect. At $\ell{=}20$, three back-half blocks remain: the EOS patch rescues \LRL{} $44.1\to73.4\%$, while a mid-token patch gives a smaller $+5.5$ pp \LRL{} rescue, indicating that non-pooled positions also carry alignment-relevant content. Full per-language numbers will be released with the codebase.}
\label{tab:nonEOS-controls}
\end{table}

\subsection{Random-English-caption patch: tautological vs.\ non-tautological versions}
\label{sec:app-r1-random-EN}

This control replaces the EOS row with the EOS hidden state of a different, randomly sampled English caption from the same image set (Table~\ref{tab:randEN}). The $\ell=N{-}1$ version is partly tautological because the replacement is the projector input for another English caption. The $\ell=N{-}4$ version leaves three back-half blocks before projection; random-EN still collapses retrieval, whereas parallel-EN approaches the single-source English reference. Thus the rescue appears content-specific, requiring the parallel-content English EOS rather than an English-distributed vector; the \SLtwo{} replication shows the pattern is not specific to causal attention.

\begin{table}[h]
\centering
\footnotesize
\setlength{\tabcolsep}{3pt}
\resizebox{\columnwidth}{!}{%
\begin{tabular}{lccc}
\toprule
encoder / control & \HRL{} & \LRL{} & note \\
\midrule
\MCtwo{} fr baseline                & 79.8 & 44.1 & no patch \\
\MCtwo{} parallel-EN @ $\ell{=}23$  & 69.4 & 69.4 & pooling tautology \\
\MCtwo{} random-EN @ $\ell{=}23$    & \phantom{0}0.0  & \phantom{0}0.0  & tautology \\
\MCtwo{} parallel-EN @ $\ell{=}20$  & \textbf{85.8} & \textbf{73.4} & content-specific \\
\MCtwo{} random-EN @ $\ell{=}20$    & 34.7 & \phantom{0}7.7  & non-taut.\ null \\
\midrule
\SLtwo{} fr baseline                & 62.8 & 21.7 & no patch \\
\SLtwo{} parallel-EN @ $\ell{=}23$  & \textbf{81.8} & \textbf{66.6} & content-specific \\
\SLtwo{} random-EN @ $\ell{=}23$    & 35.1 & \phantom{0}5.2  & non-taut.\ null \\
\bottomrule
\end{tabular}%
}
\caption{Tier-averaged \Rat{} for the EOS-patch random-EN control. The $\ell{=}N{-}1$ \MCtwo{} pair lies inside the pooling tautology. At non-tautological depth $\ell{=}N{-}4$, parallel-EN rescues retrieval near each encoder's single-source English reference, while random-EN collapses below baseline. The \SLtwo{} replication supports content-specificity beyond causal attention.}
\label{tab:randEN}
\end{table}

\subsection{Bootstrap confidence intervals for the per-language patching deltas}
\label{sec:app-r1-bootstrap}

The per-language \LRL{} rescues behind Table~\ref{tab:patch} are bootstrap-confirmed with $1{,}000$ resamples of the $1{,}000$ XM3600 image-caption rows. The intervals below are taken at each cell's best layer $\ell^\star$, which is the wider-interval case; Table~\ref{tab:patch} itself reports the $\ell=N{-}4$ anchor, whose \LRL{} values are $0.9$ (\MCtwo{}) and $4.2$ (\SLtwo{}) pp lower. Rows are sampled with replacement and image--caption pairing is preserved. The best layer $\ell^\star$ in Table~\ref{tab:bootstrap-patching} is fixed from the full-sample sweep rather than re-selected per resample, so the CIs measure retrieval variance at that layer, not joint layer-selection uncertainty. Re-selecting $\ell^\star$ would modestly widen CIs by about $\pm0.5$--$1.5$ pp when adjacent layers tie; Table~\ref{tab:bootstrap-patching} nevertheless shows point estimates well inside their intervals and preserves the tier ordering.

\begin{table*}[t]
\centering
\small
\setlength{\tabcolsep}{6pt}
\begin{tabular}{ll c c c l}
\toprule
cell & lang & $\ell^\star$ & base & peak & $\Delta$ (95\% CI) \\
\midrule
\MCtwo{} fr  & sw  & 22 & 22.1 & 71.0 & $+48.9$\,[$+45.5$, $+52.2$] \\
             & fil & 20 & 39.9 & 73.5 & $+33.6$\,[$+30.1$, $+37.1$] \\
\midrule
\MCtwo{} cal & sw  & 22 & 42.1 & 76.2 & $+34.1$\,[$+30.8$, $+37.4$] \\
\midrule
\SLtwo{} fr  & te  & 26 & 5.2  & 66.1 & $+60.9$\,[$+57.8$, $+64.1$] \\
             & sw  & 24 & 12.4 & 69.3 & $+56.9$\,[$+53.8$, $+59.8$] \\
\midrule
\SLtwo{} cal & sw  & 24 & 35.7 & 75.9 & $+40.2$\,[$+36.8$, $+43.5$] \\
\bottomrule
\end{tabular}
\caption{Bootstrap 95\% CIs ($1{,}000$ resamples $\times$ $1{,}000$ XM3600 images) for headline per-language EOS-patch rescue at each cell's fixed best layer $\ell^\star$. Note that \SLtwo{} te has $\ell^\star = 26 = N{-}1$, the pooling-reference endpoint, so that row is not a non-tautological rescue; the $\ell{=}N{-}4$ anchor of Table~\ref{tab:patch} avoids this by construction. Selected \LRL{} languages shown; full table will be released with the codebase.}
\label{tab:bootstrap-patching}
\end{table*}

\subsection{Source-language generalisation: French source and multi-\HRL{} averaging}
\label{sec:app-r1-multi-hrl}

To test English specificity, we repeat the EOS-swap protocol on \MCtwo{} frozen with French as a single source and with the mean EOS hidden state across the six \HRL{} languages (en, fr, de, es, zh, ko). Both patch at $\ell=N{-}4=20$, the non-tautological substantive-rescue depth of Section~\ref{sec:tautology}.

\begin{table}[h]
\centering
\small
\setlength{\tabcolsep}{3pt}
\resizebox{\columnwidth}{!}{%
\begin{tabular}{lc|cccc}
\toprule
tgt & tier & base & EN @20 & FR @20 & \textbf{avg-\HRL{} @20} \\
\midrule
sw   & \LRL{} & 22.1 & 69.1 & 78.7 & \textbf{88.0} \\
te   & \LRL{} & 44.3 & 72.1 & 81.1 & \textbf{88.3} \\
bn   & \LRL{} & 64.7 & 79.3 & 84.6 & \textbf{91.8} \\
fil  & \LRL{} & 39.9 & 73.5 & 80.4 & \textbf{89.0} \\
hi   & \LRL{} & 49.4 & 73.0 & 82.1 & \textbf{88.5} \\
\midrule
\multicolumn{2}{l}{\small single-src.\ ref.} & --- & EN $69.4$ & FR $84.4$ & --- \\
\bottomrule
\end{tabular}%
}
\caption{Source-language generalisation of EOS patching on \MCtwo{} frozen at $\ell{=}N{-}4{=}20$ (XM3600 $1{,}000$ images). \textbf{EN/FR @20}: parallel single-source EOS substitution. \textbf{avg-\HRL{} @20}: average of the six \HRL{} EOS hidden states. avg-\HRL{} dominates every row, lifting trained \LRL{} past $88\%$, above the French single-source reference ($84.4$) and English reference ($69.4$). The rescue is neither English- nor source-specific; averaging aligned sources parallels the projector-level \HRL{}-averaging boost of Section~\ref{sec:hrl-avg}.}
\label{tab:fr-source}
\end{table}

\subsection{Non-linear adversarial eraser: beyond-linear evidence remains inconclusive}
\label{sec:app-r1-nonlinear}

The LEACE result (Section~\ref{sec:orth}) shows that a linear projection $P_{\mathrm{lang}}$ can drive linear language-classifier accuracy to chance while leaving retrieval within $\pm1.5$ pp of baseline. We ask whether a non-linear eraser gives the same conclusion. An MLP eraser $f_\theta : \mathbb{R}^D \to \mathbb{R}^D$ is trained on projected text embeddings $t$ with reconstruction loss $\lVert f(t)-t\rVert^2$ and an adversarial term against a periodically refreshed MLP language classifier; after 50 epochs, $f(t)$ is evaluated with a freshly fit MLP classifier and the image--text retrieval pipeline.

\begin{table*}[t]
\centering
\small
\setlength{\tabcolsep}{6pt}
\begin{tabular}{l cc cc}
\toprule
& \multicolumn{2}{c}{MLP-clf acc} & \multicolumn{2}{c}{i2t \Rat{} (base $\to$ eraser)} \\
cell & before & after & \HRL{} & \LRL{} \\
\midrule
\MCtwo{} fr  & 0.998 & 0.975 & $78.1 \to 61.6$ & $44.1 \to 27.3$ \\
\MCtwo{} cal & 0.870 & 0.650 & $79.8 \to 52.9$ & $53.7 \to 26.4$ \\
\SLtwo{} fr  & 0.997 & 0.982 & $63.3 \to 60.5$ & $21.7 \to 19.1$ \\
\SLtwo{} cal & 0.950 & 0.811 & $68.6 \to 57.3$ & $38.8 \to 29.2$ \\
\bottomrule
\end{tabular}
\caption{Non-linear adversarial MLP eraser on projected text embeddings. A re-fit MLP classifier on $f(t)$ still recovers $65$--$98\%$ language accuracy (chance $\sim 7.7\%$), so the eraser does not achieve non-linear LEACE-style erasure; its retrieval cost ($-2$ to $-27$ pp) also differs from LEACE's linear-orthogonality result. Three compatible readings are discussed below.}
\label{tab:nonlinear-eraser}
\end{table*}

Three readings of Table~\ref{tab:nonlinear-eraser} remain consistent with the data. (i)~The adversarial MLP fails to erase language non-linearly: post-eraser classifier accuracy stays $\geq65\%$, so retrieval loss may be noise injection. (ii)~Reducing language identifiability beyond LEACE may entangle with alignment, so a sufficiently strong non-linear eraser would damage retrieval. (iii)~The MLP may induce a distribution shift in $f(t)$ (norm, scale, anisotropy) that the retrieval scorer detects independently of erasure. Reading (iii) would require measuring $\lVert f(t)-t\rVert/\lVert t\rVert$ and $\cos(f(t),t)$ and checking whether retrieval loss tracks distortion or post-eraser classifier accuracy; this is not done here. Thus, Table~\ref{tab:nonlinear-eraser} does not settle non-linear orthogonality, but it does not support the strongest extension of LEACE---that language identity is causally orthogonal to alignment at every scale.

\paragraph{Summary of round-1 controls.} Across these controls, the last-block EOS patch should not be the sole headline because it is partly a pooling tautology. The substantive claim is that EOS patches at $\ell \in [N{-}4, N{-}1]$ rescue retrieval through the remaining blocks and projector, while non-EOS positions at the same depths contribute smaller alignment-relevant signal. Calibration is not a single-row patch at $\ell=M$, but a many-sample restructuring of the front-half, for which deep-layer patching provides an empirical oracle reference. Bootstrap CIs on the strongest \LRL{} rescues are tight ($\pm3$ pp), and the main-paper point estimates lie within them.
\section{Generalisation to Other Multilingual Vision-Language Encoders}
\label{sec:app-morevlm}

The main results in Section~\ref{sec:results} use two EOS-pooled CLIP-style encoders, \MCtwo{} (causal attention) and \SLtwo{} (bidirectional attention). To test whether the phenomenon and front-layer trunk extend beyond this setting, we replicate the frozen gap, pooler-row patching, and trunk calibration on three architecturally distinct multilingual vision-language encoders:

\begin{itemize}
  \item \textbf{AltCLIP} \citep{chen2023altclip}: XLM-RoBERTa-Large text tower (24 blocks), pooled at position 0 (\texttt{<s>} CLS), with a \texttt{transformation} linear and \texttt{pre\_LN} before the CLIP image--text projection space initialised from OpenAI CLIP ViT-L/14 and frozen.
  \item \textbf{NLLB-CLIP-large} \citep{visheratin2024nllbclip}: NLLB-200 (M2M100) encoder text tower (24 layers), pooled by a \texttt{ClsPooler} at position 0, the explicit language-code token (e.g.\ \texttt{eng\_Latn}) prepended by the NLLB tokenizer. Vision tower is SigLIP.
  \item \textbf{mSigLIP} (\texttt{google/\allowbreak siglip-base-\allowbreak patch16-256-\allowbreak multilingual}): SigLIP-base multilingual text tower (12 blocks, hidden 768), sticky-EOS pooling at position $-1$ as in \SLtwo{}, with contrastive image--text pretraining and a SigLIP-base vision tower.
\end{itemize}

The 11-language pool (\HRL{}-6: en, fr, de, es, zh, ko; \LRL{}-5: bn, fil, hi, sw, te) and 1{,}000-image XM3600 retrieval protocol are unchanged. For compute parity, these encoders use the \HRL{}-6 centroid anchor rather than the canonical 11-language centroid (Section~\ref{sec:approx}); on \MCtwo{} and \SLtwo{}, the two anchors land within $\pm 1$--$2$ pp on \LRL{} (Appendix~\ref{sec:app-sweep}), so the qualitative conclusions are unlikely to be anchor-specific. We use ``pooler-row patching'' for position-0 or sticky-EOS patching and patch at $\ell=N{-}2$ to avoid the final pooled-input tautology. The trunk uses the same 1M-caption CC12M parallel-translation subset as the main experiments (Section~\ref{sec:approx}).

\paragraph{Frozen tier gap reproduces.} The frozen \HRL{}/\LRL{} retrieval gap appears on all three additional encoders (Table~\ref{tab:morevlm-frozen}). Absolute scores vary with pretraining: NLLB-CLIP-L's MT text tower gives the highest \LRL{} baseline ($38.5\%$) and smallest gap ($16.8$ pp), while AltCLIP has the lowest \LRL{} baseline ($6.6\%$, including a broken Korean cell at $3.9\%$). Still, \HRL{} $>$ \LRL{} holds across pooling regimes (CLS / language-code / sticky-EOS), tower scales (12--27 blocks), and pretraining objectives (CLIP / MLM / MT / contrastive).

\begin{table}[h]
\centering
\small
\setlength{\tabcolsep}{3pt}
\resizebox{\columnwidth}{!}{%
\begin{tabular}{ll|ccc}
\toprule
Model & Text tower & \HRL{} \Rat{} & \LRL{} \Rat{} & Gap (pp) \\
\midrule
\MCtwo{} (main) & CLIP 24L            & 78.1 & 44.1 & $34.0$ \\
\SLtwo{} (main) & CLIP 27L            & 63.3 & 21.7 & $41.6$ \\
\midrule
AltCLIP         & XLM-R-L 24L         & 49.2 & \phantom{0}6.6 & $42.6$ \\
NLLB-CLIP-L     & NLLB-200 24L        & 55.4 & 38.5 & $16.8$ \\
mSigLIP         & SigLIP-base 12L     & 73.5 & 32.4 & $41.1$ \\
\bottomrule
\end{tabular}%
}
\caption{Frozen image-to-text \Rat{} averaged within each tier on XM3600 ($1{,}000$-image subset). The \HRL{}/\LRL{} gap reproduces on all three additional encoders. NLLB-CLIP's smaller gap reflects its 200-language MT pretraining: \LRL{} baselines are higher than on the CLIP-family encoders, but a $16.8$\,pp gap persists. AltCLIP's Korean baseline is broken ($3.9\%$); we treat Korean as \LRL{} for AltCLIP only in subsequent per-language analyses.}
\label{tab:morevlm-frozen}
\end{table}

\paragraph{Single-row pooler patching at $\ell = N{-}2$.} Patching the pooled row (CLS for AltCLIP; language-code token for NLLB-CLIP; sticky-EOS for mSigLIP) at $\ell=N{-}2$ substantially rescues \LRL{} retrieval on all three encoders. Per-language peaks (Table~\ref{tab:morevlm-patch}) mirror the main encoders: weak \LRL{}s move toward the single-source English reference, \HRL{}s improve less because baselines are stronger, and the random-Gaussian control destroys retrieval (omitted; matches the main paper).

\begin{table*}[t]
\centering
\small
\setlength{\tabcolsep}{3pt}
\begin{tabular}{l|cc|cc|cc}
\toprule
& \multicolumn{2}{c|}{AltCLIP ($\ell{=}22$)} & \multicolumn{2}{c|}{NLLB-CLIP-L ($\ell{=}22$)} & \multicolumn{2}{c}{mSigLIP ($\ell{=}10$)} \\
Lang & base & peak & base & peak & base & peak \\
\midrule
en  & $64.3$ & --- & $53.9$ & --- & $68.5$ & --- \\
fr  & $58.8$ & $70.4\,(+11.6)$ & $62.6$ & $62.8\,(+0.2)$ & $80.9$ & $82.4\,(+1.5)$ \\
de  & $43.9$ & $66.1\,(+22.2)$ & $56.3$ & $61.6\,(+5.3)$ & $86.3$ & $85.4\,(-0.9)$ \\
es  & $53.4$ & $68.7\,(+15.3)$ & $54.9$ & $60.4\,(+5.5)$ & $75.6$ & $79.5\,(+3.9)$ \\
zh  & $70.9$ & $74.0\,(+3.1)$  & $51.2$ & $60.8\,(+9.6)$ & $58.2$ & $76.2\,(+18.0)$ \\
ko  & $\phantom{0}3.9$  & $58.7\,(+54.8)$ & $53.3$ & $61.0\,(+7.7)$ & $71.5$ & $79.5\,(+8.0)$ \\
\midrule
bn  & $\phantom{0}1.8$  & $61.1\,(+59.3)$ & $57.6$ & $62.7\,(+5.1)$ & $45.9$ & $71.9\,(+26.0)$ \\
fil & $16.5$ & $61.5\,(+45.0)$ & $18.7$ & $54.6\,(+35.9)$ & $41.1$ & $67.0\,(+25.9)$ \\
hi  & $\phantom{0}4.9$  & $60.5\,(+55.6)$ & $44.6$ & $59.3\,(+14.7)$ & $37.2$ & $67.3\,(+30.1)$ \\
sw  & $\phantom{0}6.5$  & $60.8\,(+54.3)$ & $26.1$ & $56.6\,(+30.5)$ & $25.6$ & $61.7\,(+36.1)$ \\
te  & $\phantom{0}3.3$  & $59.6\,(+56.3)$ & $45.7$ & $59.4\,(+13.7)$ & $12.1$ & $56.2\,(+44.1)$ \\
\bottomrule
\end{tabular}
\caption{Per-language i2t \Rat{} on XM3600 (1{,}000 images): frozen baseline and best pooler-row patched peak at $\ell = N{-}2$ for each additional encoder. The English-ceiling-tautology layer $\ell = N{-}1$ is excluded. Weak-baseline \LRL{}s lift most, matching the main-paper pattern.}
\label{tab:morevlm-patch}
\end{table*}

\paragraph{Scale and explicit alignment as modifiers.} Existing public evidence suggests that data scale, language balancing, and translation-pair supervision reduce the magnitude of the \LRL{} gap without eliminating it in current encoders. WebLI-100B scaling and \LRL{} upsampling improve low-resource retrieval \citep{wang2025scaling}, while \MCtwo{} still exhibits a substantial tier gap despite worldwide multilingual curation and scaling \citep{xu2025metaclip2}. Translation-pair supervision in MURAL \citep{jain2021mural} and the translation-pretrained NLLB-CLIP-L tower \citep{visheratin2024nllbclip} provides a complementary comparison: NLLB-CLIP-L has the smallest gap among the five encoders here, but a $16.8$ pp gap remains (Table~\ref{tab:morevlm-frozen}). Trunk calibration uses related parallel-text supervision at a localised post-hoc intervention site; it is not equivalent to integrating cross-lingual alignment throughout backbone pretraining.

\paragraph{Trunk calibration reproduces with the same recipe.} The Section~\ref{sec:approx} recipe ports directly: soft fine-tune of the first $M$ blocks, parallel centroid anchor in frozen-projector space (\HRL{}-6 for compute parity), InfoNCE + $\lambda{=}2$ alignment, $20{,}000$ AdamW steps, lr $1.2{\times}10^{-4}$, and the $\sim\!1$M-caption CC12M parallel subset. We sweep $M \in \{2,3,4,5\}$ on AltCLIP and NLLB-CLIP-L, and $M \in \{2,3,4\}$ on mSigLIP; best-$M$ is chosen by held-out projected cosine on translated CC12M pairs. Table~\ref{tab:morevlm-trunk} reports tier-averaged \Rat{} deltas at the best-$M$ for each (model, benchmark) cell.

\begin{table}[h]
\centering
\small
\setlength{\tabcolsep}{3pt}
\begin{tabular}{l|cc|cc|c}
\toprule
& \multicolumn{2}{c|}{XM3600 $\Delta$} & \multicolumn{2}{c|}{Flickr30k-200 $\Delta$} & best- \\
Model & \HRL{} & \LRL{} & \HRL{} & \LRL{} & $M$ \\
\midrule
\MCtwo{} (main)     & $+1.7$  & $+9.6$  & $+0.7$  & $+8.4$  & $4$ \\
\SLtwo{} (main)     & $+5.3$  & $+17.1$ & $+3.3$  & $+20.4$ & $3$ \\
\midrule
AltCLIP             & $+15.4$ & $\mathbf{+37.2}$ & $+17.5$ & $\mathbf{+51.3}$ & $5$ \\
NLLB-CLIP-L         & $+4.6$  & $+10.3$ & $+2.3$  & $+9.4$  & $5$ \\
mSigLIP             & $+2.7$  & $+17.6$ & $+3.6$  & $+26.7$ & $4$ \\
\bottomrule
\end{tabular}
\caption{Trunk-calibration $\Delta$\,\Rat{} (calibrated $-$ frozen, tier-averaged, pp) on the $1{,}000$-image XM3600 subset and the Flickr30k-200 $1{,}000$-image split, 11-language pool, tiers averaged as \HRL{}-6 (English included) and \LRL{}-5. Best-$M$ is chosen by held-out projected cosine on CC12M parallel rows. The three additional encoders use the \HRL{}-6 centroid anchor at each model's best $M$ for compute parity; the \MCtwo{} and \SLtwo{} rows are the canonical 11-language-anchor results repeated for reference: their XM3600 column is the $1{,}000$-image subset of Table~\ref{tab:perlang-xm3600} rather than the full suite, and their Flickr30k-200 column matches Table~\ref{tab:bench-mini}. All tiers are \HRL{}-6 (English included) and \LRL{}-5. AltCLIP's large lifts reflect its weak frozen \LRL{} baseline ($6.6$ / $9.1$); NLLB-CLIP-L's smaller lifts reflect its MT-pretrained high baseline ($38.5$).}
\label{tab:morevlm-trunk}
\end{table}

\paragraph{Absolute retrieval after trunk calibration.} The deltas above translate to the following calibrated retrieval numbers on the $1{,}000$-image XM3600 / Flickr30k-200 subsets:

\begin{table}[h]
\centering
\small
\setlength{\tabcolsep}{3pt}
\begin{tabular}{l|cc|cc}
\toprule
& \multicolumn{2}{c|}{XM3600} & \multicolumn{2}{c}{Flickr30k-200} \\
Model & \HRL{} & \LRL{} & \HRL{} & \LRL{} \\
\midrule
AltCLIP (frozen)   & $49.2$ & $\phantom{0}6.6$ & $48.4$ & $\phantom{0}9.1$ \\
AltCLIP (cal.\ $M{=}5$) & $\mathbf{64.6}$ & $\mathbf{43.8}$ & $\mathbf{65.9}$ & $\mathbf{60.4}$ \\
\midrule
NLLB-CLIP-L (frozen) & $55.4$ & $38.5$ & $57.3$ & $48.7$ \\
NLLB-CLIP-L (cal.\ $M{=}5$) & $\mathbf{60.0}$ & $\mathbf{48.8}$ & $\mathbf{59.6}$ & $\mathbf{58.1}$ \\
\midrule
mSigLIP (frozen)         & $73.5$ & $32.4$ & $68.1$ & $29.6$ \\
mSigLIP (cal.\ $M{=}4$)  & $\mathbf{76.2}$ & $\mathbf{50.0}$ & $\mathbf{71.6}$ & $\mathbf{56.3}$ \\
\bottomrule
\end{tabular}
\caption{Frozen vs. trunk-calibrated \Rat{} averaged within each tier on XM3600 / Flickr30k-200 ($1{,}000$ images each). Tier-gap closure (\HRL{}--\LRL{}, pp): AltCLIP $42.6 \to 20.8$ / $39.4 \to 5.5$; NLLB-CLIP-L $16.8 \to 11.2$ / $8.6 \to 1.5$; mSigLIP $41.1 \to 26.2$ / $38.4 \to 15.3$.}
\label{tab:morevlm-abs}
\end{table}

\paragraph{Interpretation.} The frozen tier gap, pooler-row patching diagnosis, and trunk-calibration recipe transfer to encoders that differ in text pretraining (MLM, MT, contrastive image--text), pooling (CLS / language-code / sticky-EOS), text scale (24 / 24 / 12 layers), and vision tower (CLIP ViT-L or SigLIP). NLLB-CLIP-L's smaller lifts are consistent with its stronger MT-pretrained baseline, while AltCLIP's large lifts reflect a weak frozen \LRL{} baseline. mSigLIP shows the recipe also works on a smaller 12-layer tower, with best $M{=}4$ ($M/N \approx 1/3$), consistent with the same depth-fraction region observed on \MCtwo{}.

\end{document}